\documentclass[lettersize,journal]{IEEEtran}

\usepackage{gensymb}
\usepackage{dsfont}
\usepackage{amsmath,amssymb,amsfonts,amsthm}
\usepackage{epsfig}
\usepackage{cite}
\usepackage{hhline}
\usepackage{multirow}
\usepackage{xcolor}
\usepackage{makecell}
\usepackage{subcaption}
\usepackage{capt-of}
\usepackage[labelformat=simple]{subcaption}

\usepackage{enumerate}
\usepackage{enumitem}
\usepackage{color}
\usepackage{graphicx}
\usepackage{algpseudocode}
\usepackage[ruled]{algorithm}
\makeatletter
\newcounter{parentalgorithm}

\makeatother

\usepackage[nolist,printonlyused]{acronym}      
\usepackage{booktabs}
\usepackage{bm}
\usepackage{pgf}
\usepackage{pgfplots}
\usepackage{tikz}
\usepackage{grffile}
\pgfplotsset{compat=newest}
\usetikzlibrary{plotmarks}
\usetikzlibrary{shapes.geometric,quotes,angles,automata,arrows,positioning,calc,3d,matrix,decorations.markings,matrix}

\tikzstyle{startstop} = [rectangle, rounded corners, minimum width=3cm, minimum height=1cm,text centered, draw=black, fill=white!30]
\tikzstyle{io} = [trapezium, trapezium left angle=70, trapezium right angle=110, minimum width=3cm, minimum height=1cm, text centered, draw=black, fill=white!30]
\tikzstyle{process} = [rectangle, minimum width=2cm, minimum height=1cm, text centered, draw=black, fill=white!30]
\tikzstyle{decision} = [diamond, minimum width=3cm, minimum height=1cm, text centered, draw=black, fill=white!30]
\tikzstyle{arrow} = [thick,->,>=stealth]

\makeatletter
\newcommand{\vast}{\bBigg@{3}}
\newcommand{\Vast}{\bBigg@{4}}
\makeatother
\renewcommand{\IEEEQED}{\IEEEQEDopen}

\showoutput
\makeatletter
\tikzset{%
  remember picture with id/.style={%
    remember picture,
    overlay,
    save picture id=#1,
  },
  save picture id/.code={%
    \edef\pgf@temp{#1}%
    \immediate\write\pgfutil@auxout{%
      \noexpand\savepointas{\pgf@temp}{\pgfpictureid}}%
  },
  if picture id/.code args={#1#2#3}{%
    \@ifundefined{save@pt@#1}{%
      \pgfkeysalso{#3}%
    }{
      \pgfkeysalso{#2}%
    }
  }
}

\def\savepointas#1#2{%
  \expandafter\gdef\csname save@pt@#1\endcsname{#2}%
}

\def\tmk@labeldef#1,#2\@nil{%
  \def\tmk@label{#1}%
  \def\tmk@def{#2}%
}

\tikzdeclarecoordinatesystem{pic}{%
  \pgfutil@in@,{#1}%
  \ifpgfutil@in@%
    \tmk@labeldef#1\@nil
  \else
    \tmk@labeldef#1,(0pt,0pt)\@nil
  \fi
  \@ifundefined{save@pt@\tmk@label}{%
    \tikz@scan@one@point\pgfutil@firstofone\tmk@def
  }{%
  \pgfsys@getposition{\csname save@pt@\tmk@label\endcsname}\save@orig@pic%
  \pgfsys@getposition{\pgfpictureid}\save@this@pic%
  \pgf@process{\pgfpointorigin\save@this@pic}%
  \pgf@xa=\pgf@x
  \pgf@ya=\pgf@y
  \pgf@process{\pgfpointorigin\save@orig@pic}%
  \advance\pgf@x by -\pgf@xa
  \advance\pgf@y by -\pgf@ya
  }%
}

\makeatother

\usepackage{comment}
\usepackage{xpatch}
\makeatletter
\xpatchcmd{\algorithmic}{\itemsep\z@}{\itemsep=-0.25mm}{}{}
\makeatother

\usepackage{matlab-prettifier}

\usepackage{multicol}
\usepackage{multirow}

\usepackage{lipsum}
\usepackage{multicol}
\begin{document}

\title{FedCau: A Proactive Stop Policy for Communication and Computation Efficient Federated Learning$^{**}$}

\author{Afsaneh Mahmoudi$^{1}$, Hossein S. Ghadikolaei$^{2}$, José Mairton Barros Da Silva Júnior$^{3}$, and Carlo Fischione$^{1,4}$
\thanks{$^{**}$ This paper version has been accepted to be published in IEEE Transactions on Wireless Communications, DOI: 10.1109/TWC.2024.3378351.}

\thanks{$^{1}$School of Electrical Engineering and Computer Science, KTH Royal Institute of Technology, Stockholm, Sweden (\{afmb, carlofi\}@kth.se).}

\thanks{$^{2}$Ericsson,~Stockholm,~Sweden (hossein.shokri.ghadikolaei@ericsson.com).}
\thanks{$^{3}$Department of Information Technology, Uppsala University, Uppsala, Sweden (mairton.barros@it.uu.se).}
\thanks{$^{3}$José Mairton B. Da Silva Jr. was jointly supported by the European Union’s Horizon Europe research and innovation program under the Marie Skłodowska-Curie project FLASH, with grant agreement No 101067652; the Ericsson Research Foundation, and the Hans Werthén Foundation.}
\thanks{$^{4}$Carlo Fischione was partially supported by Digital Futures and VR.}
}


\newtheorem{theorem}{Theorem}
\newtheorem{defin}{Definition}
\newtheorem{prop}{Proposition}
\newtheorem{lemma}{Lemma}
\newtheorem{corollary}{Corollary}
\newtheorem{alg}{Algorithm}
\newtheorem{remark}{Remark}
\newtheorem{result}{Result}
\newtheorem{conjecture}{Conjecture}
\newtheorem{example}{Example}
\newtheorem{notations}{Notations}
\newtheorem{assumption}{Assumption}
\newcommand{\combin}[2]{\ensuremath{ \left( \ba{c} #1 \\ #2 \ea \right) }}
\newcommand{\diag}{{\mbox{diag}}}
\newcommand{\rank}{{\mbox{rank}}}
\newcommand{\dom}{{\mbox{dom{\color{white!100!black}.}}}}
\newcommand{\range}{{\mbox{range{\color{white!100!black}.}}}}
\newcommand{\image}{{\mbox{image{\color{white!100!black}.}}}}
\newcommand{\herm}{^{\mbox{\scriptsize H}}}  
\newcommand{\sherm}{^{\mbox{\tiny H}}}       
\newcommand{\tran}{^{\mbox{\scriptsize T}}}  
\newcommand{\tranIn}{^{\mbox{-\scriptsize T}}}  
\newcommand{\card}{{\mbox{\textbf{card}}}}
\newcommand{\asign}{{\mbox{$\colon\hspace{-2mm}=\hspace{1mm}$}}}
\newcommand{\ssum}[1]{\mathop{ \textstyle{\sum}}_{#1}}

\newcommand{\vbar}{\raisebox{.17ex}{\rule{.04em}{1.35ex}}}
\newcommand{\vbarind}{\raisebox{.01ex}{\rule{.04em}{1.1ex}}}
\newcommand{\D}{\ifmmode {\rm I}\hspace{-.2em}{\rm D} \else ${\rm I}\hspace{-.2em}{\rm D}$ \fi}
\newcommand{\T}{\ifmmode {\rm I}\hspace{-.2em}{\rm T} \else ${\rm I}\hspace{-.2em}{\rm T}$ \fi}
\newcommand{\B}{\ifmmode {\rm I}\hspace{-.2em}{\rm B} \else \mbox{${\rm I}\hspace{-.2em}{\rm B}$} \fi}
\newcommand{\Hil}{\ifmmode {\rm I}\hspace{-.2em}{\rm H} \else \mbox{${\rm I}\hspace{-.2em}{\rm H}$} \fi}
\newcommand{\C}{\ifmmode \hspace{.2em}\vbar\hspace{-.31em}{\rm C} \else \mbox{$\hspace{.2em}\vbar\hspace{-.31em}{\rm C}$} \fi}
\newcommand{\Cind}{\ifmmode \hspace{.2em}\vbarind\hspace{-.25em}{\rm C} \else \mbox{$\hspace{.2em}\vbarind\hspace{-.25em}{\rm C}$} \fi}
\newcommand{\Q}{\ifmmode \hspace{.2em}\vbar\hspace{-.31em}{\rm Q} \else \mbox{$\hspace{.2em}\vbar\hspace{-.31em}{\rm Q}$} \fi}
\newcommand{\Z}{\ifmmode {\rm Z}\hspace{-.28em}{\rm Z} \else ${\rm Z}\hspace{-.38em}{\rm Z}$ \fi}

\newcommand{\sgn}{\mbox {sgn}}
\newcommand{\var}{\mbox {var}}
\newcommand{\E}{\mbox {E}}
\newcommand{\cov}{\mbox {cov}}
\renewcommand{\Re}{\mbox {Re}}
\renewcommand{\Im}{\mbox {Im}}
\newcommand{\cum}{\mbox {cum}}

\renewcommand{\vec}[1]{{\bf{#1}}}     

\newcommand{\vecsc}[1]{\mbox {\boldmath \scriptsize $#1$}}     
\newcommand{\itvec}[1]{\mbox {\boldmath $#1$}}
\newcommand{\itvecsc}[1]{\mbox {\boldmath $\scriptstyle #1$}}
\newcommand{\gvec}[1]{\mbox{\boldmath $#1$}}

\newcommand{\balpha}{\mbox {\boldmath $\alpha$}}
\newcommand{\bbeta}{\mbox {\boldmath $\beta$}}
\newcommand{\bgamma}{\mbox {\boldmath $\gamma$}}
\newcommand{\bdelta}{\mbox {\boldmath $\delta$}}
\newcommand{\bepsilon}{\mbox {\boldmath $\epsilon$}}
\newcommand{\bvarepsilon}{\mbox {\boldmath $\varepsilon$}}
\newcommand{\bzeta}{\mbox {\boldmath $\zeta$}}
\newcommand{\boldeta}{\mbox {\boldmath $\eta$}}
\newcommand{\btheta}{\mbox {\boldmath $\theta$}}
\newcommand{\bvartheta}{\mbox {\boldmath $\vartheta$}}
\newcommand{\biota}{\mbox {\boldmath $\iota$}}
\newcommand{\blambda}{\mbox {\boldmath $\lambda$}}
\newcommand{\bmu}{\mbox {\boldmath $\mu$}}
\newcommand{\bnu}{\mbox {\boldmath $\nu$}}
\newcommand{\bxi}{\mbox {\boldmath $\xi$}}
\newcommand{\bpi}{\mbox {\boldmath $\pi$}}
\newcommand{\bvarpi}{\mbox {\boldmath $\varpi$}}
\newcommand{\brho}{\mbox {\boldmath $\rho$}}
\newcommand{\bvarrho}{\mbox {\boldmath $\varrho$}}
\newcommand{\bsigma}{\mbox {\boldmath $\sigma$}}
\newcommand{\bvarsigma}{\mbox {\boldmath $\varsigma$}}
\newcommand{\btau}{\mbox {\boldmath $\tau$}}
\newcommand{\bupsilon}{\mbox {\boldmath $\upsilon$}}
\newcommand{\bphi}{\mbox {\boldmath $\phi$}}
\newcommand{\bvarphi}{\mbox {\boldmath $\varphi$}}
\newcommand{\bchi}{\mbox {\boldmath $\chi$}}
\newcommand{\bpsi}{\mbox {\boldmath $\psi$}}
\newcommand{\bomega}{\mbox {\boldmath $\omega$}}

\newcommand{\R}{\mathbb{R}}
\newcommand{\N}{\mathbb{N}}

\def\calA{{\mathcal A}}
\def\calB{{\mathcal B}}
\def\calC{{\mathcal C}}
\def\calD{{\mathcal D}}
\def\calE{{\mathcal E}}
\def\calF{{\mathcal F}}
\def\calG{{\mathcal G}}
\def\calH{{\mathcal H}}
\def\calI{{\mathcal I}}
\def\calJ{{\mathcal J}}
\def\calK{{\mathcal K}}
\def\calL{{\mathcal L}}
\def\calM{{\mathcal M}}
\def\calN{{\mathcal N}}
\def\calO{{\mathcal O}}
\def\calP{{\mathcal P}}
\def\calQ{{\mathcal Q}}
\def\calR{{\mathcal R}}
\def\calS{{\mathcal S}}
\def\calT{{\mathcal T}}
\def\calU{{\mathcal U}}
\def\calV{{\mathcal V}}
\def\calW{{\mathcal W}}
\def\calX{{\mathcal X}}
\def\calY{{\mathcal Y}}
\def\calZ{{\mathcal Z}}

\def\bA{\mbox {\boldmath $A$}}
\def\bB{\mbox {\boldmath $B$}}
\def\bC{\mbox {\boldmath $C$}}
\def\bD{\mbox {\boldmath $D$}}
\def\bE{\mbox {\boldmath $E$}}
\def\bF{\mbox {\boldmath $F$}}
\def\bG{\mbox {\boldmath $G$}}
\def\bH{\mbox {\boldmath $H$}}
\def\bI{\mbox {\boldmath $I$}}
\def\bJ{\mbox {\boldmath $J$}}
\def\bK{\mbox {\boldmath $K$}}
\def\bL{\mbox {\boldmath $L$}}
\def\bM{\mbox {\boldmath $M$}}
\def\bN{\mbox {\boldmath $N$}}
\def\bO{\mbox {\boldmath $O$}}
\def\bP{\mbox {\boldmath $P$}}
\def\bQ{\mbox {\boldmath $Q$}}
\def\bR{\mbox {\boldmath $R$}}
\def\bS{\mbox {\boldmath $S$}}
\def\bT{\mbox {\boldmath $T$}}
\def\bU{\mbox {\boldmath $U$}}
\def\bV{\mbox {\boldmath $V$}}
\def\bW{\mbox {\boldmath $W$}}
\def\bX{\mbox {\boldmath $X$}}
\def\bY{\mbox {\boldmath $Y$}}
\def\bZ{\mbox {\boldmath $Z$}}

\def\ba{\mbox {$\bf{a}$}}
\def\bb{\mbox {\boldmath $b$}}
\def\bc{\mbox {\boldmath $c$}}
\def\bd{\mbox {\boldmath $d$}}
\def\be{\mbox {\boldmath $e$}}
\def\bg{\mbox {\boldmath $g$}}
\def\bh{\mbox {\boldmath $h$}}
\def\bi{\mbox {\boldmath $i$}}
\def\bj{\mbox {\boldmath $j$}}
\def\bk{\mbox {\boldmath $k$}}
\def\bl{\mbox {\boldmath $l$}}
\def\bm{\mbox {\boldmath $m$}}
\def\bn{\mbox {\boldmath $n$}}
\def\bo{\mbox {\boldmath $o$}}
\def\bp{\mbox {\boldmath $p$}}
\def\bq{\mbox {\boldmath $q$}}
\def\br{\mbox {\boldmath $r$}}
\def\bs{\mbox {\boldmath $s$}}
\def\bt{\mbox {\boldmath $t$}}
\def\bu{\mbox {\boldmath $u$}}
\def\bv{\mbox {\boldmath $v$}}
\def\bw{\mbox {\boldmath $w$}}
\def\bx{\mbox {\boldmath $x$}}
\def\by{\mbox {\boldmath $y$}}
\def\bz{\mbox {\boldmath $z$}}

\newcommand{\snr}{\textup{SNR}}
\newcommand{\UE}{\mathrm{UE}}
\newcommand{\BS}{\mathrm{BS}}
\newcommand{\Passoc}{p_{_{I^{(1)}}}}
\newcommand{\Pintra}{p_{_{I^{(2)}}}}
\newcommand{\Pinter}{p_{_{I^{(3)}}}}

\newenvironment{Ex}
{\begin{adjustwidth}{0.04\linewidth}{0cm}
\begingroup\small
\vspace{-1.0em}
\raisebox{-.2em}{\rule{\linewidth}{0.3pt}}
\begin{example}
}
{
\end{example}
\vspace{-5mm}
\rule{\linewidth}{0.3pt}
\endgroup
\end{adjustwidth}}

\newcommand{\Hossein}[1]{{\textcolor{blue}{\emph{**Hossein: #1**}}}}
\newcommand{\Gabor}[1]{{\textcolor{cyan}{\emph{**Afsaneh: #1**}}}}
\newcommand{\Hadi}[1]{{\textcolor{red}{#1}}}
\newcommand{\gf}[1]{{\textcolor{cyan}{#1}}}
\newcommand{\REV}[1]{{\textcolor{blue}{#1}}}


\makeatletter
\let\save@mathaccent\mathaccent
\newcommand*\if@single[3]{%
  \setbox0\hbox{${\mathaccent"0362{#1}}^H$}%
  \setbox2\hbox{${\mathaccent"0362{\kern0pt#1}}^H$}%
  \ifdim\ht0=\ht2 #3\else #2\fi
  }
\newcommand*\rel@kern[1]{\kern#1\dimexpr\macc@kerna}
\newcommand*\widebar[1]{\@ifnextchar^{{\wide@bar{#1}{0}}}{\wide@bar{#1}{1}}}
\newcommand*\wide@bar[2]{\if@single{#1}{\wide@bar@{#1}{#2}{1}}{\wide@bar@{#1}{#2}{2}}}
\newcommand*\wide@bar@[3]{%
  \begingroup
  \def\mathaccent##1##2{%
    \let\mathaccent\save@mathaccent
    \if#32 \let\macc@nucleus\first@char \fi
    \setbox\z@\hbox{$\macc@style{\macc@nucleus}_{}$}%
    \setbox\tw@\hbox{$\macc@style{\macc@nucleus}{}_{}$}%
    \dimen@\wd\tw@
    \advance\dimen@-\wd\z@
    \divide\dimen@ 3
    \@tempdima\wd\tw@
    \advance\@tempdima-\scriptspace
    \divide\@tempdima 10
    \advance\dimen@-\@tempdima
    \ifdim\dimen@>\z@ \dimen@0pt\fi
    \rel@kern{0.6}\kern-\dimen@
    \if#31
      \overline{\rel@kern{-0.6}\kern\dimen@\macc@nucleus\rel@kern{0.4}\kern\dimen@}%
      \advance\dimen@0.4\dimexpr\macc@kerna
      \let\final@kern#2%
      \ifdim\dimen@<\z@ \let\final@kern1\fi
      \if\final@kern1 \kern-\dimen@\fi
    \else
      \overline{\rel@kern{-0.6}\kern\dimen@#1}%
    \fi
  }%
  \macc@depth\@ne
  \let\math@bgroup\@empty \let\math@egroup\macc@set@skewchar
  \mathsurround\z@ \frozen@everymath{\mathgroup\macc@group\relax}%
  \macc@set@skewchar\relax
  \let\mathaccentV\macc@nested@a
  \if#31
    \macc@nested@a\relax111{#1}%
  \else
    \def\gobble@till@marker##1\endmarker{}%
    \futurelet\first@char\gobble@till@marker#1\endmarker
    \ifcat\noexpand\first@char A\else
      \def\first@char{}%
    \fi
    \macc@nested@a\relax111{\first@char}%
  \fi
  \endgroup
}
\makeatother

\def\herm{\mathsf{H}}
\def\trans{\mathsf{T}}
\newcommand{\call}[1]{{\textsf{\small \textsc{#1}}}}
\newcommand{\callf}[1]{{\textsf{\footnotesize \textsc{#1}}}}

\def\argmax{\mathrm{arg}\max}
\def\argmin{\mathrm{arg}\min}
\renewcommand{\algorithmicrequire}{\textbf{Input:}}
\renewcommand{\algorithmicensure}{\textbf{Output:}}
\algdef{SE}[PROCEDURE]{Procedure}{EndProcedure}%
   [2]{\algorithmicprocedure\ \textproc{#1}\ifthenelse{\equal{#2}{}}{}{(#2)}}%
   {\algorithmicend\ \algorithmicprocedure}%
\algdef{SE}[FUNCTION]{Function}{EndFunction}%
   [2]{\algorithmicfunction\ \textproc{#1}\ifthenelse{\equal{#2}{}}{}{(#2)}}%
   {\algorithmicend\ \algorithmicfunction}%

\makeatletter
\newcommand\fs@betterruled{%
  \def\@fs@cfont{\bfseries}\let\@fs@capt\floatc@ruled
  \def\@fs@pre{\vspace*{5pt}\hrule height.8pt depth0pt \kern2pt}%
  \def\@fs@post{\kern2pt\hrule\relax}%
  \def\@fs@mid{\kern2pt\hrule\kern2pt}%
  \let\@fs@iftopcapt\iftrue}
\floatstyle{betterruled}
\restylefloat{algorithm}
\makeatother

\definecolor{fuzzywuzzy}{rgb}{0.8, 0.4, 0.4}

\maketitle

\begin{abstract}
This paper investigates efficient distributed training of a Federated Learning~(FL) model over a wireless network of wireless devices. The communication iterations of the distributed training algorithm may be substantially deteriorated or even blocked by the effects of the devices' background traffic, packet losses, congestion, or latency. We abstract the communication-computation impacts as an `iteration cost' and propose a cost-aware causal FL algorithm~(FedCau) to tackle this problem. We propose an iteration-termination method that trade-offs the training performance and networking costs. We apply our approach when workers use the slotted-ALOHA, carrier-sense multiple access with collision avoidance~(CSMA/CA), and orthogonal frequency-division multiple access~(OFDMA) protocols. We show that, given a total cost budget, the training performance degrades as either the background communication traffic or the dimension of the training problem increases. Our results demonstrate the importance of proactively designing optimal cost-efficient stopping criteria to avoid unnecessary communication-computation costs to achieve a marginal FL training improvement. We validate our method by training and testing FL over the MNIST {\color{black}and CIFAR-10} dataset. Finally, we apply our approach to existing communication efficient FL methods from the literature, achieving further efficiency. We conclude that cost-efficient stopping criteria are essential for the success of practical FL over wireless networks.
\end{abstract}

\section{Introduction}\label{sec: Introduction}
 The recent success of artificial intelligence and large-scale machine learning heavily relies on the advancements of distributed optimization algorithms~\cite{jordan2015machine}. The main objective of such algorithms is better training/test performance for prediction and inference tasks, such as image recognition~\cite{8542764}. However, the costs of running the algorithms over a wireless network may hinder achieving the desired training accuracy due to the communication and computation costs. The state-of-the-art of such algorithms requires powerful computing platforms with vast amounts of computational and communication resources. Although such resources are available in modern data centers that use wired networks, they are not easily available in wireless devices due to communication and energy resource constraints. Yet, there is a need to extend machine learning tasks to wireless communication scenarios. Use cases as machine leaning over IoT, edge computing, or public wireless networks serving many classes of traffic~\cite{park2018wireless}.

 One of these prominent algorithms is Federated Learning~(FL), which is a new machine learning paradigm where each individual worker has to contribute to the learning process without sharing their own data with other workers and the master node. Specifically, FL methods refer to a class of privacy-preserving distributed learning algorithms in which individual workers $[M]$ execute some local iterations and share only their parameters, with a central controller for global model aggregation~\cite{kairouz2019advances}. The FL problem consists in  optimizing a finite sum of $M$ differentiable functions $f_j$, $j \in [M]$, which take inputs from $\R^d$ for some positive $d$ and give their outputs in $ \R $, i.e., $\{f_{j}: \R^d \mapsto \R\}_{j \in [M]}$ with corresponding local parameters $\{\bw^{j} \in \R^d\}_{j \in [M]}$.
The common solution to such a problem involves an iterative procedure wherein at each global communication iteration $k$, workers have to find the local parameter $\{\bw_k^{j} \}_{j \in [M]}$ and upload them to a central controller. Then, the master node updates the model parameters as $\bw_{k+1}$ and broadcasts it to all the nodes to start the next iteration~\cite{konevcny2016federated}. 

The FL algorithm alleviates computation and privacy by parallel computations at workers using their local private data~\cite{ konevcny2016federated}. 
However, such an algorithm introduces a communication
cost: parameter vectors, such as weight and bias, must be communicated between the master and the workers to run a new iteration. The weights can be vectors of huge sizes whose frequent transmissions and reception may deplete the battery of wireless devices. Therefore, every communication iteration of these algorithms suffers some costs~\footnote{Throughout the paper, we use ``communication-computation cost'' and ``iteration cost'' interchangeably.}, 
 e.g., computation, latency, communication resource utilization, and energy. As we argue in this paper, the communication cost can be orders of magnitude larger than the computation costs, thus making the iterative procedure over wireless networks potentially very inefficient. Moreover, due to the \textit{diminishing return} rule~\cite{9563954}, the accuracy improvement of the final model gets smaller with every new iteration. Yet, it is necessary to pay an expensive communication cost to run every new communication iteration of marginal importance for training purposes.

In this paper, we investigate the problem of FL over wireless networks to ensure an efficient communication-computation cost. Specifically, we define our FL over wireless networks as follows. We consider a star network topology and focus on avoiding the extra communication-computation cost paid in FL training to attain a marginal improvement. We show that a negligible improvement in training spends valuable resources and hardly results in test accuracy progress. We propose novel and causal cost-efficient FL algorithms~(FedCau) for both convex and non-convex loss functions. We show the significant performance improvements introduced by FedCau through experimental results, where we train the FL model over the wireless networks with slotted-ALOHA, CSMA/CA, and OFDMA protocols. We apply FedCau on top of two well-known communication-efficient methods, {\color{black}Top-$q$~\cite{ 8889996}, and LAQ~\cite{9238427}} and the results show that FedCau algorithms further improve the communication efficiency of other communication-efficient methods from the literature. Our extensive results show that the FedCau methods can save the valuable resources one would spend through unnecessary iterations of FL, even when applied on top of existing methods from literature focusing on resource allocation problems~\cite{konevcny2016federated, caldas2018expanding, fan2018application, 9264742}.

\subsection{Literature Survey}
Cost-efficient distributed training is addressed in the literature through communication-efficiency~\cite{stich2018sparsified,  di2018efficient,yuan2014communication, wangni2018gradient, chen2018lag, sun2019communication, kairouz2019advances, hsieh2017gaia, yu2018parallel} or tradeoff between computation and communication primarily by resource allocation~\cite{9264742,9127160}. Mainly, we have two classes of approaches for communication-efficiency in the
literature focusing on 1) data compression, like quantization and sparsification of the local parameters in every iteration, and 2) communication iteration reduction.

The first class of approaches focuses on data compression, which reduces the amount of information exchanged in bits among nodes, thereby saving communication resources. However, we may need more iterations to compensate for quantization errors than the unquantized version.
Recent studies have shown that proper quantization approaches, together with some error feedback, can maintain the convergence of the training algorithm and the asymptotic convergence rate~\cite{stich2018sparsified, di2018efficient}.
~However, the improved convergence rates depend on the number of iterations, thus, requiring more computation resources to perform those iterations. Sparsification is an alternative approach to quantization to reduce the amount of exchanged data for running every iteration~\cite{yuan2014communication}. A prominent example of this approach is top-$q$ sparsification, where a node sends only the $q$ most significant entries, such as the ones with the highest modulus, of the stochastic gradient~\cite{stich2018sparsified, wangni2018gradient}. 

The second class of approaches focuses on the reduction of the communication iterations by eliminating the communication between some of the workers and the master node in some iterations~\cite{chen2018lag}. The work~\cite{chen2018lag} has proposed lazily aggregated gradient
(LAG) for communication-efficient distributed learning in master-worker architectures. In LAG, each worker reports its gradient vector to the master node only if the gradient changes from the last communication iteration are large enough. Hence, some nodes may skip sending their gradients at some iterations,
which saves communication resources. LAG has been extended in~\cite{sun2019communication} by sending quantized versions of the gradient vectors. 
In~\cite{yu2018parallel}, local SGD  techniques reduce the number of communication rounds needed to solve an optimization problem. In a generic FL setting, adding more local iterations may reduce the need for frequent global aggregation, leading to a lower communication overhead~\cite{kairouz2019advances}. Moreover, it allows the master node to update the global model with only a (randomly chosen) subset of the nodes at every iteration, which may further reduce the communication overhead and increase the robustness. The work in~\cite{hsieh2017gaia} has improved the random selection of the nodes and proposed the notion of significance filter, where each worker updates its local model and transmits it to the master node only when there is a significant change in the local parameters. 
Furthermore,~\cite{hsieh2017gaia} has shown that adding a memory unit at the master node and using ideas from SAGA~\cite{Defazio2014SAGA} reduce the upload frequency of each worker, thus improving the communication efficiency.


The two classes mentioned above present opportunities for reducing the cost of running distributed training algorithms and adapting them to wireless communication protocols. However, these classes focus primarily on the complexity of the iterative algorithm in terms of bits per communication round or the number of communication rounds~\cite{caldas2018expanding}. Moreover, they neglect other crucial costs associated with solving federated learning (FL) problems, such as latency~\cite{park2018wireless} and energy consumption~\cite{9127160}. These costs can render distributed algorithms ineffective in bandwidth or battery-limited wireless networks, where latency and energy consumption are critical factors.

Recent works have explored the co-design of optimization problems and communication networks, particularly in the context of computational offloading \cite{fan2018application, 9264742, 9210812}. These works have addressed task offloading, resource allocation optimization, and joint learning of wireless resource allocation and user selection. In contrast to existing literature, our approach differs by proactively designing stopping criteria to optimize tradeoffs rather than treating them as hyper-parameters set through cross-validations. This distinction makes our approach original and distinct from current state-of-the-art algorithms.

In our preliminary works, we have characterized the overall communication-computation of solving a distributed gradient descent problem where the workers had background traffic and followed a channel from medium access control~(MAC) protocols using random access, such as slotted-ALOHA~\cite{mahmoudi2020cost} or CSMA/CA~\cite{ mahmoudi20spawc} in the uplink. 
Going beyond such papers, to achieve a cost-aware training workflow, we need to consider the diminishing return rule of the optimization algorithms, which reveals that as the number of iterations increases, the improvement in training accuracy decreases. Then, we need to balance iteration cost and achievable accuracy before the algorithm's design phase. This paper constitutes a major step in addressing this important research gap. Previously in~\cite{mahmoudi2020cost, mahmoudi20spawc}, we proposed a cost-efficient framework considering the cost of each iteration of gradient descent algorithms along with minimizing a convex loss function. However, the theory of these papers was only limited to convex loss functions, the iteration costs did not consider the FedAvg algorithm and the computation latency, and there was no adequate study between the achievable test accuracy and the iteration costs. Hence, this paper proposes a new and original study compared to our preliminary works by
\begin{enumerate}
    \item Considering FedAvg algorithm;
    \item Assuming both convex and non-convex loss functions;
    \item Developing a novel theoretical framework for FedAvg that includes the communication-computation costs.  
\end{enumerate}
We apply the proposed framework to several wireless communication protocols and other communication-efficient algorithms for which we show original training and testing results. 
\subsection{Contributions}\label{subsec: contributions}
We investigate the trade-off between achievable FL loss and the overall communication-computation cost of running the FL over wireless networks as an optimization problem. {\color{black}This work focuses on training a cost-efficient FedAvg algorithm in a ``causal way'', meaning that our approach does not require the future information of the training to decide how much total cost, e.g., computation, latency, or communication energy, the training algorithm needs to spend before terminating the iterations. Different than our approach, most papers in the literature aim to train the FedAvg algorithm in resource-constraint conditions and propose the best resource allocation policies ``before'' performing the training~\cite{9264742,9127160}. These approaches rely mainly on approximating the future training information by using some lower and upper bounds of that information. In this work, we propose to train the FedAvg algorithm in a causal, communication, and computation efficient way. To this end, we utilize the well-known multi-objective optimization approach according to the scalarization procedure in~\cite{boydcnvx}. Therefore, we propose FedCau to improve the FedAvg algorithm by training in a cost-efficient manner without any need to know the future training information or any upper and lower bounds on them.
To the best of our knowledge, this is the first work that considers such causal approaches to train the FedAvg algorithm in a communication and computation efficient manner.} The main contributions of this work are summarized as: 
\begin{itemize}
    \item We propose a new multi-objective cost-efficient optimization that trades off model performance and communication costs for an FL training problem over wireless networks;
 
    \item We develop three novel causal solution algorithms, named FedCau, for the multi-objective optimization above, one with a focus on original FL and the others with a focus on stochastic FL. We establish the convergence of these algorithms for FL training problems using both convex and non-convex loss functions; 
    
    \item We investigate the training and test performance of the proposed algorithms using MNIST {\color{black} and CIFAR-10} datasets, over the communication protocols: slotted-ALOHA, CSMA/CA, and OFDMA. We consider these protocols because they are the dominant communication protocols in most wireless local area networks, such as IEEE~802.11-based products~\cite{ziouva2002csma}, or fixed assignment access protocol like OFDMA~\cite{8422767};
    
    \item  We apply our proposed FedCau on top of top-$q$ sparsification and lazily aggregated quantized gradient~(LAQ) methods showing vast applications of the FedCau~\cite{9238427, 8889996}; 
    
    \item  The experimental results highlight the ability of our proposed FedCau to achieve efficient and accurate training. We conclude that a co-design of distributed optimization algorithms and communication protocols is essential for the success of cost-efficient FL over wireless networks, including its applications to edge computing and IoT. 
    

\end{itemize}

The rest of this paper is organized as follows. Section~\ref{section: System_Model} describes the general system model and problem formulation. In Section~\ref{section: results}, we derive some useful results and propose our non-causal and causal FL algorithms~(FedCau), which are by design intended to run over communication networks. In the analysis, we consider both convex and non-convex loss functions.  In Section~\ref{section: Protocols}, we apply our algorithms to slotted-ALOHA, CSMA/CA, and OFDMA. In Section~\ref{sec: numerical_results}, we analyze the performance of the FedCau algorithms. We then conclude the paper in Section~\ref{section: Conclusion}. We moved all the proofs and extra materials to the Appendix. 

\emph{Notation:} Normal font $w$, bold font small-case $\bw$, bold-font capital letter $\bW$, and calligraphic font $\calW$ denote scalar, vector, matrix, and set, respectively. We define the index set $[N] = \{1,2,\ldots,N\}$ for any integer $N$. We denote by $\|\cdot\|$ the $l_2$-norm, by $|\calA|$ the cardinality of set $\calA$, by $[\bw]_i$ the entry $i$ of vector $\bw$, by $\bw{\tran}$ the transpose of $\bw$, and $\mathds{1}_{x}$ is an indicator function taking $1$ if and only if $x$ is true, and takes $0$ otherwise. 


\section{System Model and Problem Formulation}\label{section: System_Model}
In this section, we represent the system model and the problem formulation. First, we discuss the FedAvg algorithm, and afterward, we propose the main approach of this paper. 
\subsection{Federated Learning}\label{subsec: FLs}
Consider a star network of $M$ workers that cooperatively solve a distributed training problem involving a loss function $f(\bw)$. Consider $D$ as the whole dataset distributed among each worker $j \in [M]$ with $D_j$ data samples. Let tuple $(\bx_{ij}, y_{ij})$ denote data sample $i$ of $|D_j|$ samples of worker $j$ and $\bw \in \R^d$ denote the model parameter at the master node. Considering $\sum_{j =1}^{ M} |D_j| = |D|$, and $j, j' \in [M]$, $j \neq j'$, we assume $D_j \cap D_{j'}=\emptyset$, and defining $\rho_j := {|D_j|}/{|D|}$, we formulate the following training problem
\begin{equation}\label{eq: our-optimization}
\bw^* \in \argmin_{\bw\in\R^d} f(\bw)=\sum_{j =1}^{ M} 
{\rho_j f_j(\bw)},
\end{equation}
where $f_j(\bw) := \sum_{i =1}^{|D_j|} {f(\bw; \bx_{ij}, y_{ij})}/{|D_j|}$. Optimization problem~\eqref{eq: our-optimization} applies to a large group of functions as convex and non-convex (such as deep neural networks). 

The standard iterative procedure to solve problem~\eqref{eq: our-optimization} with the initial vector $\bw_0$ is
\begin{equation}\label{eq: iterative}
    \bw_{k} = \sum_{j=1}^M \rho_j \bw_k^j,~\hspace{2mm} k=1,\ldots, K.
\end{equation}
For a differentiable loss function~$f(\bw)$, we choose to perform~\eqref{eq: iterative} by the Federated Averaging~(FedAvg) algorithm.

Initializing the training process with~$\bw_0$, Federated Averaging~(FedAvg) is a distributed learning algorithm in which the master node sends $\bw_{k-1}$ to the workers at the beginning of each iteration~$k \geq 1$. Every worker $j\in[M]$ performs a number $E$ of local iterations, $i=1, \ldots,E$, of stochastic gradient descent~\cite{8889996} with data subset of $\xi_k^j \le |D_j|$, and computes its local parameter~$\bw_{i,k}^j$, considering the initial point of $\bw_{0,k}^{j} = \bw_{k-1}$,~\cite{li2019convergence}, for any~$k=1,\ldots,K$,
\begin{equation}\label{eq: local trains}
  \bw_{i,k}^{j}=\bw_{i-1,k}^{j}-\frac{\alpha_k}{\xi_k^j} \sum_{n=1}^{\xi_k^j} \nabla_{\bw} f(\bw_{i-1,k}^{j}; \bx_{nj}, y_{nj}), 
 \end{equation}
 where $\bw_k^j = \bw_{E,k}^{j}$. Then each worker transmits $\bw_{k}^{j}$ to the master node for updating $\bw_k$ according to~\eqref{eq: iterative}.  
Note that in FedAvg, when $E = 1$ and we use the exact gradient vector in the place of the stochastic gradient, we achieve the basic FL algorithm. 
Considering the FedAvg solver~\eqref{eq: local trains} for the updating process in~\eqref{eq: iterative}, and without enforcing convexity for $f(\bw)$, we use the following Remark throughout the paper.
\begin{remark}{\label{remark: descent property}}
[Theorem 11.7 of~\cite{DBLP:books/daglib/0022091}] Consider any differentiable loss function~$f(\bw): \R^d \mapsto \R$ with Lipschitz continuous gradient $\nabla_{\bw}f(\bw)$, i.e., $\|\nabla_{\bw}f(\bw_1)~-~\nabla_{\bw}f(\bw_2)\|~\le~L\|\bw_1~-~\bw_2\|$, for some constant $0 < L < \infty$, and let $\bw_1, \ldots, \bw_{k}$ be the sequence obtained from the FL algorithm updates in Eq.~\eqref{eq: iterative}. Then, by $\alpha_k \geq \gamma \|\nabla_w f(\bw_k)\|^2$ and for an appropriate constant $\gamma > 0$, the following inequality holds: $f(\bw_1)\geq \ldots \geq f(\bw_{k})$.
\end{remark} 
The workers use the FedAvg algorithm~\eqref{eq: local trains} to compute their local parameters~$\bw_{k}^{j}$, while the master node performs the iterations of~\eqref{eq: iterative} until a convergence criteria for $\left\| f(\bw_k)-f(\bw^*)\right\|$ is met~\cite{bertsekas1989parallel}. We denote by $K$ the first iteration at which the stopping criteria of the FedAvg algorithm is met, namely
\begin{equation}\label{eq: k-bar}
K := {\textrm{the first value of}}~k \mid \left\| f(\bw_k)-f(\bw^*)\right\| < \epsilon \:,
\end{equation}
where $\epsilon > 0$ is the decision threshold for terminating the algorithm at iteration $K$ and~$f(\bw^*)$ is the optimum of the loss function at the optimal parameter~ $\bw^*$. The state-of-the-art literature defines the threshold $\epsilon$ independently before training. However, an optimal threshold must be designed to optimize communication-computation resources in solving~\eqref{eq: our-optimization}. Since knowing $f(\bw^*)$ beforehand is not realistic, we propose an alternative approach to find $K$ in \eqref{eq: k-bar} without this prior knowledge. Our main contribution is determining $K$ as a function of the communication-computation cost and the loss function of the FedAvg algorithm~\eqref{eq: local trains}. We will substantiate this significant result in Section~\ref{subsec: problem formulation}. 

Let $c_k > 0$, ${k=1,2,\ldots}$ denote the cost of performing a complete communication iteration $k$. Accordingly, when we run FedAvg, namely an execution of~\eqref{eq: iterative} and~\eqref{eq: local trains}, the complete training process will cost $\sum_{k=1}^{K} c_k$. Some examples of $c_k$ in real-world applications are:
\begin{itemize}
    \item \emph{Communication cost}: $c_k$ is the number of bits transmitted in every communication iteration $k$;
    \item \emph{Energy consumption}: $c_k$ is the energy needed for performing a global iteration to receive $\bw_k$ at a worker and send $\{\bw_k^j\}_{j\in [M]}$ to the master node over the wireless channel; 
    \item \emph{Latency}: $c_k$ is the overall delay to compute and send parameters from and to the workers and the master node over the wireless channel~\cite{9264742}.
\end{itemize}
Considering latency as the iteration cost, the term $c_k$ for running every training iteration of the FedAvg algorithm~\eqref{eq: local trains} is generally given by the sum of four latency components:
\begin{enumerate}
\item $\ell_{1,k}$: communication latency in broadcasting parameters by master node;
\item $\ell_{2,k}$: the computation latency in computing $\bw_k^j$ for every worker $j$;
\item $\ell_{3,k}$: communication latency in sending $\bw_k^j$ to master node;
\item $\ell_{4,k}$: computation latency in updating parameters at the master node.
\end{enumerate}
See Section~\ref{section: Protocols} for more detailed modeling of the components of $c_k$ for slotted-ALOHA, CSMA/CA, and OFDMA protocols. 

\subsection{Problem Formulation}\label{subsec: problem formulation}
To solve optimization problem~\eqref{eq: our-optimization} over a wireless network, the FedAvg algorithm~\eqref{eq: local trains} faces two major challenges: 
\begin{enumerate}
    \item \emph{Computation-communication cost}: It lacks the incorporation of computation and communication costs related to local parameters and model updates. These costs depend on factors such as computation power, communication protocols, energy consumption, and overall communication resources of the local device;
    
    \item \emph{{Number of iterations}: The termination iteration~$K$ in~\eqref{eq: k-bar} significantly impacts the communication-computation cost of the algorithm~\eqref{eq: local trains}. A lower~$K$ would consume fewer resources while leading to a negligible degradation in training performance, compared to a higher~$K$ that can result in substantial communication-computation costs without significant improvements in training optimality.} 
\end{enumerate}
The termination iteration~$K$ in~\eqref{eq: k-bar} strongly impacts the overall training costs for solving the optimization problem~\eqref{eq: our-optimization}. Thus, selecting an appropriate value for $K$ is crucial to prevent potentially adverse effects on communication-computation resource utilization in FedAvg~\eqref{eq: local trains} over wireless networks. 

We propose an original optimization of the termination iteration~$K$ in the FedAvg algorithm~\eqref{eq: local trains} to tackle the mentioned challenges. By explicitly considering the cost of training iterations, we aim at obtaining an optimal stopping iteration that solves the following optimization problem.
\begin{subequations}\label{eq: general}
\begin{alignat}{3}
\label{general1}
  \underset{K}{\mathrm{minimize}} & \quad \left[f(\bw_{K}), \sum_{k=1}^{K} c_k\right] \: \\ 
  \text{subject to}& 
 \label{general2}
 \quad \bw_{k} = \sum_{j =1}^{ M}  \rho_j \bw_k^j,\quad k=1,\ldots, K \: \\
 \label{general3}
& \quad \bw_{0,k}^{j} = \bw_{k-1}, \quad k=1,\ldots, K \: \\
\nonumber
&\quad\bw_{i,k}^{j}=\bw_{i-1,k}^{j}-
\: \\
\nonumber
&\quad \frac{\alpha_k}{\xi_k^j} \sum_{n=1}^{\xi_k^j} \nabla_{\bw} f(\bw_{i-1,k}^{j}; \bx_{nj}, y_{nj}),~k\le K,
\end{alignat}
\end{subequations}
where $\sum_{k=1}^{K} c_k$ quantifies the overall iteration-cost expenditure for the training of loss function~$f(\bw)$ when transmitting in a particular wireless channel in uplink.  Note that~\eqref{general1} represents a multi-objective function, which aims at minimizing the training loss function $f(\bw)$, and the overall iteration cost $\sum_{k=1}^{K} c_k$. Note that the values of $c_k$, for $k \le {K}$, can be, in general, a function of the parameter $\bw_k$, but neither $c_k$ nor $\bw_k$ are optimization variables of problem~\eqref{general1}. Optimization problem~\eqref{eq: general} states to devote communication-computation resources as efficiently as possible while performing FedAvg algorithms~\eqref{eq: local trains} to achieve an accurate training result for loss function~$f(\bw)$. Thus, by solving optimization problem~\eqref{general1}, we can obtain the optimal number of iterations for FedAvg algorithm~\eqref{eq: local trains}, which minimizes the communication-computation costs while also minimizing the loss function of FedAvg. 

\begin{remark}\label{re: causal and non-causal}
We have formulated optimization problem~\eqref{eq: general} according to the ``unfolding method" of iterative algorithms~\cite{balatsoukas2019deep}, where it is ideally assumed that the optimizer knows beforehand (before iterations~\eqref{eq: iterative} and~\eqref{eq: local trains} occur) what the cost of each communication iteration in~\eqref{eq: iterative} would be and when~they would terminate. Such an ideal formulation cannot occur in the real world since it assumes knowledge of the future, thus being called {\bf ``non-causal setting"}. However, this formulation is useful because its solution gives the best optimal value of the stopping iteration~$k^*$. In this paper, we show that we can convert such a non-causal solution of problem~\eqref{eq: general} into a practical algorithm in a so-called {\bf``causal setting"}. We will show that the solution to the causal setting given by the practical algorithm is very close to $k^*$. 
\end{remark}

Solving~\eqref{eq: general} presents several challenges: it is multi-objective, involves integer variables, and contains non-analytical objective and constraint functions with non-explicit dependencies on $K$. Additionally, the problem is non-causal, making it difficult to determine the optimal $K$ without knowing $\bw_{k}$'s in advance. Thus, addressing such non-explicit and non-causal optimization problems can be highly challenging~\cite{boydcnvx}. In the next section, we propose a practical solution to problem \eqref{eq: general}. 

\section{Solution Algorithms} \label{section: results}
In this section, we present preliminary technical results, propose an iterative solution to~\eqref{eq: general}, and demonstrate that the proposed methods achieve optimal or sub-optimal solutions while converging in a finite number of iterations. 

\subsection{Preliminary Solution Steps}
In this subsection, we develop some preliminary results to arrive at a solution to the optimization problem~\eqref{eq: general}. We start by transforming~\eqref{eq: general} according to the scalarization procedure of multi-objective optimization~\cite{boydcnvx}.
Specifically, we define the joint communication-computation cost and the loss function of FedAvg algorithm~\eqref{eq: local trains} as a scalarization of the overall iteration-cost function $\sum_{k=1}^{K} c_k$ and the loss function $f(\bw_{K})$. Note that such a joint cost is general in the sense that, depending on the values of $c_k$, it can naturally model many communication-computation costs, including constant charge per computation and mission-critical applications.

We transform the multi-objective optimization problem~\eqref{eq: general} into its scalarized version as 
\begin{subequations}\label{eq: cost-efficient-distributed-optim-3}
\begin{alignat}{3}
\label{general13}
 k^*&\quad\in \quad \underset{K}{\argmin}\quad G(K) \: \\ 
 \text{subject to}& 
 \label{general22}
 \quad \bw_{k} = \sum_{j =1}^{ M}  \rho_j \bw_k^j,\quad k\le K \: \\
 \label{general33}
& \quad \bw_{0,k}^{j} = \bw_{k-1}, \quad k\le K \: \\
&\quad\bw_{i,k}^{j}=\bw_{i-1,k}^{j}- 
\: \\
\nonumber
&\quad \frac{\alpha_k}{\xi_k^j} \sum_{n=1}^{\xi_k^j} \nabla_{\bw} f(\bw_{i-1,k}^{j}; \bx_{nj}, y_{nj}),~k\le K,
 \end{alignat}
 \end{subequations}
where $G(K)$ and $C(K)$ are defined as
 \begin{equation}\label{eq: G_define}
     G(K) := \left(\beta C(K)+(1-\beta)f(\bw_{K})\right),
 \end{equation}
\begin{equation}\label{eq: C(K)}
    \vspace{-0.005\textheight}{C(K):= \sum_{k=1}^{K}\hspace{1.1mm}c_k.} 
\end{equation}
$C(K)$ is the iteration-cost function representing all the costs the network spends from the beginning of the training until the termination iteration~$K$ and $\beta \in (0,1)$ is the scalarization factor of the multi-objective scalarization method~\cite{boydcnvx}.

The following lemma states that if $G(K)$ is monotonically decreasing, we can find $k^*$ where $G(K)$ is minimized.
\begin{lemma}\label{lemma: G}
Consider optimization problem~\eqref{eq: cost-efficient-distributed-optim-3}. Let $G(K)$ be a non-increasing function of all $K \le k^*$. Then, $k^*$ indicates the index at which the sign of discrete derivation~\cite{weigand2014discrete} of $G(K)$ changes for the first time, i.e. 
\begin{equation}\label{eq: finding k}
    k^* \in \min \{ K | G(K+1)-G(K) > 0 \}
\end{equation}
\end{lemma}
\begin{IEEEproof}
See Appendix~A-A in~\cite{mahmoudi2022fedcau}.
\end{IEEEproof}
In the following section, we present three algorithms to solve optimization problem~\eqref{eq: cost-efficient-distributed-optim-3}. First, we discuss the non-causal setting for characterizing the minimizer, then, introduce a causal setting to design algorithms that achieve practical minimizers for convex and non-convex loss functions. Finally, we establish the optimality and convergence of the algorithms. 

\subsection{Non-causal Setting}\label{subsec: non-causal}
An ideal approach to solve problem~\eqref{eq: cost-efficient-distributed-optim-3} is an exhaustive search over the discrete set of $K \in [0,+\infty)$. However, this approach requires knowing in advance the sequences $(f(\bw_k)){k}$ and $(c_k){k}$ for all $k \in [0,+\infty)$, which is not practical as the sequence of parameters $(\bw_k)_k$, and consequently $(f(\bw_k))_k$, are not available in advance. For analytical purposes, our non-causal setting assumes that all these values are available at $k=0$, enabling us to find the ultimate minimizer $k^*$. While this approach is not feasible in practice, we investigate it to establish a benchmark for the performance evaluation of subsequent causal solution algorithms (see Section~\ref{sec: numerical_results}).

\subsection{FedCau for Convex Loss Functions} \label{Causal Setting}
Here, we propose an approximation of the optimal stopping iteration $k^*$, referred to as $k_c$. Our analysis demonstrates that $k_c$ can be practically computed using a causal setting scenario. Under certain conditions, we establish that $k_c$ corresponds to $k^*$ or $k^*+1$. Specifically, when $k^* = K^{\max}$, with $K^{\max}$ denoting the maximum allowable number of iterations, we have $k_c = k^*$, otherwise, $k_c = k^*+1$ (see Section~\ref{subsec: convergence}).
\begin{algorithm}[t]
    \centering
    \caption{Cost-efficient batch FedCau.}
\scriptsize
\label{alg: synchronous alg}
\begin{algorithmic}[1]
\State \textbf{Inputs:} $\bw_0$, ${(\bx_{ij}, y_{ij})}_{i,j}$, $\alpha_k$, $M$, $\{|D_j|\}_{j\in[M]}$, $\rho_j$.
\vspace{0.003\textheight}
\State \textbf{Initialize:} $k_c = +\infty$, $G(0)=+\infty$
\vspace{0.0025\textheight}
\State Master node broadcasts $\bw_0$ to all nodes
\vspace{0.0025\textheight}
\For{$k \leq k_c$} \Comment{{\color{black}Global iterations}}
\vspace{0.0025\textheight}
\For{$j \in [M]$} 
\State Calculate $f_k^j := \sum_{i =1}^{|D_j|} f(\bw_k; \bx_{ij}, y_{ij})/|D_j|$ 
\vspace{0.0025\textheight}
\State Set $\bw_{0,k+1}^j=\bw_{k} $ 
\vspace{0.0035\textheight}
\For{$h \in [E]$} \Comment{{\color{black}local iterations}}
\vspace{0.0025\textheight}
\State Compute $\bw_{h,k+1}^j \leftarrow \bw_{h-1,k+1} - \alpha_k \nabla_w f_k^j$
\EndFor
\vspace{0.0025\textheight}
\State Set $\bw_{k+1}^j=\bw_{E,k+1}^j $ 
\vspace{0.0035\textheight}
\State Send $\bw_{k+1}^j$ and $f_k^j$ to the master node
\EndFor
\vspace{0.0025\textheight}
\State Wait until master node collects all $\{\bw_{k+1}^j\}_j$ and set $\bw_{k+1} \leftarrow \sum_{j =1}^{ M}  \rho_j \bw_{k+1}^j$
\vspace{0.0025\textheight}
\State Calculate $f(\bw_{k}) :=\sum_{j =1}^{ M} \rho_j f_k^j$
\vspace{0.0025\textheight}
\State Calculate $c_k$ and $G(k)$
\vspace{0.0025\textheight}
\If{$G(k) < G(k-1)$} \Comment{{\color{black}Evaluating~\eqref{eq: finding k}}}
\vspace{0.0025\textheight}
\State Master node broadcasts $\bw_{k+1}$ to the workers
\Else
\State Set $k_c = k$, Break and go to line 24

\EndIf
\State Set $k \leftarrow k+1 $
\EndFor
\State \textbf{Return} $\bw_{k_c}$, $k_c$, $G(k_c)$
\end{algorithmic}
\end{algorithm}
Thus, we develop three implementation variations of FedAvg algorithm~\eqref{eq: local trains}, Algorithms~\ref{alg: synchronous alg}-\ref{alg: Nonconvex alg}, with our causal termination approach, FedCau, for solving~\eqref{eq: cost-efficient-distributed-optim-3}. Algorithms~\ref{alg: synchronous alg} and \ref{alg: asynchronous alg} are batch and mini-batch implementations using convex loss functions, while Algorithm~\ref{alg: Nonconvex alg} considers non-convex loss functions.

In the batch update of Algorithm~\ref{alg: synchronous alg}, workers compute $\{\bw_k^j ,f_k^j\}_{j\in[M]}$ and transmit them to the master node (see lines~6-12). We assume that the local parameter of each worker consists of the value of local FL model $\bw_k^j$ and the local loss function $f_k^j$~\footnote{We assumed that $f_k^j \in \R$ and $\bw_k^j \in \R^d$, then the communication overhead, in term of the number of bits, for transmission of $f_k^j$ is negligible compared to the local FL model $\bw_k^j$. Thus, we consider the local parameters to consist of both the local FL model and the local loss function value.}. The master node updates $\bw_k$ and $f(\bw_k)$ upon receiving all local parameters $\{\bw_k^j ,f_k^j\}_{j\in[M]}$ from workers at each iteration~$k$ (see lines~14-15). Then, the iteration cost $c_k$, representing the iteration cost, is calculated. To prevent termination in the first iteration, we initialize $G(0)=+\infty$, and subsequently, the multi-objective cost function $G(k)$ is updated (line~16). A comparison between $G(k)$ and its previous value $G(k-1)$ is made (see line~17) to determine the termination of iterations (see lines~19-24).

In FedAvg, there are many scenarios where specific workers can upload their local parameters to the master node, resulting in implicit sub-sampling and approximations of $f(\bw_k)$ denoted as $\hat{f}(\bw_k)$. This sub-sampling results in approximating the joint communication-computation and FL cost function, $\hat{G}(K)$. Algorithm~\ref{alg: asynchronous alg} employs mini-batch updates to avoid excessive resource consumption for marginal test accuracy improvements. It leverages the descent property of FedAvg algorithm~\eqref{eq: local trains} for a monotonic decreasing loss function~$f(\bw)$, as described in Remark~\ref{remark: descent property}. Algorithm~\ref{alg: asynchronous alg} aims at achieving non-increasing sequences of $f(\bw_k)_k$ and $G(k)_{k \le k^*}$.

{\color{black}
\begin{algorithm}[t]
\caption{{\color{black}Stochastic cost-efficient mini-batch FedCau.}}
\scriptsize
\label{alg: asynchronous alg}
\begin{algorithmic}[1]
\State \textbf{Inputs:}  $\bw_0$, ${(\bx_{ij}, y_{ij})}_{i,j}$, $\rho_j$, $F_f$, $\alpha_k$, $M$, $\left\{|D_j|\right\}_{j\in[M]}$
\State \textbf{Initialize:} $k_c = +\infty$, $T_s = +\infty$, $j_s = 0$, $\hat{G}(0)=+\infty$, $\calM_n^1 = \left\{[M]\right\}$, and Fair-Count $=\mathbf{1}_{M\times 1}$, $t_1^s = 1$

\State Master node broadcast $\bw_0$ to all nodes

\For{$k \leq k_c$} \Comment{{\color{black}Global iterations}}
\vspace{0.0025\textheight}
\State $\calM_k^a = \{ \}$
\vspace{0.0035\textheight}
\For{$j \in [M]$}  
\vspace{0.0025\textheight}
\State Calculate $f_k^j := \sum_{i =1}^{|D_j|} f(\bw_k; \bx_{ij}, y_{ij})/|D_j|$ 
\vspace{0.0025\textheight}
\State Set $\bw_{0,k+1}^j=\bw_{k} $ 
\vspace{0.0025\textheight}
\For{$h \in [E]$} \Comment{{\color{black}local iterations}}
\State Compute $\bw_{h,k+1}^j \leftarrow \bw_{h-1,k+1} - \alpha_k \nabla_w f_k^j$
\EndFor
\vspace{0.0025\textheight}
\State Set $\bw_{k+1}^j=\bw_{E,k+1}^j $ 
\vspace{0.0035\textheight}
\If{$j \in \calM_k^n$} \Comment{{\color{black}Partial participation}}
\vspace{0.0025\textheight}
\State Send $ \bw_{k+1}^j$ and $f_k^j$ to the master node
\EndIf
\EndFor

\If{$k=1$} master node:
 \vspace{0.0025\textheight}
\For{$t_1^s \leq T_s$} until $\calM_k^n= \{ \}$ \Comment{{\color{black}Computing $T_s$}}
\vspace{0.5mm}
\vspace{0.0025\textheight}
\State $\calM_k^a \leftarrow \calM_k^a \cup \{j_s\}$
\vspace{0.5mm}
\vspace{0.0025\textheight}
\State  $\calM_k^n\leftarrow \calM_k^n\setminus \calM_k^a$ \Comment{{\color{black}Worker participation}}
\vspace{0.0025\textheight}
\EndFor 
\vspace{0.0025\textheight}
\State Set $T_s = t_1^s$
\vspace{0.0025\textheight}
\State Set $\bw_{k+1} \leftarrow \sum_{j =1}^{ M}  \rho_j \bw_{k+1}^j$ 
\vspace{0.0025\textheight}
\Comment{{\color{black}Global update}}
\State Calculate $f(\bw_{k}) :=\sum_{j =1}^{ M} \rho_j f_k^j$ and $G(k)$
\vspace{0.5mm}
\vspace{0.0025\textheight}
\State Set a time budget $T\leq T_s$, and $t_k^s = 0$ 
\vspace{0.0025\textheight}
\State Set $\calM_{k+1}^n= \left\{[M]\right\}$
\Else \Comment{{\color{black}if $k\ge 2$}} 

\For{$t_k^s \leq T$}  \Comment{{\color{black}Assigning time budget~$T$}}
\vspace{0.0025\textheight}
\State Every node $j\in \calM_k^n$ send $\bw_{k+1}^j$
\vspace{0.0025\textheight}
\If{Successful node $j_s \in \calM_n$}
\vspace{0.003\textheight}
\State $\calM_k^a \leftarrow \calM_k^a \cup \{j_s\}$
\vspace{0.0035\textheight}
\State ${\text{Fair-Count}}[j_s] = {\text{Fair-Count}}[j_s] +1$ 
\EndIf
\EndFor

\State Master node set \Comment{{\color{black}Global update with replacements}}
\vspace{0.0025\textheight}

$\bw_{k+1} \leftarrow \sum_{j\in \calM_k^a} \rho_j \bw_{k+1}^j + \sum_{j'\notin \calM_k^a} \rho_{j'} \bw_{k}^{j'}$
\vspace{1mm}
\vspace{0.0025\textheight}
\State Master node calculate 

\vspace{1mm}

$\hat{f}(\bw_{k}) :=\sum_{j \in \calM_k^a} \rho_j f_k^j + \sum_{{j'} \notin \calM_k^a} \rho_{j'} f_{k-1}^{j'}$ and $\hat{G}(k)$
\vspace{1mm}
\State Master node update 
\vspace{0.0025\textheight}

$\calM_k^n=\left\{j |{\text{Fair-Count}}[j] < F_f\right\} $ \Comment{{\color{black}Fairness evaluation}}
\vspace{1mm}
\vspace{0.0025\textheight}
\If{$\calM_k^n= \{\}$}
\vspace{0.0035\textheight}
\State $\calM_{k+1}^n= \left\{[M]\right\}$ 
\vspace{0.0025\textheight}
\Comment{{\color{black}Update partial participation}}
\vspace{0.0025\textheight}
\State Fair-Count $=\mathbf{0}_{M\times 1}$
\EndIf
\EndIf

\If{$\hat{G}(k) < \hat{G}(k-1)$} \Comment{{\color{black}Evaluating~\eqref{eq: finding k}}}
\vspace{0.0025\textheight}
\State Master node broadcast $\bw_{k+1}$ to the workers
\Else
\State Set $k_c = k$, and $\hat{\bw}_{k_c}\leftarrow \bw_{k+1}$
\vspace{0.0025\textheight}
\State Break and go to line {51} \Comment{{\color{black}Terminating the training}}
\EndIf
\State Set $k \leftarrow k+1 $, and $t_{k+1}^s = 1$ 
\EndFor
\State \textbf{Return} $\hat{\bw}_{k_c}$, $k_c$, $\hat{G}(k_c)$
\end{algorithmic}
\end{algorithm}
}
 

 Algorithm~\ref{alg: asynchronous alg} introduces partial worker participation and fairness in training FedCau. $\mathcal{M}_k^n$ represents the node selection subset at each communication iteration~$k$, and Fair-count$[j]$ denotes the counter for the number of successfully-sent local parameters by worker $j \in \mathcal{M}_k^n$. We introduce a ``Fairness-Factor'' $F_f \le K^{\max}$ that restricts workers from transmitting more than $F_f$ local parameters until all workers satisfy $F_f$ local parameter transmission. At the first communication iteration $k=1$, once a worker $j \in \mathcal{M}_1^n$ successfully transmits its local parameter $\boldsymbol{w}_1^j$, it is removed from the selected node subset $\mathcal{M}_1^n$ (see lines7-21). Thus, worker $j$ will not transmit any packets until all workers send their local parameters. The master node computes the resource used to perform the first communication iteration as $T_s$. It considers $T_s$ as a benchmark to determine $T \le T_s$ as the maximum allowable time slots for future iterations $k=2, \ldots, K$ (see line~25)~\footnote{Here, we allocate an equal portion of the resource to each iteration. However, one can assign a different portion of resources to each iteration, which is out of the scope of this paper.}.
 Note that in $k=1$, low-power workers have a higher probability of transmitting their local parameter, and the latency $T$ is smaller compared to full worker participation. After completing communication iteration $k=1$, partial worker participation begins at $k \geq 2$ when the master node updates $\calM_{k+1}^2$ (see line 26).
 
For $k \ge 2$, the selected workers $j \in \mathcal{M}_k^n$ have a time budget $T$ to compute and transmit their local parameters. This constraint creates competition among the selected workers to communicate with the master node. However, some workers may fail to send their local parameters. To address this challenge, we introduce the set $\mathcal{M}_k^a$, which contains the indexes of the successful workers $j_s \in \mathcal{M}_k^n$ that managed to transmit during iteration $k$ (see line 31). Additionally, the fairness counter of each successful worker, Fair-count$[j_s]$, is increased (see line 32) to influence future selections for communication iterations. Afterward, the master node updates the global parameter by the local parameters it has received, $\bw_k^j, j \in \calM_k^a$, and then replaces the missing local parameters by the values of the previous iteration, for the local parameters $\bw_k^{j'} = \bw_{k-1}^{j'}, j' \notin \calM_k^a$~\footnote{For simplicity, we use the notation $f_k^j:=f_j(\bw_k)$.} and local functions $f_k^{j'} = f_{k-1}^{j'},  j' \notin \calM_k^a$ (see lines~30-31). Algorithm~2 utilizes this replacement strategy to ensure the non-increasing behavior of $G(k), k=1,\ldots, k^*$, and maintain a descent sequence of $\hat{f}(\boldsymbol{w}_k), k=1, \ldots, k^*$. Since Algorithm~2 considers convex loss functions, the replacement of missing parameters guarantees the descent behavior of the sequence $\hat{f}(\boldsymbol{w}_k), k=1, \ldots, k^*$ (Lemma~\ref{lemma: replacing}). Additionally, the master node updates the selected workers based on the fairness factor, ensuring fair worker participation for the upcoming communication iterations (see lines 37-40). This process requires the master node to retain a memory of all previous local parameters. The remaining part of Algorithm~\ref{alg: asynchronous alg} (lines 43-51) handles parameter updates and checks for the potential stopping iteration $k_c$, similar to lines 12-20 in Algorithm~\ref{alg: synchronous alg}.
 \begin{lemma}\label{lemma: replacing}
 Let $f_k^j$ be the local loss function at the communication iteration $k$ for each worker $j \in [M]$. Suppose that $f_j(\bw)$ be a convex function w.r.t. $\bw$. Then, Algorithm~\ref{alg: asynchronous alg} guarantees the decreasing behavior of $\hat{f}(\bw_k), \forall k$. 
 \end{lemma}
 \begin{IEEEproof}
 See Appendix~A-B in~\cite{mahmoudi2022fedcau}.
 \end{IEEEproof}


As explained above, Algorithm 2 allows for both full and partial participation, offering fairness in worker participation based on the parameter $T$. The distinction between full and partial participation lies in the fact that in partial participation, the update of the global parameter $\boldsymbol{w}_k$ depends on the new local parameters from the subset $\mathcal{M}_k^n$. However, it remains uncertain which workers within the subset successfully transmit their local parameters and which ones fail, particularly when workers possess non-iid training data. To address this challenge, we introduce the fairness-factor $F_f$ to mitigate the impact on the global update. The value of $F_f$ can be tailored to the specific training application, enabling customization of the partial participation scheme.

\begin{algorithm}[t]
\caption{Stochastic non-convex cost-efficient mini-batch FedCau.}
\scriptsize
\label{alg: Nonconvex alg}
\begin{algorithmic}[1]
\State \textbf{Inputs:}  $\bw_0$, ${(\bx_{ij}, y_{ij})}_{i,j}$, $\alpha_k$, $M$, $\{|D_j|\}_{j\in[M]}$, $\rho_j$, $E$, $\xi_k^j $ 
\State \textbf{Initialize:} $k_c^u = k_c^l = 0$, $k_{\text{max}}^l = 2$
\vspace{0.002\textheight}
\State Master node broadcasts $\bw_0$ to all nodes

\For{$k\geq 1$} \Comment{{\color{black}Global iterations}}
\State $\calM_a = \{ \}$

\Statex Each node $j$ calculates: \Comment{{\color{black}local iterations}}

\For{$j \in [M]$}  
\If{$k=0$}
\State Randomly select a subset of data with size $\xi_k^j$
\State $F_0^j := \sum_{i =1}^{\xi_k^j} F(\bw_0; \bx_{ij}, y_{ij})/\xi_k^j$ 
\vspace{0.0025\textheight}
\EndIf
\State Set $\bw_{0,k+1}^j = \bw_{k}$, $F_{0,{k+1}}^j = F_k^j$ 
\vspace{0.0025\textheight}
\For{$i \in [E]$}
\State Randomly select a subset of data with size $\xi_k^j$
\State $\bw_{i,k+1}^j \leftarrow \bw_{i_{E-1},k+1} - \alpha_k \nabla_w F_{i_{E-1},{k+1}}^j$
\vspace{0.0025\textheight}
\State $F_{i,k+1}^j = \sum_{i =1}^{\xi_k^j} F(\bw_{i,k+1}^j; \bx_{ij}, y_{ij})/\xi_k^j$
\vspace{0.002\textheight}
\EndFor
\vspace{0.0015\textheight}
\State Set $\bw_{E,k+1}^j = \bw_{k+1}^{j}$, and $F_{k+1}^j = F_{E,k+1}^j$
\vspace{0.0025\textheight}
\State Send $\bw_{k+1}^j$ and $F_{k+1}^j$ to the master node
\EndFor

\Statex Master node calculates:
\Comment{{\color{black}Global update with replacement}} 
\vspace{0.0025\textheight}
\State $\calM_a \leftarrow \calM_a \cup \{\text{Successful workers}\}$
\State $\bw_{k+1} \leftarrow \sum_{j\in \calM_a} \rho_j \bw_{k+1}^j + \sum_{j'\notin \calM_a} \rho_{j'} \bw_{k}^{j'}$
\vspace{0.0025\textheight}
\State $\Tilde{F}(\bw_{k}) :=\sum_{j \in \calM_a} \rho_j F_k^j + \sum_{{j'} \notin \calM_a} \rho_{j'} F_{k-1}^{j'}$ 
\vspace{0.0015\textheight}
\State Update $C(k)$
\If{$k \leq 2$}
\vspace{0.0015\textheight}
\State Set $F_u(\bw_{k})=F_l(\bw_{k}) = \Tilde{F}(\bw_{k})$
\Else \Comment{{\color{black}Update $F_u(\bw_{k})$ and $F_l(\bw_{k})$}}
\If{$ \Tilde{F}(\bw_{k}) \geq \Tilde{F}(\bw_{k-1})$}
\vspace{0.0025\textheight}
\State Set $F_u(\bw_k) = \Tilde{F}(\bw_{k})$
\vspace{0.0025\textheight}
\If{$\Tilde{F}(\bw_{k}) \geq F_u(\bw_{k-1})$}
\vspace{0.0025\textheight}
\State $k_{\max}^u = \underset{ k_u < k}{\max} \{ k_u | F_u(\bw_{k_u})>\Tilde{F}(\bw_{k})\}$
\vspace{0.0025\textheight}
\State Update $(F_u(\bw_i))_{i = k_{\max}^u, \ldots, k}$ as~\eqref{eq: mapping}
\EndIf
\State Calculate $\delta_k^u = F_u(\bw_{k})-F_u(\bw_{k-1})$ 
\vspace{0.0025\textheight}
\State Set $F_l(\bw_{k})=F_l(\bw_{k-1})~+~\delta_k^u$
\vspace{0.0025\textheight}
\State Update $(G_u(K))_{K=k_{\max}^u,\ldots, k}$~and $G_l(k)$ 
\Else  \Comment{{\color{black} $\Tilde{F}(\bw_{k}) < \Tilde{F}(\bw_{k-1})$}}
\If{$\Tilde{F}(\bw_{k}) < F_l(\bw_{k_{\text{max}}^l})$}
\vspace{0.0025\textheight}
\State Set $F_l(\bw_{k}) = \Tilde{F}(\bw_{k})$
\vspace{0.0025\textheight}
\State Update $(F_l(\bw_i))_{i = k_{\max}^l, \ldots, k}$ as~\eqref{eq: mapping}
\vspace{0.0025\textheight}
\State Calculate~$\delta_k^l = F_l(\bw_{k})-F_l(\bw_{k-1})$
\vspace{0.0025\textheight}
\State Set $F_u(\bw_{k})~=~F_u(\bw_{k-1})~+~\delta_k^l$
\vspace{0.0025\textheight}
\State Update $(G_l(K))_{K=k_{\max}^l,\ldots, k}$~and $G_u(k)$
\vspace{0.0025\textheight}
\State Set $k_{\text{max}}^l = k$
\Else \Comment{{\color{black}If $\Tilde{F}(\bw_{k}) \geq F_l(\bw_{k_{\text{max}}^l})$}}
\vspace{0.0025\textheight}
\State Set $F_u(\bw_k) = \Tilde{F}(\bw_{k})$ 
\vspace{0.0025\textheight}
\State Calculate $\delta_k^u = F_u(\bw_{k})-F_u(\bw_{k-1})$ 
\vspace{0.0025\textheight}
\State Set $F_l(\bw_{k})=F_l(\bw_{k-1})+\delta_k^u$
\vspace{0.0025\textheight}
\State Update $G_u(k)$ and $G_l(k)$
\EndIf
\EndIf
\EndIf 

\Comment{{\color{black}Evaluate~\eqref{eq: finding k} for $G_l$ and $G_u$}}
\State $k_{c}^l ={\min}\{K| G_l( K)>G_l( K-1)\}$
\vspace{0.0025\textheight}
\State $k_{c}^u ={\min}\{K| G_u( K)>G_u( K-1)\}$

\vspace{0.0015\textheight}
\If{$k_{c}^u\neq 0, k_{c}^l\neq 0$}
\vspace{0.0015\textheight}
\State Break and go to line 60 \Comment{{\color{black}Terminating the training}}
\Else
\State Master node broadcast $\bw_{k+1}$ to the workers
\State Set $k \leftarrow k+1 $ 
\EndIf
\EndFor
\State \textbf{Return} $k_c^l$, $k_c^u$, $(\bw_k)_{k=k_c^l, \ldots, k_c^u}$
\end{algorithmic}
\end{algorithm}

Another challenge in the partial participation of Algorithm 2 is determining the appropriate time budget $T$ for each iteration. Algorithm 2 suggests selecting a value for $T < T_s$ by causally computing $T_s$ in the first iteration, considering full worker participation and excluding background traffic. However, the choice of $T$ depends on the specific learning application, such as healthcare, autonomous driving, or video surveillance. One should consider a suitable time budget of $T$ based on the requirements of the learning application. For latency-sensitive scenarios like autonomous driving, where quick decisions are crucial to prevent accidents, a smaller value of $T$ is preferred.
\subsection{FedCau for Non-convex Loss Functions}\label{subsec: non-convex f}
Here, we extend Algorithms~\ref{alg: synchronous alg} and~\ref{alg: asynchronous alg} to include non-convex loss functions. We consider FedAvg~\cite{konevcny2016federated}, where each worker~$j$ performs $E \geq 1$ local iterations over its local data subset, $\xi_k^j \le |D_j|$. The master node updates the global parameter $\bw_{k+1}$ by averaging and broadcasting the local parameters to the workers. Additionally, the master node calculates~$\Tilde{F}(\bw_k)$ by averaging the local loss functions of the workers~\cite{konevcny2016federated}.

We design a cost-efficient algorithm which optimizes~$\Tilde{G}(K)$, an estimate of the multi-objective cost function~${G}(K)$ defined as $\Tilde{G}(K) := \beta C(K) + (1-\beta) \Tilde{F}(\bw_{K})$, where recall that $C(K)$ is the iteration-cost function at $K$. We design such an estimate since the stochastic nature of the sequences of $(\Tilde{F}(\bw_k))_k$, arises from the local updates by $\xi_k^j \le |D_j|$ using mini-batches, results in a stochastic sequence of $(\Tilde{G}(k))_k$. This sequence $(\Tilde{G}(k))_k$ hinders the application of Algorithms~\ref{alg: synchronous alg}, \ref{alg: asynchronous alg} and might lead to their early stopping at a communication iteration. Thus, we need to develop an alternative algorithm. 

We propose a causal approach to establish non-increasing upper and lower bounds, $G_u(K)$ and $G_l(K)$, for the stochastic sequence $(\Tilde{G}(k))_k$. As this sequence is not necessarily non-increasing and may have multiple local optimum points, we aim at obtaining an interval $k_c^u \le k_c \le k_c^l$, where $k_c^u$ and $k_c^l$ represent the stopping iteration for $G_u(K)$ and $G_l(K)$ functions, respectively. According to the definition of $G(K)$ function, we define $G_u(K)$, and $G_l(K)$ functions as
\begin{subequations}\label{eq: G bounds}
\begin{alignat}{3}
\label{b: upperG}
    G_u(K) &:= \beta C(K)+(1-\beta)F_u(\bw_{K}) ,\\
\label{c: lowerG}
    G_l(K) &:= \beta C(K)+(1-\beta)F_l(\bw_{K})\:,
\end{alignat}
\end{subequations}
where $F_u(\bw_{K})$ and $F_l(\bw_{K})$ represent the estimation of the loss function at upper and lower bounds. 
To obtain the sequences of $(G_u(k)){k}$ and $(G_l(k)){k}$, the master node computes the upper and lower bounds for $\Tilde{F}(\bw_k)$ while ensuring the monotonic decreasing behavior of $(F_u(\bw_{k}))k$ and $(F_l(\bw{k}))_k$ to satisfy Remark~\ref{remark: descent property}. In the following, we now concentrate on the process of obtaining the bounds for $\Tilde{F}(\bw_k)$.

Algorithm~\ref{alg: Nonconvex alg} shows the steps required for the cost-efficient FedAvg with causal setting and non-convex loss function~$F(\bw)$. Lines~3-18 summarize the local and global iterations of FedAvg. Here, we introduce $\calM_a$ as the set of workers which successfully transmit their local parameters to the master node (see line~20). We initialize $F_u(\bw_k)= F_l(\bw_k) = \Tilde{F}(\bw_k), k \le 2$ for the first two iterations (see line~25). For iterations $k \geq 3$, if the new value of loss function fulfils $\Tilde{F}(\bw_k) \geq \Tilde{F}_(\bw_{k-1})$, the algorithm updates $F_u(\bw_k) = \Tilde{F}(\bw_k)$ (see line~28). Then, the algorithm checks if $\Tilde{F}(\bw_k)$, which is now equal to $F_u(\bw_k)$, is greater than the previous value of $F_u(\bw_{k-1})$ (see line~29). This checking is important because we must develop a monotonic decreasing sequence of $F_u(\bw_i)_{i=1:k_c^u}$. When $\Tilde{F}(\bw_k) \geq \Tilde{F_u}(\bw_{k-1})$, the master node returns to the history of $F_u(\bw_i)_{i=1:k-1}$ and checks for $i < k$, when the condition $F_u(\bw_i)> F_u(\bw_k)$ is satisfied. Since at each communication iteration $k$ we carefully check the monotonic behavior of $F_u(\bw_k)$, we are sure that if we find the proper maximum communication iteration $i$ that fulfills $i<k$, for which $F_u(\bw_{i+1}) <F_u(\bw_k)<F_u(\bw_i)$, we have the result of $F_u(\bw_j) < F_u(\bw_k), j<i$. Let us define this communication iteration $i$ as $k_{\text{max}}^u$ (see line 30). Thus, it is enough to find such $i$ to update the sequence of $F_u(\bw_{i_1}), i_1=i, \ldots, k$. 

 Now, we need to update the sequences of $F_u(\bw_{i_2}), i_2 = k_{\text{max}}^u, \ldots, k$ to obtain the monotonic decreasing upper bound. We choose the monotonic linear function because it satisfies the sufficient decrease condition~(see~\cite{DBLP:books/daglib/0022091}, Section~11.5). Therefore, we satisfy the decreasing behavior for $F_u(\bw_k)$ and the upper bound behavior, which means that  the maximum values of $\Tilde{F}(\bw_k)$ are always lower than $F_u(\bw_k)$. Thus, we update the sequences of $F_u(\bw_i), i = k_{\text{max}}^u, \ldots, k$ according to \eqref{eq: mapping}, with $k_1 = k_{\text{max}}^u$, $k_2 = k$, and $F_u(k_i) = F_{\text{linear}}(k_i), k_i \in [k_1,k_2]$.      
We define $F_{\text{Apxt}}(k)$ as the linear approximation of $F(\bw_k)$ in an interval $k \in [k_1,k_2]$
\begin{equation} \label{eq: mapping}
    F_{\text{Apxt}}(k_i) = a k_i + b, \hspace{2mm} k_1\le k_i\le k_2, 
\end{equation}
where
{
\begin{subequations}\label{eq: F approx}
\begin{alignat}{3}
\label{eq: F approx1}
    a &= \frac{\Tilde{F}(\bw_{k_2})-\Tilde{F}(\bw_{k_1})}{k_2-k_1},\\
\label{eq: F approx2}
    b &= F(\bw_{k_2}) - a k_2.
\end{alignat}
\end{subequations}}
Next, we need to update $F_l(\bw_k)$. Here, let us define the difference between two consecutive values of $F_u(\bw_k)$ and $F_u(\bw_{k-1})$ as $\delta_k^u$, and the difference between $F_l(\bw_k)$ and $F_l(\bw_{k-1})$ as $\delta_k^l$. Then, we update the corresponding values for $(G_u(K))_{K=k_{\max}^u,\ldots, k}$~and $G_l(k)$, respectively (see lines~34-35). Afterward, we need to check the condition at which $\Tilde{F}(\bw_k) < {F_l}(\bw_{k_{\text{max}}^l})$, where $k_{\text{max}}^l$ represents the last communication iteration at which the value of $\Tilde{F}(\bw_{k_{\text{max}}^l})$ has been considered as $F_l(\bw_{{k_{\text{max}}^l}})$. If $\Tilde{F}(\bw_k) < \Tilde{F}(\bw_{k_{\text{max}}^l})$, we need to update the lower bound sequences (see lines~37-38) to avoid over-decreasing the lower bound function $F_l(\bw_k)$ by the approximation of line 34. Subsequently, we need to calculate~$\delta_k^l = F_l(\bw_{k})-F_l(\bw_{k-1})$, then update $F_u(\bw_{k})~=~F_u(\bw_{k-1})~+~\delta_k^l$ and the value of $k_{\text{max}}^l = k$ (see lines~41-43). 

The last condition to check is when $ {F_l}(\bw_{k_{\text{max}}^l}) < \Tilde{F}(\bw_k) < \Tilde{F}(\bw_{k-1})$. In this condition, the monotonic decreasing behavior of $\Tilde{F}(\bw_k)$ is satisfied, whereas the decreasing behavior is not satisfied for the lower bound $F_l(\bw_k)$. Thus we set  ${F_u}(\bw_k) = \Tilde{F}(\bw_k)$, and $\delta_k^u = F_u(\bw_{k})-F_u(\bw_{k-1})$, $F_l(\bw_{k})=F_l(\bw_{k-1})+\delta_k^u$, and update $G_u(k)$ and $G_l(k)$ (see lines~45-48). Finally, lines 52-61 show when to stop the algorithm. 


\subsection{Optimality and Convergence Analysis}\label{subsec: convergence}
In this subsection, we investigate the existence and optimality of the solution to problem~\eqref{eq: cost-efficient-distributed-optim-3} and the convergence of the algorithms that return the optimal solutions. We are ready to give the following proposition, which provides us with the required analysis of Algorithms~\ref{alg: synchronous alg}, and~\ref{alg: asynchronous alg}.

First, we start with the monotonic behavior of $G(K), K \le k^*$. In practice, we have this desired monotonically decreasing behavior, as we show in the following proposition: 
\begin{prop}\label{prop: beta}
Consider $G(K)$ defined in Eq.~\eqref{eq: G_define}. Define $\Delta_k:=f_{k-1}-f_k$, $\Delta_0 = f_0$, 
by choosing $\beta$ as 
\vspace{-0.005\textheight}
\begin{equation}
    0 < \frac{1}{1+ \frac{{\rm max}_k \hspace{1mm} c_k}{\Delta_{0}}} \le \beta \le \frac{1}{1+ \frac{{\rm min}_k \hspace{1mm} c_k}{\Delta_{0}}} <1, \hspace{2mm} k\le k^*,
\end{equation}
\vspace{-0.005\textheight}
the function $G(K)$, $K \le k^*$, is non-increasing at $K$.
\end{prop}
\vspace{-0.001\textheight}
\begin{IEEEproof}
 See Appendix~A-C in~\cite{mahmoudi2022fedcau}.
\end{IEEEproof}
\vspace{-0.005\textheight}
\begin{remark}\label{remark: beta}
The previous proposition implies that, without loss of generality, we can assume that $\textrm{max}_k~c_k$ is high enough and $\textrm{min}_k~c_k$ is close to zero (setting the initial cost to zero, for example). Thus $\beta$ can, in practice, vary between $0$ and $1$, without restricting the applicability range of the multi-objective optimization.
\end{remark}
\begin{prop}\label{prop: convergence}
Optimization problem~\eqref{eq: cost-efficient-distributed-optim-3} has a finite optimal solution $k^*$.
\end{prop}
\begin{IEEEproof}
 See Appendix~A-D in~\cite{mahmoudi2022fedcau}.
\end{IEEEproof}
{\color{black} Proposition~\ref{prop: convergence} implies that when $G(K)$ is monotonically decreasing with $K$, $k^*$ is equal to~$K^{\max}$. According to the training setup, the maximum number of iterations is set as $K^{\max}$ at the beginning of the training. Thus, monotonically decreasing $G(K)$ results in $k^* = K^{\max}$, which means that the value of the FL loss function is dominant in $G(K)$, and the FedCau procedure is similar to the FedAvg method.}

The following Theorem clarifies an important relation between the non-causal and causal solutions of Algorithms~\ref{alg: synchronous alg}~and~2.  
\begin{theorem}\label{thrm: causal-alg}
Let $k^*$ be the solution to optimization problem~\eqref{eq: cost-efficient-distributed-optim-3}, and let $k_{c}$ denote the approximate solution obtained in the non-causal and causal settings of Algorithm~\ref{alg: synchronous alg}, and~\ref{alg: asynchronous alg}, respectively. Then, the following statements hold
\vspace{-0.005\textheight}
\begin{subequations}
\label{eq: causal k}
\begin{alignat}{3}
\label{a: causal-stopp}
  k_c  \in \{& k^*, k^* +1\}  \:, \\
\label{b: causal-stopp}
f(\bw_{k_c}) & \le f(\bw_{k^*}) \: ,~ \text{and}\\
\label{c: causal-stopp}
G(k_c) & \ge G(k^*) \:.
\end{alignat}
\end{subequations}
\end{theorem}
\vspace{-0.001\textheight}
\begin{IEEEproof}
 See Appendix~A-E in~\cite{mahmoudi2022fedcau}.
\end{IEEEproof}

\begin{remark}
Note that $k^*$ and $k_c$ are fundamentally different. $k_c$ is obtained from Algorithms~\ref{alg: synchronous alg} or~\ref{alg: asynchronous alg}, while $k^*$ is the optimal stopping iteration that we would compute if we knew beforehand the evolution of the iterations of FedAvg algorithm~\eqref{eq: local trains}, thus non-causal. Nevertheless, we show that the computation of the stopping iteration $k_c$ that we propose in the causal setting of Algorithms~\ref{alg: synchronous alg} and~\ref{alg: asynchronous alg} is almost identical to $k^*$.    
\end{remark}

Theorem~\ref{thrm: causal-alg} is a central result in our paper, showing that we can develop a simple yet close-to-optimal algorithm for optimization problem~\eqref{eq: cost-efficient-distributed-optim-3}. In other words, Algorithms~\ref{alg: synchronous alg} and~\ref{alg: asynchronous alg} in causal setting solve problem~\eqref{eq: cost-efficient-distributed-optim-3} by taking at most one extra iteration compared to the non-causal to compute the optimal termination communication iteration number. 

Next, we focus on the convergence analysis of Algorithm~\ref{alg: Nonconvex alg}. From Section~\ref{subsec: non-convex f}, we define $F_u(\bw_k)$ and $F_l(\bw_k)$ as the upper and lower bound functions for $\Tilde{F}(\bw_k)$, respectively, such that for every $k \geq 1$, inequalities $F_l(\bw_k) \le \Tilde{F}(\bw_k) \le F_u(\bw_k)$ hold. The following remark highlights the important monotonic behavior of $F_u(\bw_k)$ and $F_l(\bw_k)$.
\begin{remark}\label{remark: nonoconvex alg}
The proposed functions $F_u(\bw_k)$ and $F_l(\bw_k)$ are monotonic decreasing w.r.t. $k$, i.e., $F_u(\bw_{k}) < F_u(\bw_{k-1})$, and $F_l(\bw_{k}) < F_l(\bw_{k-1})$ for $\forall k \geq 1$. These results hold because we consider a linear function, which is monotonically decreasing, w.r.t. $k$, for updating each value of $F_l(\bw_k)$ and $F_u(\bw_k)$ for $k \geq 1$. Since the monotonically decreasing linear function fulfills the sufficient decreasing condition~(see~\cite{DBLP:books/daglib/0022091}, Section~11.5), we claim that $F_u(\bw_k)$ and $F_l(\bw_k)$ are monotonic decreasing w.r.t. $k$.
\end{remark}
 Remark~\ref{remark: nonoconvex alg} indicates that we can apply the batch FedCau update of Algorithm~\ref{alg: synchronous alg} to obtain the causal stopping point for $G_u(K)$ and $G_l(K)$ denoted as $k_c^u$ and $k_c^l$, respectively. Therefore, according to Proposition~\ref{prop: convergence}, there are finite optimal stopping iterations for minimizing $G_u(K)$ and $G_l(K)$. Thus, Theorem~\ref{thrm: causal-alg} is valid for $k_c^u$ and $k_c^l$, and we guarantee the convergence of $G_u(K)$ and $G_l(K)$. The following Proposition characterizes the relation of causal stopping iteration $k_c$ of $\Tilde{G}(K)$ with $k_c^u$ and $k_c^l$.
 
 \begin{prop}\label{prop: k_c nonconvex}
 Let $k_c$, $k_c^u$, and $k_c^l$ be the causal stopping iterations for minimizing $\Tilde{G}(K)$, $G_u(K)$, and $G_l(K)$, respectively. Then, the inequalities $k_c^u~\le~k_c~\le~k_c^l$ hold.
 \begin{IEEEproof}
 See Appendix~A-F in~\cite{mahmoudi2022fedcau}.
 \end{IEEEproof}
 \end{prop}
 
 \vspace{-0.005\textheight}
 Proposition~\ref{prop: k_c nonconvex} characterizes an interval in which $k_c$ can take values to stop Algorithm~\ref{alg: Nonconvex alg}. As $k_c^u \le k_c \le k_c^l$, it is enough that we find $k_c^u$ and terminate the algorithm. However, the maximum allowable number of iterations is $k_c^l$, which can be achieved if the resource budget allows us.  Using Proposition~\ref{prop: k_c nonconvex}, we can obtain a sub-optimal $k_c$ by applying the FedCau update Algorithm~\ref{alg: Nonconvex alg} to non-convex loss functions. 
{\color{black} 
 \begin{lemma}\label{lemma: tightness}
 Let $F_u(\bw_k)$ and $F_l(\bw_k)$ be respectively the upper bound and the lower bound of $\Tilde{F}(\bw_k)$ obtained from the stochastic non-convex cost-efficient mini-batch FedCau Algorithm~\ref{alg: Nonconvex alg}. Let us define $\Tilde{F}_{\max} := \max_{k \in [2,K]} \Tilde{F}(\bw_k) $ and $\Tilde{F}_{\min} := \min_{k \in [2,K]} \Tilde{F}(\bw_k)$. Assuming that $|\Tilde{F}(\bw_k)| < \infty, k =1,\ldots, K$, then $|F_u(\bw_k) - F_l(\bw_k)| \le \Tilde{F}_{\max}-\Tilde{F}_{\min}$ is the tightness between the upper bound~$F_u(\bw_k)$ and the lower bound~$F_l(\bw_k)$.
 \end{lemma}
 \begin{IEEEproof}
 See Appendix~\ref{P: lemma: tightness}. 
 \end{IEEEproof}
 Lemma~\ref{lemma: tightness} specifies that the maximum distance between the upper and lower bound functions $F_u(\bw_k)$ and $F_l(\bw_k)$, $k =1,\ldots,K$, is determined by the variations of non-convex sequence~$\Tilde{F}(\bw_k), k=1,\ldots, K$. In the following Proposition, we investigate the tightness of the interval~$(k_c^u, k_c^l)$.
\begin{prop}\label{prop: k_u and k_l}
Let~$K^{\text{max}}$, $k_c^u$ and $k_c^l$ be the maximum number of iterations, the causal stopping iteration for minimizing $G_u(K)$, and the causal stopping iteration for minimizing~$G_l(K)$, respectively. Recall the definition of~$k_{\max}^u$ in line~29 of Algorithm~3. Then,
\begin{alignat}{3}\label{eq: k_u and k_l}
k_c^l &&&= 1 + \\
\nonumber
    &&&\max \left \{k_c^u, k_{\max}^u + \left \lceil \frac{(1-\beta)}{\beta c_{k_d}}\left \{F_u(\boldsymbol{w}_{k_{\max}^u})-\Tilde{F}(\boldsymbol{w}_{{k_d}}) \right \}  \right \rceil \right \},
\end{alignat}
where, for $k~\in [k_c^u+1, K^{\text{max}}]$,
\begin{equation}
   k_d :=  {\textrm{the first value of}}~k \mid  \Tilde{F}(\boldsymbol{w}_{k})>F_u(\boldsymbol{w}_{k-1}).
\end{equation}
\end{prop}
\begin{IEEEproof}
 See Appendix~\ref{P: prop: k_u and k_l}. 
 \end{IEEEproof}
 Proposition~\ref{prop: k_u and k_l} denotes that the tightness of the interval~$(k_c^u, k_c^l)$ is mainly determined by $c_k$ and the variations of the non-convex sequence~$\Tilde{F}(\bw_k), k=k_c^u+1,\ldots, K^{\text{max}}$.
 
 To summarize, FedCau is applicable for both full and partial worker participation, as well as when $f(\boldsymbol{w}_k)$ is monotonically decreasing and not monotonic decreasing. Specifically, we have used the FedCau theory to propose Algorithm~3 that obtains a suboptimal solution for $k_c$ when $f(\boldsymbol{w}_k)$ is not monotonically decreasing.
\subsection{Complexity Analysis of Algorithms~\ref{alg: synchronous alg}-\ref{alg: Nonconvex alg}}\label{subsection: complexity}

In this part, we analyze the computation complexities of Algorithms~1-3 and compare them with the computation complexity of FedAvg. Recall that in FedCau of Algorithms~1-3, the stopping iteration~$k_c \le K^{\max}$,  $K^{\max}$ is the number of FedAvg global iterations. By assuming the training is done considering a neural network with $N_l$ number of layers, $d_N$ as the maximum number of neurons, the backpropagation of local gradients in each worker $j$ after $E$ local iterations, results in a complexity of~$\mathcal{O}(E N_l d_N^3)$. Thus, Algorithm~1 has the complexity of $\mathcal{O}(k_c E (|D|d + N_l d_N^3))$, which is less than or equal to the complexity of FedAvg as $\mathcal{O}(K^{\max}E( |D|d+ N_l d_N^3))$. Similarly, the complexity of Algorithm~2 is obtained as $\mathcal{O}(|D|Ed + (k_c-1)F_f^{-1}|D|Ed  + Ek_c N_l d_N^3)$. Finally, the complexity of Algorithm~3, by considering the complexity from the neural network setting we mentioned before, is obtained as $\mathcal{O}(|D|Ed(k_c^l)^3+ Ek_c^l N_l d_N^3)$.

 }
\section{Application to communication Protocols}\label{section: Protocols}

We consider wireless communication scenarios with a broadcast channel in the downlink from the master node to the workers. In the uplink, we consider three communication protocols, slotted-ALOHA~\cite{bertsekas2004data} and CSMA/CA~\cite{ziouva2002csma} with a binary exponential backoff retransmission policy~\cite{Yang2003Delay}, and OFDMA~\cite{ 8422767} by which the workers transmit their local parameters to the master node. We assume that in each communication iteration~$k$, local parameters are set at the head of the line of each node's queue and ready to be transmitted. Thus, upon receiving $\bw_k$, each worker $j \in [M]$ computes its local parameter $\bw_k^j$ and puts it in the head of the line of its transmission queue. In a parallel process, each worker may generate some background traffic and put them on the same queue, and send them by the first-in-first-out queuing policy. We obtain the average end-to-end communication-computation latency at each iteration~$k$, denoted by $c_k$, for slotted-ALOHA and CSMA/CA protocols: by taking an average over the randomness of the protocols. Hence, at the end of each communication iteration $K$, the network has faced a latency equal to $\sum_{k = 1}^{K} c_k$. It means that we consider each time slot (in $ms$) and sum up the spent computation delay and time slots in each communication iteration $k$ to achieve $c_k$, thus following the Algorithms~\ref{alg: synchronous alg},~\ref{alg: asynchronous alg}, and~\ref{alg: Nonconvex alg} to solve optimization problem~\eqref{eq: cost-efficient-distributed-optim-3}.

The critical point to consider is that we should choose a stable network in which packet saturation will not happen. We only consider the latency of transmitting local parameters, positioned at the head of line queues, which is influenced by the number of workers~$M$, transmission probability~$p_x$, and packet arrival probability~$p_r$ at each time slot. Local parameters at each iteration~$k$ are distinct from background traffic packets influenced by the probability of $p_r$.



Recall the definition of the communication-computation cost components $\ell_{1,k}, \ell_{2,k}, \ell_{3,k}$ and $\ell_{4,k}$ in Section~\ref{subsec: FLs}. For $\ell_{1,k}$, we consider a broadcast channel with data rate $R$ bits/s and parameter size of $b$ bits (which includes the payload and headers), leading to a constant latency of $\ell_{1,k}~=~b/R$ s. Also, it is natural to assume that $\ell_{4,k}$ is a given constant for updating parameters at the master node~\cite{Forough:2019}. The computation latency $\ell_{2,k}^j$ in each iteration~$k$ at each worker $j \in [M]$ is calculated as $\ell_{2,k}^j = {a_k^j |D_{k}^j|}/{\nu_k^j}$, where $a_k^j$ is the number of processing cycles to execute one sample of data~(cycles/sample), $|D_{k}^j| \le |D_j|$ is a subset of local dataset each worker chooses to update its local parameter~$\bw_k^j$, and~$\nu_k^j$ is the central processing unit~(cycles/s)~\cite{nguyen2020efficient}. Without any loss of generality, we consider that $|D_{k}^j| = |D_j|, k=1,\ldots,K$. We assume that all the worker nodes start transmitting their local parameters simultaneously. Thus, the network must wait for the slowest worker to complete its computation. Therefore, $\ell_{2,k} = \max_{j \in [M]} \ell_{2,k}^j$. The third term, $\ell_{3,k}$, is determined by the channel capacity, resource allocation policy, and network traffic. We characterize this term for two batch and mini-batch update cases with a defined time budget. Further, every specific broadcast channel imposes a particular $R$ and $b$, which do not change during the optimizing process. Therefore, to compute the iteration-cost function $\sum_k c_k$, we take into account the $\ell_{3,k}$ and $\ell_{2,k}$ terms and ignore the latency terms of $\ell_{1,k}$, and $\ell_{4,k}$ because they do not play a role in the optimization problem~\eqref{eq: cost-efficient-distributed-optim-3} in the presence of shared wireless channel for the uplink. Note that in this paper, without loss of generality, $c_k:= \ell_{2,k} + \ell_{3,k}$, in which $\ell_{2,k}$ is independent of the communication channels/protocols. 
We wish to obtain the upper bound for communication delay when the users in the network follow MAC protocols, such as slotted-ALOHA and CSAMA/CA, to transmit their local parameters of FedAvg algorithm~\eqref{eq: local trains} to the master node~\cite{Yang2003Delay, 4155680}. There are many papers in the literature computing the average transmission delay for MAC protocols. However, we have a specific assumption that at each communication iteration~$k$, each worker puts its local parameter at the head of the line in its queue and makes it ready for transmission. Note that in FedAvg algorithm~\eqref{eq: local trains}, the master node needs to receive all the local parameters to update the new global parameter~$\bw_k$. Accordingly, we calculate the average latency of the system while all workers must successfully transmit at least one packet to the master node. 
The following Proposition establishes bounds of the average transmission latency $\ell_{3,k}$.
\vspace{-0.005\textheight}
\begin{prop}\label{prop: csmaca}
Consider random access MAC protocols in which the local parameters of FedAvg algorithm~\eqref{eq: local trains} are head-of-line packets at each iteration~$k$. Let $M$, $p_x$, and $p_r$ be the number of nodes, the transmission probability at each time slot, and the background packet arrival probability at each time slot. Consider each time slot to have a duration of~$t_s$~seconds. Then, the average transmission delay, $\mathbb{E}\{\ell_{3,k}\}$ is bounded by
\vspace{-0.003\textheight}
\begin{equation}\label{eq: upper comm}
 \sum_{i=0}^{M-1}t_s p_{i,i+1} \le \mathbb{E}\left \{\ell_{3,k}\right\} \le \sum_{i=0}^{M-1}t_s   \left\{p_{i,i+1}+\frac{p_{i,i}}{(1-p_{i,i})^2}\right\}, 
\end{equation}
where 
\begin{alignat}{3}
\nonumber
p_{i,i} =& \hat{p} + (1-p_x)^{M-i}, i = 0, 1&\\
\nonumber
p_{i,i} =& i p_rp_x \sum_{j=1}^{i-1}\frac{(i-1)!}{j!(i-1-j)!} \left\{ p_r^j(1-p_x)^j(1-p_r)^{i-1-j} \right\}\\
\nonumber
&+\hat{p} + (1-p_x)^M, i \ge 2 &
\end{alignat}
and
\vspace{-0.005\textheight}
\begin{alignat}{3}
\nonumber
&p_{i,i+1} = \\
\nonumber
&\left(M-i\right)(1-p_x)^{M-i-1}p_x{\sum_{j=1}^{i} p_r^j(1-p_x)^j(1-p_r)^{i-j}},
\end{alignat}
where $\hat{p}$ is the probability of an idle time slot.
\end{prop}
\vspace{-0.001\textheight}{
\begin{IEEEproof}
See Appendix~A-G in~\cite{mahmoudi2022fedcau}.
\end{IEEEproof}}
%
\vspace{0mm}{
\begin{figure*}[t]
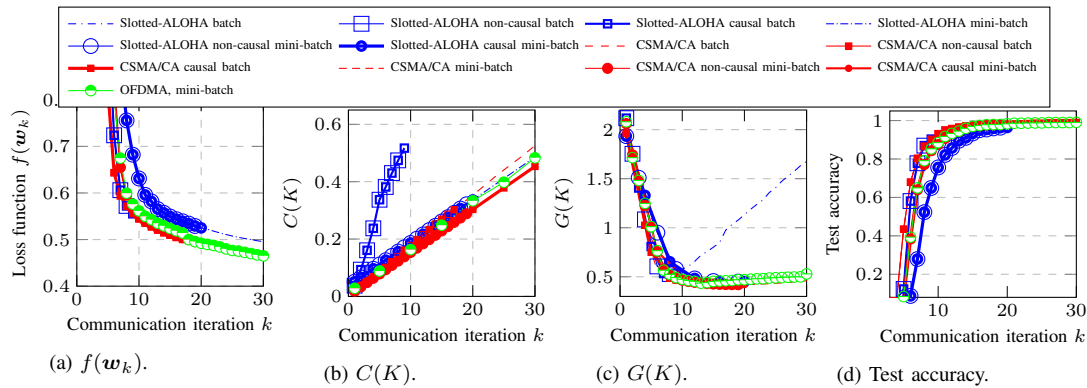

\vspace{15mm}
\centering
\begin{minipage}{0.28\columnwidth}
\vspace{-15mm}
{\scriptsize\input{./figs/f}}
\subcaption{$f(\bw_k)$.}
\label{subfig: f}
\end{minipage}
\hspace{0.1\columnwidth}
\begin{minipage}{0.28\columnwidth}
{\scriptsize\input{./figs/C}}
\subcaption{$C(K)$.}
\label{subfig: C}
\end{minipage}
\hspace{0.1\columnwidth}
\begin{minipage}{0.28\columnwidth}
{\scriptsize\input{./figs/G}}
\subcaption{$G(K)$.}
\label{subfig: G}
\end{minipage}
\hspace{0.1\columnwidth}
\begin{minipage}{0.28\columnwidth}
{\scriptsize\input{./figs/AC}}
\subcaption{Test accuracy.}
\label{subfig: AC}
\end{minipage}
\caption{Illustration of non-causal and FedCau batch update of Algorithm~\ref{alg: synchronous alg} and FedCau mini-batch update of Algorithm~\ref{alg: asynchronous alg} with $T = 0.3$s the presence of slotted ALOHA and CSMA/CA, and OFDMA for $M = 100$, $p_x = 1$, and $p_r = 0.2$. a) Loss function~$f(\bw_k)$ b) Iteration-cost function $C(K)$ c) Multi-objective cost function $G(K)$, and d) Test accuracy.
}
\label{fig: Batch1k}

\end{figure*} 
}


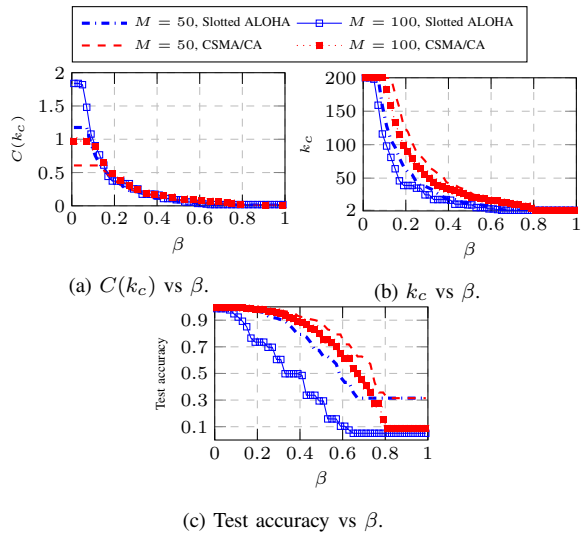
\begin{figure}[t]
\vspace{4mm}
\centering
\begin{minipage}{0.4\columnwidth}
\vspace{-0.035\textheight}
{\scriptsize\begin{tikzpicture}

\begin{axis}[%
width=0.8\columnwidth,
height=0.5\columnwidth,
at={(0,0)},
scale only axis,
xmin=0, 
xmax=1,
xlabel={$\beta$},
xtick={0,0.2,0.4,0.6,0.8,1},
xtick style={color=black},
ymin=0,
ymax=2,
ytick={0, 0.5, 1, 1.5, 2, 2.5},
grid style={dashed},
grid=both,
ymajorgrids,
ylabel near ticks,
ylabel={\tiny{$C(k_c)$}},
axis background/.style={fill=white},
legend style={at={(0.0,1.5)},legend columns=2, font = \tiny,  anchor=north west, legend cell align=left, align=left}
]
\addplot [color=blue, dashdotted, very thick]
  table[row sep=crcr]{%
0.01	1.17686068682298\\
0.03	1.17686068682298\\
0.05	1.17686068682298\\
0.07	1.17686068682298\\
0.09	0.93123942959118\\
0.11	0.753954869048859\\
0.13	0.629628170288071\\
0.15	0.564301033667008\\
0.17	0.471011914258495\\
0.19	0.395590992651445\\
0.21	0.348758335442912\\
0.23	0.325475570978149\\
0.25	0.260832657668224\\
0.27	0.23004260866699\\
0.29	0.223579552857543\\
0.31	0.206453946109202\\
0.33	0.189417808162442\\
0.35	0.161072387653931\\
0.37	0.137207223895067\\
0.39	0.113047239132897\\
0.41	0.113047239132897\\
0.43	0.106092530956667\\
0.45	0.0946430806057722\\
0.47	0.0767071421402546\\
0.49	0.0767071421402546\\
0.51	0.0651034646171107\\
0.53	0.0651034646171107\\
0.55	0.0582297044042158\\
0.57	0.0524849552588908\\
0.59	0.0362885459212553\\
0.61	0.0295907410181364\\
0.63	0.0295907410181364\\
0.65	0.0177071539586176\\
0.67	0.0115517004098445\\
0.69	0.0115517004098445\\
0.71	0.0115517004098445\\
0.73	0.0115517004098445\\
0.75	0.0115517004098445\\
0.77	0.0115517004098445\\
0.79	0.0115517004098445\\
0.81	0.0115517004098445\\
0.83	0.0115517004098445\\
0.85	0.0115517004098445\\
0.87	0.0115517004098445\\
0.89	0.0115517004098445\\
0.91	0.0115517004098445\\
0.93	0.0115517004098445\\
0.95	0.0115517004098445\\
0.97	0.0115517004098445\\
0.99	0.0115517004098445\\
};
\addlegendentry{$M=50$, Slotted ALOHA}

\addplot [color=blue, mark=square, mark options={scale=0.6}]
  table[row sep=crcr]{%
0.01	1.83883336509111\\
0.03	1.83883336509111\\
0.05	1.81985871109157\\
0.07	1.48044781765102\\
0.09	1.07824533366628\\
0.11	0.90280943523672\\
0.13	0.765437896346362\\
0.15	0.612012242551538\\
0.17	0.442470523022959\\
0.19	0.372430685451879\\
0.21	0.372430685451879\\
0.23	0.372430685451879\\
0.25	0.33311845920842\\
0.27	0.33311845920842\\
0.29	0.253839097954727\\
0.31	0.253839097954727\\
0.33	0.175143844520049\\
0.35	0.175143844520049\\
0.37	0.175143844520049\\
0.39	0.175143844520049\\
0.41	0.160797802480281\\
0.43	0.109741469027216\\
0.45	0.109741469027216\\
0.47	0.109741469027216\\
0.49	0.0908907166862506\\
0.51	0.0908907166862506\\
0.53	0.0587647866403802\\
0.55	0.0587647866403802\\
0.57	0.0587647866403802\\
0.59	0.0360553826465224\\
0.61	0.0360553826465224\\
0.63	0.0268158589639084\\
0.65	0.018620649260747\\
0.67	0.018620649260747\\
0.69	0.018620649260747\\
0.71	0.018620649260747\\
0.73	0.018620649260747\\
0.75	0.018620649260747\\
0.77	0.018620649260747\\
0.79	0.018620649260747\\
0.81	0.018620649260747\\
0.83	0.018620649260747\\
0.85	0.018620649260747\\
0.87	0.018620649260747\\
0.89	0.018620649260747\\
0.91	0.018620649260747\\
0.93	0.018620649260747\\
0.95	0.018620649260747\\
0.97	0.018620649260747\\
0.99	0.018620649260747\\
};
\addlegendentry{$M=100$, Slotted ALOHA}

\addplot [color=red, dashed, thick]
  table[row sep=crcr]{%
0.01	0.606790988761746\\
0.03	0.606790988761746\\
0.05	0.606790988761746\\
0.07	0.606790988761746\\
0.09	0.606790988761746\\
0.11	0.606790988761746\\
0.13	0.606790988761746\\
0.15	0.533789241238538\\
0.17	0.470334238075031\\
0.19	0.387376954608982\\
0.21	0.359933878411559\\
0.23	0.31192556036935\\
0.25	0.269440453590549\\
0.27	0.250803332451313\\
0.29	0.229637960774479\\
0.31	0.199354804406779\\
0.33	0.187263778467005\\
0.35	0.17221399092959\\
0.37	0.166405840814429\\
0.39	0.1335031603931\\
0.41	0.124293326490597\\
0.43	0.120853633618675\\
0.45	0.108937139189729\\
0.47	0.102814832355077\\
0.49	0.0941288515502604\\
0.51	0.0821195322140738\\
0.53	0.0700848734881669\\
0.55	0.0666927566132706\\
0.57	0.0575092963612932\\
0.59	0.0575092963612932\\
0.61	0.045234947693849\\
0.63	0.045234947693849\\
0.65	0.0391416001617348\\
0.67	0.0330989314090611\\
0.69	0.0330989314090611\\
0.71	0.0300287282096924\\
0.73	0.0271131470006574\\
0.75	0.0153900593422289\\
0.77	0.00917958177395897\\
0.79	0.00917958177395897\\
0.81	0.00592269449930438\\
0.83	0.00592269449930438\\
0.85	0.00592269449930438\\
0.87	0.00592269449930438\\
0.89	0.00592269449930438\\
0.91	0.00592269449930438\\
0.93	0.00592269449930438\\
0.95	0.00592269449930438\\
0.97	0.00592269449930438\\
0.99	0.00592269449930438\\
};
\addlegendentry{$M=50$, CSMA/CA}

\addplot [color=red, mark=square*, thin,mark options={scale=0.6}, dotted]
  table[row sep=crcr]{%
0.01	0.973394895093958\\
0.07	0.973394895093958\\
0.11	0.886100411757409\\
0.15	0.647222012144639\\
0.19	0.481993578992194\\
0.23	0.38497144601334\\
0.27	0.311491742279844\\
0.31	0.243301932442993\\
0.35	0.18868263909625\\
0.39	0.164066895058395\\
0.43	0.149395030203335\\
0.47	0.130514419154076\\
0.53	0.102018292752338\\
0.57	0.0917361555371851\\
0.61	0.0771178842765263\\
0.65	0.0675761631809527\\
0.69	0.0577081395471571\\
0.73	0.0480445960889244\\
0.75	0.0384647410896419\\
0.79	0.0240241532073453\\
0.91	0.00983463070189923\\
0.99	0.00983463070189923\\
};
\addlegendentry{$M=100$, CSMA/CA}

\end{axis}
\end{tikzpicture}%
}

\subcaption{ $C(k_c)$ vs $\beta$.}
\label{subfig: ckvsbeta}
\end{minipage}
\hspace{0.005\columnwidth}
\begin{minipage}{0.4\columnwidth}
{\scriptsize\begin{tikzpicture}

\begin{axis}[%
width=0.8\columnwidth,
height=0.5\columnwidth,
at={(0,0)},
scale only axis,
xmin=0, 
xmax=1,
xlabel={$\beta$},
xtick={0,0.2,0.4,0.6,0.8,1},
xtick style={color=black},
ymin=1,
ymax=200,
ytick={2, 50, 100, 150, 200},
grid style={dashed},
grid=both,
ymajorgrids,
ylabel near ticks,
ylabel={\tiny{$k_c$}},
axis background/.style={fill=white},
legend style={at={(0.3,1.5)},legend columns=4, font = \tiny,  anchor=north west, legend cell align=left, align=left}
]
\addplot [color=blue, dashdotted, very thick]
  table[row sep=crcr]{%
0.01	200\\
0.03	200\\
0.05	200\\
0.07	200\\
0.09	158\\
0.11	128\\
0.13	107\\
0.15	96\\
0.17	80\\
0.19	67\\
0.21	59\\
0.23	55\\
0.25	44\\
0.27	39\\
0.29	38\\
0.31	35\\
0.33	32\\
0.35	27\\
0.37	23\\
0.39	19\\
0.41	19\\
0.43	18\\
0.45	16\\
0.47	13\\
0.49	13\\
0.51	11\\
0.53	11\\
0.55	10\\
0.57	9\\
0.59	6\\
0.61	5\\
0.63	5\\
0.65	3\\
0.67	2\\
0.69	2\\
0.71	2\\
0.73	2\\
0.75	2\\
0.77	2\\
0.79	2\\
0.81	2\\
0.83	2\\
0.85	2\\
0.87	2\\
0.89	2\\
0.91	2\\
0.93	2\\
0.95	2\\
0.97	2\\
0.99	2\\
};

\addplot [color=blue, mark=square, mark options={scale=0.6}]
  table[row sep=crcr]{%
0.01	200\\
0.03	200\\
0.05	198\\
0.07	159\\
0.09	116\\
0.11	98\\
0.13	81\\
0.15	64\\
0.17	46\\
0.19	39\\
0.21	39\\
0.23	39\\
0.25	35\\
0.27	35\\
0.29	25\\
0.31	25\\
0.33	18\\
0.35	18\\
0.37	18\\
0.39	18\\
0.41	17\\
0.43	11\\
0.45	11\\
0.47	11\\
0.49	10\\
0.51	10\\
0.53	6\\
0.55	6\\
0.57	6\\
0.59	4\\
0.61	4\\
0.63	3\\
0.65	2\\
0.67	2\\
0.69	2\\
0.71	2\\
0.73	2\\
0.75	2\\
0.77	2\\
0.79	2\\
0.81	2\\
0.83	2\\
0.85	2\\
0.87	2\\
0.89	2\\
0.91	2\\
0.93	2\\
0.95	2\\
0.97	2\\
0.99	2\\
};

\addplot [color=red, dashed, thick]
  table[row sep=crcr]{%
0.01	200\\
0.03	200\\
0.05	200\\
0.07	200\\
0.09	200\\
0.11	200\\
0.13	200\\
0.15	176\\
0.17	155\\
0.19	128\\
0.21	119\\
0.23	103\\
0.25	89\\
0.27	83\\
0.29	76\\
0.31	66\\
0.33	62\\
0.35	57\\
0.37	55\\
0.39	44\\
0.41	41\\
0.43	40\\
0.45	36\\
0.47	34\\
0.49	31\\
0.51	27\\
0.53	23\\
0.55	22\\
0.57	19\\
0.59	19\\
0.61	15\\
0.63	15\\
0.65	13\\
0.67	11\\
0.69	11\\
0.71	10\\
0.73	9\\
0.75	5\\
0.77	3\\
0.79	3\\
0.81	2\\
0.83	2\\
0.85	2\\
0.87	2\\
0.89	2\\
0.91	2\\
0.93	2\\
0.95	2\\
0.97	2\\
0.99	2\\
};

\addplot [color=red, mark=square*, thin,mark options={scale=0.6}, dotted]
  table[row sep=crcr]{%
0.01	200\\
0.03	200\\
0.05	200\\
0.07	200\\
0.09	200\\
0.11	182\\
0.13	159\\
0.15	133\\
0.17	115\\
0.19	99\\
0.21	90\\
0.23	79\\
0.25	68\\
0.27	64\\
0.29	56\\
0.31	50\\
0.33	44\\
0.35	39\\
0.37	37\\
0.39	34\\
0.41	33\\
0.43	31\\
0.45	28\\
0.47	27\\
0.49	24\\
0.51	21\\
0.53	21\\
0.55	20\\
0.57	19\\
0.59	17\\
0.61	16\\
0.63	16\\
0.65	14\\
0.67	13\\
0.69	12\\
0.71	11\\
0.73	10\\
0.75	8\\
0.77	8\\
0.79	5\\
0.81	2\\
0.83	2\\
0.85	2\\
0.87	2\\
0.89	2\\
0.91	2\\
0.93	2\\
0.95	2\\
0.97	2\\
0.99	2\\
};

\end{axis}
\end{tikzpicture}%
}

\subcaption{$k_c$ vs $\beta$.}
\label{subfig: kcvsbeta}
\end{minipage}
\begin{minipage}{0.4\columnwidth}

{\scriptsize\begin{tikzpicture}

\begin{axis}[%
width=0.8\columnwidth,
height=0.5\columnwidth,
at={(0,0)},
scale only axis,
xmin=0, 
xmax=1,
xlabel={$\beta$},
xtick={0,0.2,0.4,0.6,0.8,1},
xtick style={color=black},
ymin=0,
ymax=1,
ytick={0.1, 0.3, 0.5, 0.7, 0.9},
grid style={dashed},
grid=both,
ymajorgrids,
ylabel near ticks,
ylabel={\tiny{Test accuracy}},
axis background/.style={fill=white},
legend style={at={(0.3,1.5)},legend columns=4, font = \tiny,  anchor=north west, legend cell align=left, align=left}
]
\addplot [color=blue, dashdotted, very thick]
  table[row sep=crcr]{%
0.01	0.996217494089835\\
0.03	0.996217494089835\\
0.05	0.996217494089835\\
0.07	0.996217494089835\\
0.09	0.994326241134752\\
0.11	0.992434988179669\\
0.13	0.987706855791962\\
0.15	0.983924349881797\\
0.17	0.981087470449173\\
0.19	0.973522458628842\\
0.21	0.964066193853428\\
0.23	0.95839243498818\\
0.25	0.931914893617021\\
0.27	0.920567375886525\\
0.29	0.914893617021277\\
0.31	0.904491725768321\\
0.33	0.893144208037825\\
0.35	0.865721040189125\\
0.37	0.830732860520095\\
0.39	0.79290780141844\\
0.41	0.79290780141844\\
0.43	0.773995271867612\\
0.45	0.733333333333333\\
0.47	0.672813238770686\\
0.49	0.672813238770686\\
0.51	0.6274231678487\\
0.53	0.6274231678487\\
0.55	0.602836879432624\\
0.57	0.570685579196218\\
0.59	0.473286052009456\\
0.61	0.432624113475177\\
0.63	0.432624113475177\\
0.65	0.356973995271868\\
0.67	0.314420803782506\\
0.69	0.314420803782506\\
0.71	0.314420803782506\\
0.73	0.314420803782506\\
0.75	0.314420803782506\\
0.77	0.314420803782506\\
0.79	0.314420803782506\\
0.81	0.314420803782506\\
0.83	0.314420803782506\\
0.85	0.314420803782506\\
0.87	0.314420803782506\\
0.89	0.314420803782506\\
0.91	0.314420803782506\\
0.93	0.314420803782506\\
0.95	0.314420803782506\\
0.97	0.314420803782506\\
0.99	0.314420803782506\\
};

\addplot [color=blue, mark=square, mark options={scale=0.6}]
  table[row sep=crcr]{%
0.01	0.986761229314421\\
0.03	0.986761229314421\\
0.05	0.986761229314421\\
0.07	0.977304964539007\\
0.09	0.947990543735225\\
0.11	0.923404255319149\\
0.13	0.892198581560284\\
0.15	0.850591016548463\\
0.17	0.76548463356974\\
0.19	0.735224586288416\\
0.21	0.735224586288416\\
0.23	0.735224586288416\\
0.25	0.69645390070922\\
0.27	0.69645390070922\\
0.29	0.604728132387707\\
0.31	0.604728132387707\\
0.33	0.497872340425532\\
0.35	0.497872340425532\\
0.37	0.497872340425532\\
0.39	0.497872340425532\\
0.41	0.484633569739953\\
0.43	0.336170212765957\\
0.45	0.336170212765957\\
0.47	0.336170212765957\\
0.49	0.293617021276596\\
0.51	0.293617021276596\\
0.53	0.15839243498818\\
0.55	0.15839243498818\\
0.57	0.15839243498818\\
0.59	0.10354609929078\\
0.61	0.10354609929078\\
0.63	0.075177304964539\\
0.65	0.0505910165484633\\
0.67	0.0505910165484633\\
0.69	0.0505910165484633\\
0.71	0.0505910165484633\\
0.73	0.0505910165484633\\
0.75	0.0505910165484633\\
0.77	0.0505910165484633\\
0.79	0.0505910165484633\\
0.81	0.0505910165484633\\
0.83	0.0505910165484633\\
0.85	0.0505910165484633\\
0.87	0.0505910165484633\\
0.89	0.0505910165484633\\
0.91	0.0505910165484633\\
0.93	0.0505910165484633\\
0.95	0.0505910165484633\\
0.97	0.0505910165484633\\
0.99	0.0505910165484633\\
};

\addplot [color=red, dashed, thick]
  table[row sep=crcr]{%
0.01	0.996217494089835\\
0.03	0.996217494089835\\
0.05	0.996217494089835\\
0.07	0.996217494089835\\
0.09	0.996217494089835\\
0.11	0.996217494089835\\
0.13	0.996217494089835\\
0.15	0.996217494089835\\
0.17	0.994326241134752\\
0.19	0.992434988179669\\
0.21	0.990543735224586\\
0.23	0.983924349881797\\
0.25	0.982978723404255\\
0.27	0.982033096926714\\
0.29	0.97919621749409\\
0.31	0.971631205673759\\
0.33	0.966903073286052\\
0.35	0.962174940898345\\
0.37	0.95839243498818\\
0.39	0.931914893617021\\
0.41	0.923404255319149\\
0.43	0.921513002364066\\
0.45	0.907328605200946\\
0.47	0.900709219858156\\
0.49	0.890307328605201\\
0.51	0.865721040189125\\
0.53	0.830732860520095\\
0.55	0.820330969267139\\
0.57	0.79290780141844\\
0.59	0.79290780141844\\
0.61	0.718203309692671\\
0.63	0.718203309692671\\
0.65	0.672813238770686\\
0.67	0.6274231678487\\
0.69	0.6274231678487\\
0.71	0.602836879432624\\
0.73	0.570685579196218\\
0.75	0.432624113475177\\
0.77	0.356973995271868\\
0.79	0.356973995271868\\
0.81	0.314420803782506\\
0.83	0.314420803782506\\
0.85	0.314420803782506\\
0.87	0.314420803782506\\
0.89	0.314420803782506\\
0.91	0.314420803782506\\
0.93	0.314420803782506\\
0.95	0.314420803782506\\
0.97	0.314420803782506\\
0.99	0.314420803782506\\
};

\addplot [color=red, mark=square*, thin,mark options={scale=0.6}, dotted]
  table[row sep=crcr]{%
0.01	0.99338061465721\\
0.03	0.99338061465721\\
0.05	0.99338061465721\\
0.07	0.99338061465721\\
0.09	0.99338061465721\\
0.11	0.99338061465721\\
0.13	0.992434988179669\\
0.15	0.989598108747045\\
0.17	0.986761229314421\\
0.19	0.983924349881797\\
0.21	0.981087470449173\\
0.23	0.978250591016548\\
0.25	0.967848699763593\\
0.27	0.964066193853428\\
0.29	0.960283687943262\\
0.31	0.95177304964539\\
0.33	0.931914893617021\\
0.35	0.916784869976359\\
0.37	0.911111111111111\\
0.39	0.88936170212766\\
0.41	0.884633569739953\\
0.43	0.867612293144208\\
0.45	0.838297872340426\\
0.47	0.833569739952719\\
0.49	0.802364066193853\\
0.51	0.751300236406619\\
0.53	0.751300236406619\\
0.55	0.732387706855792\\
0.57	0.703073286052009\\
0.59	0.649172576832151\\
0.61	0.613238770685579\\
0.63	0.613238770685579\\
0.65	0.534751773049645\\
0.67	0.497872340425532\\
0.69	0.45531914893617\\
0.71	0.40709219858156\\
0.73	0.359810874704492\\
0.75	0.271867612293144\\
0.77	0.271867612293144\\
0.79	0.152718676122931\\
0.81	0.0855791962174941\\
0.83	0.0855791962174941\\
0.85	0.0855791962174941\\
0.87	0.0855791962174941\\
0.89	0.0855791962174941\\
0.91	0.0855791962174941\\
0.93	0.0855791962174941\\
0.95	0.0855791962174941\\
0.97	0.0855791962174941\\
0.99	0.0855791962174941\\
};

\end{axis}
\end{tikzpicture}%
}
\subcaption{Test accuracy vs $\beta$.}
\label{subfig: accvsbeta}
\end{minipage}
\caption{{\color{black}Illustration of the effect of $\beta$ on the performance of batch FedCau update of Algorithm~\ref{alg: synchronous alg}, $M=50, 100$ and CSMA/CA with $p_x = 1$, $p_r = 0.01$. a) The causal iteration-cost $C(k_c)$ decreases while $\beta$ increases. b) The causal stopping iteration~$k_c$ is smaller for larger $\beta$. c) Test accuracy also decreases when $\beta$ increases.} 
}
\label{fig: BetaPerf}
\end{figure}
Proposition~\ref{prop: csmaca} introduces the bounds for transmission delay, thus for $c_k$, while considering slotted-ALOHA and CSMA/CA communication protocols. Recall that $c_k = \ell_{2,k} + \ell_{3,k}$, then by considering the slowest worker in local iteration, the iteration cost $c_k$ is bounded by
\begin{alignat}{3}\label{eq: ck_bounds}
\nonumber
&\sum_{i=0}^{M-1}t_s p_{i,i+1} + \min_{j\in [M]} \left \{\frac{|D_j| a_k^j}{\nu_k^j}\right\} \le  c_k \le 
\\
& |D|\max_{j\in [M]} \left \{\frac{a_k^j}{\nu_k^j}\right\}+ t_s \sum_{i=0}^{M-1} \left\{p_{i,i+1}+\frac{p_{i,i}}{(1-p_{i,i})^2}\right\},
\end{alignat}
which helps us to design the communication-computation parameters for FedCau. {\color{black}Note that we consider a setup where the transmission starts simultaneously for all the workers. This is an important setup by which we have developed Algorithms~1-3 and the bounds on the iteration-cost $c_k$ in Proposition~\ref{prop: csmaca} and inequalities~\eqref{eq: ck_bounds}. The assumption that all workers transmit at each iteration~$k$ is only for Algorithm~1. However, in the updated Algorithm~2, we can consider either partial or full worker participation, which allows us to skip the slowest worker and not wait for it at \textit{each} iteration~$k$. Finally, in Algorithm~3, we have developed a general approach by which FedCau can be applied to any scenario, e.g., full or partial worker participation, non-convex loss functions $f(\boldsymbol{w})$ or any $G(K)$ with various local optimum points. Thus, the assumption that workers start transmissions to the master node simultaneously does not contradict the cost-efficiency of FedCau because we have considered various scenarios, like full or partial worker participation, in Algorithms~1-3.}
\begin{figure*}[t]
\centering
\begin{minipage}{0.4\columnwidth}
{\scriptsize\definecolor{amber}{rgb}{1.0, 0.49, 0.0}
\definecolor{taupe}{rgb}{0.28, 0.24, 0.2}
\definecolor{tealgreen}{rgb}{0.0, 0.51, 0.5}
\begin{tikzpicture}
\begin{axis}[%
width=0.8\columnwidth,
height=0.5\columnwidth,
at={(0,0)},
scale only axis,
xmin=0, 
xmax=200,
xlabel={Communication iteration $k$},
xtick style={color=black},
ymin=0,
ymax=0.45,
ytick={0.1,0.3},
grid style={dashed},
ymajorgrids,
grid = both,
ylabel near ticks,
ylabel={$C(k_c)$},
axis background/.style={fill=white},
legend style={at={(0,0.99)}, anchor=north east,  legend cell align=left,font = \tiny}
]
\addplot [color=tealgreen, mark=halfcircle* ]
  table[row sep=crcr]{%
1	0.0040728\\
8	0.0325824\\
16	0.0651648\\
24	0.0977472\\
32	0.1303296\\
40	0.162912\\
48	0.1954944\\
56	0.2280768\\
64	0.2606592\\
72	0.2932416\\
80	0.325824\\
88	0.3584064\\
96	0.3909888\\
};
\addplot [color=amber, mark=otimes, mark options={scale=0.85}]
  table[row sep=crcr]{%
1	0.0034338\\
8	0.0223338\\
16	0.0439338\\
24	0.0655338\\
32	0.0871338\\
40	0.1087338\\
48	0.1303338\\
56	0.1519338\\
64	0.1735338\\
72	0.1951338\\
80	0.2167338\\
88	0.2383338\\
96	0.2599338\\
104	0.2815338\\
112	0.3031338\\
120	0.3247338\\
126	0.3409338\\
};
\addplot [color=taupe, dashed, line width=2pt] 
  table[row sep=crcr]{%
1	0.0038844\\
2	0.0053844\\
3	0.0068844\\
4	0.0083844\\
5	0.0098844\\
6	0.0113844\\
7	0.0128844\\
8	0.0143844\\
9	0.0158844\\
10	0.0173844\\
11	0.0188844\\
12	0.0203844\\
13	0.0218844\\
14	0.0233844\\
15	0.0248844\\
16	0.0263844\\
17	0.0278844\\
18	0.0293844\\
19	0.0308844\\
20	0.0323844\\
21	0.0338844\\
22	0.0353844\\
23	0.0368844\\
24	0.0383844\\
25	0.0398844\\
26	0.0413844\\
27	0.0428844\\
28	0.0443844\\
29	0.0458844\\
30	0.0473844\\
31	0.0488844\\
32	0.0503844\\
33	0.0518844\\
34	0.0533844\\
35	0.0548844\\
36	0.0563844\\
37	0.0578844\\
38	0.0593844\\
39	0.0608844\\
40	0.0623844\\
41	0.0638844\\
42	0.0653844\\
43	0.0668844\\
44	0.0683844\\
45	0.0698844\\
46	0.0713844\\
47	0.0728844\\
48	0.0743844\\
49	0.0758844\\
50	0.0773844\\
51	0.0788844\\
52	0.0803844\\
53	0.0818844\\
54	0.0833844\\
55	0.0848844\\
56	0.0863844\\
57	0.0878844\\
58	0.0893844\\
59	0.0908844\\
60	0.0923844\\
61	0.0938844\\
62	0.0953844\\
63	0.0968844\\
64	0.0983844\\
65	0.0998844\\
66	0.1013844\\
67	0.1028844\\
68	0.1043844\\
69	0.1058844\\
70	0.1073844\\
71	0.1088844\\
72	0.1103844\\
73	0.1118844\\
74	0.1133844\\
75	0.1148844\\
76	0.1163844\\
77	0.1178844\\
78	0.1193844\\
79	0.1208844\\
80	0.1223844\\
81	0.1238844\\
82	0.1253844\\
83	0.1268844\\
84	0.1283844\\
85	0.1298844\\
86	0.1313844\\
87	0.1328844\\
88	0.1343844\\
89	0.1358844\\
90	0.1373844\\
91	0.1388844\\
92	0.1403844\\
93	0.1418844\\
94	0.1433844\\
95	0.1448844\\
96	0.1463844\\
97	0.1478844\\
98	0.1493844\\
99	0.1508844\\
100	0.1523844\\
101	0.1538844\\
102	0.1553844\\
103	0.1568844\\
104	0.1583844\\
105	0.1598844\\
106	0.1613844\\
107	0.1628844\\
108	0.1643844\\
109	0.1658844\\
110	0.1673844\\
111	0.1688844\\
112	0.1703844\\
113	0.1718844\\
114	0.1733844\\
115	0.1748844\\
116	0.1763844\\
117	0.1778844\\
118	0.1793844\\
119	0.1808844\\
120	0.1823844\\
121	0.1838844\\
122	0.1853844\\
123	0.1868844\\
124	0.1883844\\
125	0.1898844\\
126	0.1913844\\
127	0.1928844\\
128	0.1943844\\
129	0.1958844\\
130	0.1973844\\
131	0.1988844\\
132	0.2003844\\
133	0.2018844\\
134	0.2033844\\
135	0.2048844\\
136	0.2063844\\
137	0.2078844\\
138	0.2093844\\
139	0.2108844\\
140	0.2123844\\
141	0.2138844\\
142	0.2153844\\
143	0.2168844\\
144	0.2183844\\
145	0.2198844\\
146	0.2213844\\
147	0.2228844\\
148	0.2243844\\
149	0.2258844\\
150	0.2273844\\
151	0.2288844\\
152	0.2303844\\
153	0.2318844\\
154	0.2333844\\
155	0.2348844\\
156	0.2363844\\
157	0.2378844\\
158	0.2393844\\
159	0.2408844\\
160	0.2423844\\
161	0.2438844\\
162	0.2453844\\
163	0.2468844\\
164	0.2483844\\
165	0.2498844\\
166	0.2513844\\
167	0.2528844\\
168	0.2543844\\
169	0.2558844\\
170	0.2573844\\
171	0.2588844\\
172	0.2603844\\
173	0.2618844\\
174	0.2633844\\
175	0.2648844\\
176	0.2663844\\
177	0.2678844\\
178	0.2693844\\
179	0.2708844\\
180	0.2723844\\
181	0.2738844\\
182	0.2753844\\
183	0.2768844\\
184	0.2783844\\
185	0.2798844\\
186	0.2813844\\
187	0.2828844\\
188	0.2843844\\
189	0.2858844\\
190	0.2873844\\
191	0.2888844\\
};
\end{axis}
\end{tikzpicture}%
}

\subcaption{Causal iteration-cost $M=50$, $F_f = 10$.}
\label{subfig: CSminiC}
\end{minipage}
\hspace{0.02\columnwidth}
\begin{minipage}{0.4\columnwidth}

{\scriptsize\definecolor{amber}{rgb}{1.0, 0.49, 0.0}
\definecolor{taupe}{rgb}{0.28, 0.24, 0.2}
\definecolor{tealgreen}{rgb}{0.0, 0.51, 0.5}
\begin{tikzpicture}
\begin{axis}[%
width=0.8\columnwidth,
height=0.5\columnwidth,
at={(0,0)},
scale only axis,
xmin=0, 
xmax=200,
xlabel={Communication iteration $k$},
xtick style={color=black},
ymin=0.08,
ymax=1.1,
ytick={0,0.2, 0.4, 0.6, 0.8, 1},
grid style={dashed},
ymajorgrids,
grid=both,
ylabel near ticks,
ylabel={Test accuracy},
axis background/.style={fill=white},
legend style={at={(0.99,0)}, anchor=south east, legend cell align=left, font = \tiny}
]
\addplot [color=tealgreen, mark=halfcircle* ]
  table[row sep=crcr]{%
1	0.433569739952719\\
6	0.549881796690307\\
11	0.640661938534279\\
18	0.707801418439716\\
25	0.764539007092199\\
31	0.808983451536643\\
37	0.842080378250591\\
48	0.872340425531915\\
56	0.892198581560284\\
65	0.91016548463357\\
72	0.920567375886525\\
80	0.93096926713948\\
87	0.938534278959811\\
92	0.945153664302601\\
96	0.946099290780142\\
};
\addlegendentry{$T=1.1$}
\addplot [color=amber, mark=otimes, mark options={scale=0.85}] 
  table[row sep=crcr]{%
1	-0.00520094562647744\\
2	0.00520094562647755\\
4	0.0260047281323877\\
8	0.0884160756501182\\
12	0.183924349881797\\
16	0.298345153664303\\
20	0.448699763593381\\
24	0.589598108747045\\
28	0.695508274231678\\
32	0.785342789598109\\
36	0.848699763593381\\
40	0.888416075650118\\
44	0.916784869976359\\
47	0.928132387706856\\
53	0.95177304964539\\
65	0.973522458628842\\
81	0.984869976359338\\
126	0.989598108747045\\
};
\addlegendentry{$T=0.9$}
\addplot [color=taupe, dashed, line width=2pt] 
  table[row sep=crcr]{%
1	-0.0572104018912529\\
2	-0.0250591016548463\\
3	0.00898345153664304\\
4	0.0534278959810874\\
5	0.0997635933806147\\
6	0.146099290780142\\
7	0.185815602836879\\
8	0.229314420803783\\
9	0.264302600472813\\
10	0.307801418439716\\
11	0.339007092198582\\
12	0.387234042553191\\
13	0.416548463356974\\
14	0.447754137115839\\
15	0.47612293144208\\
16	0.498817966903073\\
17	0.52434988179669\\
18	0.544208037825059\\
19	0.5725768321513\\
20	0.596217494089834\\
21	0.613238770685579\\
22	0.633096926713948\\
23	0.652955082742317\\
24	0.670921985815603\\
25	0.686997635933806\\
26	0.700236406619385\\
27	0.712529550827423\\
28	0.721985815602837\\
29	0.730496453900709\\
30	0.741843971631206\\
31	0.747517730496454\\
32	0.755082742316785\\
33	0.761702127659574\\
34	0.768321513002364\\
35	0.774940898345154\\
36	0.782505910165485\\
37	0.788179669030733\\
38	0.794799054373522\\
39	0.802364066193853\\
40	0.808037825059102\\
41	0.817494089834515\\
42	0.824113475177305\\
43	0.828841607565012\\
44	0.831678486997636\\
45	0.833569739952719\\
46	0.837352245862884\\
47	0.846808510638298\\
48	0.854373522458629\\
49	0.859101654846336\\
50	0.862884160756501\\
51	0.866666666666667\\
52	0.872340425531915\\
53	0.873286052009456\\
54	0.87612293144208\\
55	0.879905437352246\\
56	0.884633569739953\\
57	0.884633569739953\\
58	0.890307328605201\\
59	0.897872340425532\\
60	0.902600472813239\\
61	0.906382978723404\\
62	0.91016548463357\\
63	0.911111111111111\\
64	0.913002364066194\\
65	0.913002364066194\\
66	0.913947990543735\\
67	0.916784869976359\\
68	0.919621749408983\\
69	0.923404255319149\\
70	0.926241134751773\\
71	0.928132387706856\\
72	0.929078014184397\\
73	0.93096926713948\\
74	0.931914893617021\\
75	0.932860520094563\\
76	0.933806146572104\\
77	0.933806146572104\\
78	0.935697399527187\\
79	0.93758865248227\\
80	0.938534278959811\\
81	0.941371158392435\\
82	0.941371158392435\\
83	0.944208037825059\\
84	0.945153664302601\\
85	0.948936170212766\\
86	0.949881796690307\\
87	0.950827423167849\\
88	0.952718676122931\\
89	0.954609929078014\\
90	0.955555555555556\\
91	0.956501182033097\\
92	0.95839243498818\\
93	0.95839243498818\\
94	0.959338061465721\\
95	0.962174940898345\\
96	0.963120567375886\\
97	0.963120567375886\\
98	0.963120567375886\\
99	0.964066193853428\\
100	0.965011820330969\\
101	0.965011820330969\\
102	0.966903073286052\\
103	0.966903073286052\\
104	0.969739952718676\\
105	0.969739952718676\\
106	0.969739952718676\\
107	0.969739952718676\\
108	0.969739952718676\\
109	0.969739952718676\\
110	0.969739952718676\\
111	0.969739952718676\\
112	0.970685579196217\\
113	0.971631205673759\\
114	0.973522458628842\\
115	0.974468085106383\\
116	0.975413711583924\\
117	0.976359338061466\\
118	0.976359338061466\\
119	0.976359338061466\\
120	0.976359338061466\\
121	0.976359338061466\\
122	0.977304964539007\\
123	0.977304964539007\\
124	0.978250591016548\\
125	0.978250591016548\\
126	0.97919621749409\\
127	0.97919621749409\\
128	0.97919621749409\\
129	0.97919621749409\\
130	0.980141843971631\\
131	0.980141843971631\\
132	0.980141843971631\\
133	0.980141843971631\\
134	0.981087470449173\\
135	0.981087470449173\\
136	0.981087470449173\\
137	0.981087470449173\\
138	0.981087470449173\\
139	0.981087470449173\\
140	0.981087470449173\\
141	0.981087470449173\\
142	0.981087470449173\\
143	0.981087470449173\\
144	0.981087470449173\\
145	0.981087470449173\\
146	0.981087470449173\\
147	0.982033096926714\\
148	0.982033096926714\\
149	0.983924349881797\\
150	0.983924349881797\\
151	0.983924349881797\\
152	0.983924349881797\\
153	0.983924349881797\\
154	0.983924349881797\\
155	0.983924349881797\\
156	0.983924349881797\\
157	0.983924349881797\\
158	0.983924349881797\\
159	0.983924349881797\\
160	0.983924349881797\\
161	0.983924349881797\\
162	0.983924349881797\\
163	0.984869976359338\\
164	0.984869976359338\\
165	0.984869976359338\\
166	0.984869976359338\\
167	0.984869976359338\\
168	0.985815602836879\\
169	0.986761229314421\\
170	0.986761229314421\\
171	0.986761229314421\\
172	0.986761229314421\\
173	0.986761229314421\\
174	0.986761229314421\\
175	0.987706855791962\\
176	0.987706855791962\\
177	0.987706855791962\\
178	0.987706855791962\\
179	0.987706855791962\\
180	0.987706855791962\\
181	0.987706855791962\\
182	0.988652482269504\\
183	0.988652482269504\\
184	0.988652482269504\\
185	0.989598108747045\\
186	0.989598108747045\\
187	0.989598108747045\\
188	0.989598108747045\\
189	0.989598108747045\\
190	0.989598108747045\\
191	0.989598108747045\\
};
\addlegendentry{$T=0.5$}
\end{axis}
\end{tikzpicture}%
}
\subcaption{FedCau test accuracy $M=50$, $F_f = 10$.}
\label{subfig: CSminiACC}
\end{minipage}
\hspace{0.02\columnwidth}
\begin{minipage}{0.4\columnwidth}

{\scriptsize\definecolor{amber}{rgb}{1.0, 0.49, 0.0}
\definecolor{taupe}{rgb}{0.28, 0.24, 0.2}
\definecolor{tealgreen}{rgb}{0.0, 0.51, 0.5}
\begin{tikzpicture}
\begin{axis}[%
width=0.8\columnwidth,
height=0.5\columnwidth,
at={(0,0)},
scale only axis,
xmin=0, 
xmax=200,
xlabel={Communication iteration $k$},
xtick style={color=black},
ymin=0.08,
ymax=1.1,
ytick={0,0.2, 0.4, 0.6, 0.8, 1},
grid style={dashed},
ymajorgrids,
grid=both,
ylabel near ticks,
ylabel={Test accuracy},
axis background/.style={fill=white},
legend style={at={(0.99,0)}, anchor=south east, legend cell align=left, font = \tiny}
]
\addplot [color=taupe, dashed, line width=2pt] 
  table[row sep=crcr]{%
1	-0.0572104018912529\\
2	-0.0250591016548463\\
3	0.00898345153664304\\
4	0.0534278959810874\\
5	0.0997635933806147\\
6	0.146099290780142\\
7	0.185815602836879\\
8	0.229314420803783\\
9	0.264302600472813\\
10	0.307801418439716\\
11	0.339007092198582\\
12	0.387234042553191\\
13	0.416548463356974\\
14	0.447754137115839\\
15	0.47612293144208\\
16	0.498817966903073\\
17	0.52434988179669\\
18	0.544208037825059\\
19	0.5725768321513\\
20	0.596217494089834\\
21	0.613238770685579\\
22	0.633096926713948\\
23	0.652955082742317\\
24	0.670921985815603\\
25	0.686997635933806\\
26	0.700236406619385\\
27	0.712529550827423\\
28	0.721985815602837\\
29	0.730496453900709\\
30	0.741843971631206\\
31	0.747517730496454\\
32	0.755082742316785\\
33	0.761702127659574\\
34	0.768321513002364\\
35	0.774940898345154\\
36	0.782505910165485\\
37	0.788179669030733\\
38	0.794799054373522\\
39	0.802364066193853\\
40	0.808037825059102\\
41	0.817494089834515\\
42	0.824113475177305\\
43	0.828841607565012\\
44	0.831678486997636\\
45	0.833569739952719\\
46	0.837352245862884\\
47	0.846808510638298\\
48	0.854373522458629\\
49	0.859101654846336\\
50	0.862884160756501\\
51	0.866666666666667\\
52	0.872340425531915\\
53	0.873286052009456\\
54	0.87612293144208\\
55	0.879905437352246\\
56	0.884633569739953\\
57	0.884633569739953\\
58	0.890307328605201\\
59	0.897872340425532\\
60	0.902600472813239\\
61	0.906382978723404\\
62	0.91016548463357\\
63	0.911111111111111\\
64	0.913002364066194\\
65	0.913002364066194\\
66	0.913947990543735\\
67	0.916784869976359\\
68	0.919621749408983\\
69	0.923404255319149\\
70	0.926241134751773\\
71	0.928132387706856\\
72	0.929078014184397\\
73	0.93096926713948\\
74	0.931914893617021\\
75	0.932860520094563\\
76	0.933806146572104\\
77	0.933806146572104\\
78	0.935697399527187\\
79	0.93758865248227\\
80	0.938534278959811\\
81	0.941371158392435\\
82	0.941371158392435\\
83	0.944208037825059\\
84	0.945153664302601\\
85	0.948936170212766\\
86	0.949881796690307\\
87	0.950827423167849\\
88	0.952718676122931\\
89	0.954609929078014\\
90	0.955555555555556\\
91	0.956501182033097\\
92	0.95839243498818\\
93	0.95839243498818\\
94	0.959338061465721\\
95	0.962174940898345\\
96	0.963120567375886\\
97	0.963120567375886\\
98	0.963120567375886\\
99	0.964066193853428\\
100	0.965011820330969\\
101	0.965011820330969\\
102	0.966903073286052\\
103	0.966903073286052\\
104	0.969739952718676\\
105	0.969739952718676\\
106	0.969739952718676\\
107	0.969739952718676\\
108	0.969739952718676\\
109	0.969739952718676\\
110	0.969739952718676\\
111	0.969739952718676\\
112	0.970685579196217\\
113	0.971631205673759\\
114	0.973522458628842\\
115	0.974468085106383\\
116	0.975413711583924\\
117	0.976359338061466\\
118	0.976359338061466\\
119	0.976359338061466\\
120	0.976359338061466\\
121	0.976359338061466\\
122	0.977304964539007\\
123	0.977304964539007\\
124	0.978250591016548\\
125	0.978250591016548\\
126	0.97919621749409\\
127	0.97919621749409\\
128	0.97919621749409\\
129	0.97919621749409\\
130	0.980141843971631\\
131	0.980141843971631\\
132	0.980141843971631\\
133	0.980141843971631\\
134	0.981087470449173\\
135	0.981087470449173\\
136	0.981087470449173\\
137	0.981087470449173\\
138	0.981087470449173\\
139	0.981087470449173\\
140	0.981087470449173\\
141	0.981087470449173\\
142	0.981087470449173\\
143	0.981087470449173\\
144	0.981087470449173\\
145	0.981087470449173\\
146	0.981087470449173\\
147	0.982033096926714\\
148	0.982033096926714\\
149	0.983924349881797\\
150	0.983924349881797\\
151	0.983924349881797\\
152	0.983924349881797\\
153	0.983924349881797\\
154	0.983924349881797\\
155	0.983924349881797\\
156	0.983924349881797\\
157	0.983924349881797\\
158	0.983924349881797\\
159	0.983924349881797\\
160	0.983924349881797\\
161	0.983924349881797\\
162	0.983924349881797\\
163	0.984869976359338\\
164	0.984869976359338\\
165	0.984869976359338\\
166	0.984869976359338\\
167	0.984869976359338\\
168	0.985815602836879\\
169	0.986761229314421\\
170	0.986761229314421\\
171	0.986761229314421\\
172	0.986761229314421\\
173	0.986761229314421\\
174	0.986761229314421\\
175	0.987706855791962\\
176	0.987706855791962\\
177	0.987706855791962\\
178	0.987706855791962\\
179	0.987706855791962\\
180	0.987706855791962\\
181	0.987706855791962\\
182	0.988652482269504\\
183	0.988652482269504\\
184	0.988652482269504\\
185	0.989598108747045\\
186	0.989598108747045\\
187	0.989598108747045\\
188	0.989598108747045\\
189	0.989598108747045\\
190	0.989598108747045\\
191	0.989598108747045\\
};
\addlegendentry{FedCau}

\addplot [color=black] 
  table[row sep=crcr]{%
1	-0.00236406619385332\\
2	0.0392434988179668\\
3	0.0647754137115839\\
4	0.0761229314420804\\
5	0.11016548463357\\
6	0.113947990543735\\
7	0.133806146572104\\
8	0.171631205673759\\
9	0.175413711583924\\
10	0.199054373522459\\
11	0.205673758865248\\
12	0.236879432624113\\
13	0.242553191489362\\
14	0.273758865248227\\
15	0.27565011820331\\
16	0.312529550827423\\
17	0.315366430260047\\
18	0.345626477541371\\
19	0.348463356973995\\
20	0.376832151300236\\
21	0.380614657210402\\
22	0.408037825059102\\
23	0.438297872340425\\
24	0.44113475177305\\
25	0.448699763593381\\
26	0.451536643026005\\
27	0.45531914893617\\
28	0.481796690307329\\
29	0.484633569739953\\
30	0.48936170212766\\
31	0.492198581560284\\
32	0.494089834515366\\
33	0.495981087470449\\
34	0.502600472813239\\
35	0.51016548463357\\
36	0.511111111111111\\
37	0.515839243498818\\
38	0.536643026004728\\
39	0.540425531914894\\
40	0.543262411347518\\
41	0.546099290780142\\
42	0.549881796690307\\
43	0.554609929078014\\
44	0.557446808510638\\
45	0.561229314420804\\
46	0.565957446808511\\
47	0.569739952718676\\
48	0.575413711583924\\
49	0.578250591016549\\
50	0.599054373522459\\
51	0.602836879432624\\
52	0.610401891252955\\
53	0.619858156028369\\
54	0.621749408983452\\
55	0.624586288416076\\
56	0.626477541371158\\
57	0.629314420803783\\
58	0.656737588652482\\
59	0.658628841607565\\
60	0.660520094562648\\
61	0.667139479905437\\
62	0.66903073286052\\
63	0.671867612293144\\
64	0.674704491725768\\
65	0.676595744680851\\
66	0.680378250591017\\
67	0.680378250591017\\
68	0.695508274231678\\
69	0.703073286052009\\
70	0.71725768321513\\
71	0.719148936170213\\
72	0.721040189125296\\
73	0.722931442080378\\
74	0.725768321513002\\
75	0.726713947990544\\
76	0.726713947990544\\
77	0.729550827423168\\
78	0.729550827423168\\
79	0.733333333333333\\
80	0.735224586288416\\
81	0.74468085106383\\
82	0.747517730496454\\
83	0.749408983451537\\
84	0.751300236406619\\
85	0.752245862884161\\
86	0.753191489361702\\
87	0.755082742316785\\
88	0.760756501182033\\
89	0.760756501182033\\
90	0.76548463356974\\
91	0.768321513002364\\
92	0.771158392434988\\
93	0.773049645390071\\
94	0.774940898345154\\
95	0.776832151300236\\
96	0.776832151300236\\
97	0.787234042553192\\
98	0.787234042553192\\
99	0.794799054373522\\
100	0.796690307328605\\
101	0.801418439716312\\
102	0.805200945626477\\
103	0.808037825059102\\
104	0.808983451536643\\
105	0.808983451536643\\
106	0.808983451536643\\
107	0.820330969267139\\
108	0.821276595744681\\
109	0.821276595744681\\
110	0.821276595744681\\
111	0.822222222222222\\
112	0.822222222222222\\
113	0.823167848699764\\
114	0.823167848699764\\
115	0.825059101654846\\
116	0.826950354609929\\
117	0.82789598108747\\
118	0.830732860520095\\
119	0.844917257683215\\
120	0.845862884160757\\
121	0.846808510638298\\
122	0.846808510638298\\
123	0.847754137115839\\
124	0.848699763593381\\
125	0.849645390070922\\
126	0.849645390070922\\
127	0.850591016548463\\
128	0.850591016548463\\
129	0.850591016548463\\
130	0.850591016548463\\
131	0.850591016548463\\
132	0.850591016548463\\
133	0.851536643026005\\
134	0.852482269503546\\
135	0.853427895981087\\
136	0.85531914893617\\
137	0.86193853427896\\
138	0.862884160756501\\
139	0.871394799054374\\
140	0.873286052009456\\
141	0.874231678486998\\
142	0.87612293144208\\
143	0.87612293144208\\
144	0.877068557919622\\
145	0.877068557919622\\
146	0.877068557919622\\
147	0.878959810874705\\
148	0.878959810874705\\
149	0.879905437352246\\
150	0.880851063829787\\
151	0.883687943262411\\
152	0.884633569739953\\
153	0.885579196217494\\
154	0.885579196217494\\
155	0.885579196217494\\
156	0.887470449172577\\
157	0.894089834515366\\
158	0.898817966903073\\
159	0.898817966903073\\
160	0.900709219858156\\
161	0.901654846335697\\
162	0.902600472813239\\
163	0.902600472813239\\
164	0.902600472813239\\
165	0.904491725768321\\
166	0.905437352245863\\
167	0.905437352245863\\
168	0.906382978723404\\
169	0.906382978723404\\
170	0.906382978723404\\
171	0.906382978723404\\
172	0.907328605200946\\
173	0.907328605200946\\
174	0.907328605200946\\
175	0.909219858156028\\
176	0.909219858156028\\
177	0.91016548463357\\
178	0.91016548463357\\
179	0.914893617021277\\
180	0.914893617021277\\
181	0.914893617021277\\
182	0.914893617021277\\
183	0.914893617021277\\
184	0.914893617021277\\
185	0.915839243498818\\
186	0.915839243498818\\
187	0.915839243498818\\
188	0.915839243498818\\
189	0.915839243498818\\
190	0.915839243498818\\
191	0.916784869976359\\
};
\addlegendentry{FedAvg}

\end{axis}
\end{tikzpicture}%
}
\subcaption{FedCau vs FedAvg test accuracy, $T = 0.5$.}
\label{subfig: CSminiAcc_FedAvg1}
\end{minipage}
\hspace{0.02\columnwidth}
\begin{minipage}{0.4\columnwidth}

{\scriptsize\definecolor{amber}{rgb}{1.0, 0.49, 0.0}
\definecolor{taupe}{rgb}{0.28, 0.24, 0.2}
\definecolor{tealgreen}{rgb}{0.0, 0.51, 0.5}
\begin{tikzpicture}
\begin{axis}[%
width=0.8\columnwidth,
height=0.5\columnwidth,
at={(0,0)},
scale only axis,
xmin=0, 
xmax=127,
xlabel={Communication iteration $k$},
xtick style={color=black},
ymin=0.08,
ymax=1.1,
ytick={0,0.2, 0.4, 0.6, 0.8, 1},
grid style={dashed},
ymajorgrids,
grid=both,
ylabel near ticks,
ylabel={Test accuracy},
axis background/.style={fill=white},
legend style={at={(0.99,0)}, anchor=south east, legend cell align=left, font = \tiny}
]

\addplot [color=amber, mark=otimes, mark options={scale=0.85}] 
  table[row sep=crcr]{%
1	-0.00520094562647744\\
2	0.00520094562647755\\
4	0.0260047281323877\\
8	0.0884160756501182\\
12	0.183924349881797\\
16	0.298345153664303\\
20	0.448699763593381\\
24	0.589598108747045\\
28	0.695508274231678\\
32	0.785342789598109\\
36	0.848699763593381\\
40	0.888416075650118\\
44	0.916784869976359\\
47	0.928132387706856\\
53	0.95177304964539\\
65	0.973522458628842\\
81	0.984869976359338\\
126	0.989598108747045\\
};
\addlegendentry{FedCau}

\addplot [color=black]
  table[row sep=crcr]{%
1	-0.361702127659574\\
2	-0.327659574468085\\
3	-0.300236406619385\\
4	-0.280378250591017\\
5	-0.233096926713948\\
6	-0.188652482269503\\
7	-0.164066193853428\\
8	-0.111111111111111\\
9	-0.0855791962174941\\
10	-0.0647754137115839\\
11	-0.00520094562647744\\
12	0.0269503546099291\\
13	0.0817966903073286\\
14	0.132860520094563\\
15	0.192434988179669\\
16	0.2274231678487\\
17	0.269976359338061\\
18	0.29645390070922\\
19	0.320094562647754\\
20	0.339007092198582\\
21	0.391016548463357\\
22	0.408983451536643\\
23	0.451536643026005\\
24	0.465721040189125\\
25	0.513947990543735\\
26	0.531914893617021\\
27	0.561229314420804\\
28	0.575413711583924\\
29	0.599054373522459\\
30	0.617966903073286\\
31	0.64160756501182\\
32	0.652955082742317\\
33	0.677541371158392\\
34	0.707801418439716\\
35	0.735224586288416\\
36	0.741843971631206\\
37	0.749408983451537\\
38	0.763593380614657\\
39	0.774940898345154\\
40	0.783451536643026\\
41	0.790070921985816\\
42	0.791962174940898\\
43	0.797635933806147\\
44	0.798581560283688\\
45	0.801418439716312\\
46	0.820330969267139\\
47	0.832624113475177\\
48	0.845862884160757\\
49	0.848699763593381\\
50	0.849645390070922\\
51	0.852482269503546\\
52	0.862884160756501\\
53	0.873286052009456\\
54	0.877068557919622\\
55	0.879905437352246\\
56	0.886524822695036\\
57	0.894089834515366\\
58	0.898817966903073\\
59	0.904491725768321\\
60	0.908274231678487\\
61	0.913947990543735\\
62	0.917730496453901\\
63	0.920567375886525\\
64	0.922458628841608\\
65	0.925295508274232\\
66	0.926241134751773\\
67	0.927186761229314\\
68	0.928132387706856\\
69	0.928132387706856\\
70	0.931914893617021\\
71	0.935697399527187\\
72	0.938534278959811\\
73	0.940425531914894\\
74	0.942316784869976\\
75	0.943262411347518\\
76	0.945153664302601\\
77	0.946099290780142\\
78	0.946099290780142\\
79	0.947990543735225\\
80	0.948936170212766\\
81	0.948936170212766\\
82	0.948936170212766\\
83	0.949881796690307\\
84	0.95177304964539\\
85	0.954609929078014\\
86	0.957446808510638\\
87	0.95839243498818\\
88	0.95839243498818\\
89	0.959338061465721\\
90	0.959338061465721\\
91	0.959338061465721\\
92	0.959338061465721\\
93	0.959338061465721\\
94	0.960283687943262\\
95	0.961229314420804\\
96	0.962174940898345\\
97	0.963120567375886\\
98	0.964066193853428\\
99	0.964066193853428\\
100	0.965957446808511\\
101	0.965957446808511\\
102	0.966903073286052\\
103	0.966903073286052\\
104	0.966903073286052\\
105	0.966903073286052\\
106	0.966903073286052\\
107	0.966903073286052\\
108	0.966903073286052\\
109	0.966903073286052\\
110	0.966903073286052\\
111	0.966903073286052\\
112	0.968794326241135\\
113	0.968794326241135\\
114	0.969739952718676\\
115	0.969739952718676\\
116	0.969739952718676\\
117	0.9725768321513\\
118	0.973522458628842\\
119	0.973522458628842\\
120	0.973522458628842\\
121	0.974468085106383\\
122	0.974468085106383\\
123	0.974468085106383\\
124	0.975413711583924\\
125	0.975413711583924\\
126	0.977304964539007\\
};
\addlegendentry{FedAvg}

\end{axis}
\end{tikzpicture}%
}
\subcaption{FedCau vs FedAvg test accuracy, $T = 0.9$.}
\label{subfig: CSminiAcc_FedAvg2}
\end{minipage}
\begin{minipage}{0.4\columnwidth}

{\scriptsize\definecolor{amber}{rgb}{1.0, 0.49, 0.0}
\definecolor{taupe}{rgb}{0.28, 0.24, 0.2}
\definecolor{tealgreen}{rgb}{0.0, 0.51, 0.5}
\begin{tikzpicture}
\begin{axis}[%
width=0.8\columnwidth,
height=0.5\columnwidth,
at={(0,0)},
scale only axis,
xmin=0, 
xmax=100,
xlabel={Communication iteration $k$},
xtick style={color=black},
ymin=0.08,
ymax=1.1,
ytick={0.28, 0.4, 0.6, 0.8, 1},
grid style={dashed},
ymajorgrids,
grid=both,
ylabel near ticks,
ylabel={Test accuracy},
axis background/.style={fill=white},
legend style={at={(0.99,0)}, anchor=south east, legend cell align=left, font = \tiny}
]
\addplot [color=tealgreen, mark=halfcircle* ]
  table[row sep=crcr]{%
1	0.433569739952719\\
6	0.549881796690307\\
11	0.640661938534279\\
18	0.707801418439716\\
25	0.764539007092199\\
31	0.808983451536643\\
37	0.842080378250591\\
48	0.872340425531915\\
56	0.892198581560284\\
65	0.91016548463357\\
72	0.920567375886525\\
80	0.93096926713948\\
87	0.938534278959811\\
92	0.945153664302601\\
96	0.946099290780142\\
};
\addlegendentry{FedCau}

\addplot [color=black]
  table[row sep=crcr]{%
1	0.293617021276596\\
2	0.324822695035461\\
3	0.352245862884161\\
4	0.371158392434988\\
5	0.393853427895981\\
6	0.411820330969267\\
7	0.424113475177305\\
8	0.438297872340425\\
9	0.460047281323877\\
10	0.47612293144208\\
11	0.492198581560284\\
12	0.501654846335697\\
13	0.522458628841608\\
14	0.535697399527187\\
15	0.55177304964539\\
16	0.565957446808511\\
17	0.587706855791962\\
18	0.595271867612293\\
19	0.609456264775414\\
20	0.617021276595745\\
21	0.628368794326241\\
22	0.640661938534279\\
23	0.651063829787234\\
24	0.658628841607565\\
25	0.666193853427896\\
26	0.670921985815603\\
27	0.681323877068558\\
28	0.688888888888889\\
29	0.702127659574468\\
30	0.71063829787234\\
31	0.719148936170213\\
32	0.725768321513002\\
33	0.735224586288416\\
34	0.739007092198582\\
35	0.747517730496454\\
36	0.752245862884161\\
37	0.753191489361702\\
38	0.75886524822695\\
39	0.766430260047281\\
40	0.774940898345154\\
41	0.776832151300236\\
42	0.780614657210402\\
43	0.785342789598109\\
44	0.791962174940898\\
45	0.798581560283688\\
46	0.800472813238771\\
47	0.808983451536643\\
48	0.81371158392435\\
49	0.816548463356974\\
50	0.819385342789598\\
51	0.823167848699764\\
52	0.826004728132388\\
53	0.830732860520095\\
54	0.83451536643026\\
55	0.84113475177305\\
56	0.845862884160757\\
57	0.847754137115839\\
58	0.849645390070922\\
59	0.852482269503546\\
60	0.858156028368794\\
61	0.862884160756501\\
62	0.864775413711584\\
63	0.867612293144208\\
64	0.871394799054374\\
65	0.87612293144208\\
66	0.878959810874705\\
67	0.879905437352246\\
68	0.881796690307329\\
69	0.884633569739953\\
70	0.888416075650118\\
71	0.891252955082742\\
72	0.894089834515366\\
73	0.897872340425532\\
74	0.900709219858156\\
75	0.901654846335697\\
76	0.906382978723404\\
77	0.907328605200946\\
78	0.91016548463357\\
79	0.912056737588652\\
80	0.913947990543735\\
81	0.914893617021277\\
82	0.916784869976359\\
83	0.919621749408983\\
84	0.920567375886525\\
85	0.923404255319149\\
86	0.925295508274232\\
87	0.926241134751773\\
88	0.927186761229314\\
89	0.928132387706856\\
90	0.929078014184397\\
91	0.93096926713948\\
92	0.933806146572104\\
93	0.936643026004728\\
94	0.936643026004728\\
95	0.936643026004728\\
96	0.936643026004728\\
};
\addlegendentry{FedAvg}

\end{axis}
\end{tikzpicture}%
}
\subcaption{FedCau vs FedAvg test accuracy, $T = 1.1$.}
\label{subfig: CSminiAcc_FedAvg3}
\end{minipage}
\hspace{0.02\columnwidth}
\begin{minipage}{0.4\columnwidth}
{\scriptsize\definecolor{upmaroon}{rgb}{0.48, 0.07, 0.07}
\definecolor{wildstrawberry}{rgb}{1.0, 0.26, 0.64}
\definecolor{ultramarine}{rgb}{0.07, 0.04, 0.56}
\definecolor{uclablue}{rgb}{0.33, 0.41, 0.58}
\begin{tikzpicture}
\begin{axis}[%
width=0.8\columnwidth,
height=0.5\columnwidth,
at={(0,0)},
scale only axis,
xmin=0, 
xmax=200,
xlabel={Fairness factor $F_f$},
xtick style={color=black},
ymin=0.32,
ymax=0.4,
ytick={0.32, 0.34, 0.36},
grid style={dashed},
ymajorgrids,
grid=both,
ylabel near ticks,
ylabel={$C(k_c)$},
axis background/.style={fill=white},
legend style={at={(0.99,0)}, anchor=south east, legend cell align=left, font = \tiny}
]
\addplot [color=ultramarine, mark = *, mark options={scale=0.9}, dotted]
  table[row sep=crcr]{%
3 0.3834\\
10 0.3697\\
20	0.3454\\
40	0.3435\\
50	0.3481\\
70	0.3535\\
100	0.351\\
120	0.355\\
150	0.3665\\
180	0.374\\
200	0.3830\\
};

\end{axis}
\end{tikzpicture}%
}

\subcaption{Causal iteration-cost.}
\label{subfig: CSminiC1}
\end{minipage}
\hspace{0.02\columnwidth}
\begin{minipage}{0.4\columnwidth}

{\scriptsize\definecolor{upmaroon}{rgb}{0.48, 0.07, 0.07}
\definecolor{wildstrawberry}{rgb}{1.0, 0.26, 0.64}
\definecolor{ultramarine}{rgb}{0.07, 0.04, 0.56}
\definecolor{uclablue}{rgb}{0.33, 0.41, 0.58}
\begin{tikzpicture}
\begin{axis}[%
width=0.8\columnwidth,
height=0.5\columnwidth,
at={(0,0)},
scale only axis,
xmin=0, 
xmax=200,
xlabel={Fairness factor $F_f$},
xtick style={color=black},
ymin=0.9,
ymax=1,
grid style={dashed},
ymajorgrids,
grid=both,
ylabel near ticks,
ylabel={Test accuracy},
axis background/.style={fill=white},
legend style={at={(0.99,0)}, anchor=south east, legend cell align=left, font = \tiny}
]
\addplot [color=ultramarine, mark = *, mark options={scale=0.9}, dotted]
  table[row sep=crcr]{%
3 0.9877\\ 
10 0.9811\\
20	0.9631\\
40	0.9697\\
50	0.9650\\
70	0.9423\\
100	0.941\\
120	0.94\\
150	0.9253\\
180	0.921\\
200	0.92\\
};

\end{axis}
\end{tikzpicture}%
}
\subcaption{Test accuracy.}
\label{subfig: CSminiACC1}
\end{minipage}

\caption{{\color{black}Illustration of the mini-batch FedCau update of Algorithm~\ref{alg: asynchronous alg} for CSMA/CA, $p_x = 1$, $p_r = 0.01$, and $M=50$. a) $C(k_c)$ for~$F_f = 10$, $T = 0.5, 0.9, 1.1 $s, b) Test accuracy for~$F_f = 10$, $T = 0.5, 0.9, 1.1 $s, c) Test accuracy for~$T = 0.5$s FedCau vs FedAvg, d) Test accuracy for~$T = 0.9$s FedCau vs FedAvg, e) Test accuracy for~$T = 1.1$s FedCau vs FedAvg, f) Iteration-cost function $C(k_c)$ for $T = 1.1$s vs different fairness factor $F_f$, and g) Test accuracy for $T = 1.1$s vs $F_f$.}
}
\label{fig: mini1}
\end{figure*}
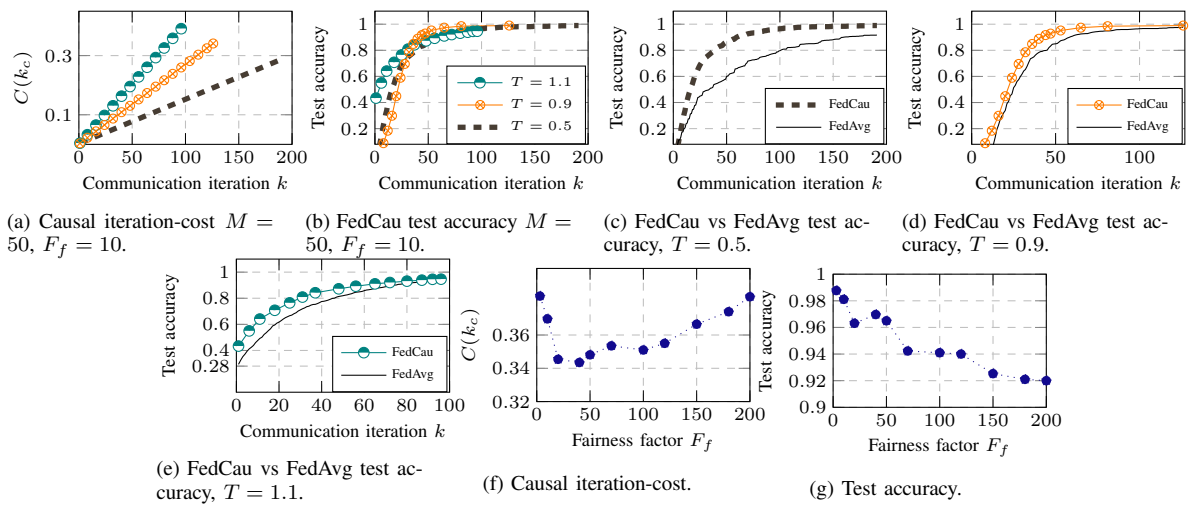

Finally, in OFDMA, we consider uplink transmissions in a single-cell wireless system with $s = 1,\ldots, S_c$ orthogonal subchannels~\cite{8086180}. Let $h_{l}^s$, $p_{l}^s$ be the channel gain and the transmit power of link $l$ on subchannel $s$ by which worker $j$ sends its local parameters to the master node. Therefore, the signal-to-noise ratio (SNR) for the uplink is defined $\text{SNR}(p_{l}^s, h_{l}^s):={p_{l}^s h_{l}^s}/{\sigma_{l}^{s}}$. 
 The corresponding data rate (bps/Hz) is as $R_p (\text{SNR}) = \sum _{s=1}^{S_c}\text{log}_2( 1+\text{SNR}(p_{l}^s, h_{l}^s))$. The master node randomly decides at each iteration $k$ which worker should use which subchannel link, and the remaining workers will not participate in the parameter uploading.

\section{Numerical Results}\label{sec: numerical_results}
In this section, we illustrate our results from the previous sections. We numerically show the extensive impact of the iteration costs when running the FedAvg algorithm~\eqref{eq: local trains} training problem over a wireless network. We use a network with $M$ workers and simulation to implement slotted-ALOHA, CSMA/CA (both with binary exponential backoff), and the OFDMA. In each of these networks, we apply our proposed Algorithms~\ref{alg: synchronous alg},~\ref{alg: asynchronous alg},~and~\ref{alg: Nonconvex alg}. {\color{black}We train the FedCau by the well-known MNIST dataset with non-iid distribution among workers and the CIFAR-10 dataset with both iid and non-iid cases}. {\color{black}For the non-iid implementation, we first sort the dataset w.r.t. the label numbers of~$y_i= i$, where $i \in \{0,1,\ldots, 9 \}$, where $i$ is the index of each data sample with size~$|D_i|$. Moreover, in the MNIST dataset, the labels are the same as the digits, while in CIFAR-10, the labels demonstrate airplane, automobile, bird, cat, deer, dog, frog, horse, ship, and truck.
Afterward, we assign an equal portion of data to each worker~$j \in \{1,\ldots, M\}$, starting from the beginning of the sorted dataset. According to the size of each dataset, CIFAR-10 with $50000$ and MNIST with $60000$ data samples, the data portion of every class in the datasets assigned to each worker is different.} Finally, we apply our proposed FedCau on top of existing methods from the literature, such as top-$q$ and LAQ.

\subsection{Simulation Settings}
First, we consider solving a convex regression problem over a wireless network using a real-world dataset.
To this end, we extract a binary dataset from MNIST (hand-written digits) by keeping samples of digits 0 and 1 and then setting their labels to -1 and +1, respectively. We then randomly split the resulting dataset of 12600 samples among $M$ workers, each having $\{(\bx_{ij}, y_{ij})\}$, where  $\bx_{ij} \in \R ^{784}$ is a data sample $i$, and a vectorized image at node $j \in [M]$ with corresponding digit label $y_{ij} \in \{-1, +1\}$. We use the loss function \cite{koh2007interior}
\begin{equation}
    f(\bw) = \sum_{j =1}^M \rho_j \sum_{i =1}^{|D_j|} \frac{1}{|D_j|} \log\left(1 + e^{-\bw^T\bx_{ij} y_{ij}}\right),
\end{equation}
where we consider that each worker $j \in [M]$ has $|D_j| = |D_i| = |D|/M, \forall i, j \in [M]$.

Second, we consider a non-convex image classification problem with the workers using convolutional neural networks (CNNs) with a cross-entropy loss function. The architecture of the CNN consists of a convolutional layer,
Conv2D(32, (3, 3)), a MaxPooling2D layer with a pool size of (2, 2), a Flatten layer, two Dense (fully connected) layers with size 64 and 10, and a final layer that produces probability distributions over 10 classes of the CIFAR-10 dataset. Overall, the CNN has 462410 parameters.

 We implement the network with $M$ workers performing local updates of $\bw_k^j, \forall j \in [M]$ and imposing computation latency of $\ell_{2,k}$ to the system. We assume a synchronous network in which all workers start the local iteration of $\bw_k^j$ simultaneously right after receiving $\bw_{k-1}$. Note that the latency counting of $c_k$ at each iteration~$k$ starts from the beginning of the local iterations until the uplink process is complete. Regarding the computation latency, we consider $\nu_k \in [10^6, 3\times10^9]$ cycles/s, and $a_k = [10,30]\time 16$ cycle/sample for $ k=1, \ldots, M$. In slotted-ALOHA, we consider a capacity of one packet per slot and a slot duration of $1$ ms. In CSMA/CA, we consider the packet length of $10$~kb with a packet rate of $1$~k packets per second, leading to a total rate of $1$~Mbps. We set the duration of SIFS, DIFS, and each time slot to be $10~\micro$s, $50~\micro$s, and $10~\micro$s respectively~\cite{IEEE802.11Standard} and run the network for $1000$ times. In the OFDMA setup, we consider the uplink in a single cell system with the coverage radius of $r_c = 1$ Km. There are $L_p$ cellular links on $S_c$ subchannels. We model the subchannel power gain $h_l^s = \zeta/r^3$, following the Rayleigh fading, where $\zeta$ has an exponential distribution with unitary mean. We consider the noise power in each subchannel as $-170$ dBm/Hz and the maximum transmit power of each link as $23$ dBm. We assume that $S_c=64$ subchannels, the total bandwidth of $10$ MHz, and the subchannel bandwidth of $150$ KHz. We define $c_k$ as the latency caused by the slowest worker to send the local parameters to the master node.
%

\subsection{Performance of FedCau Update from Algorithms~\ref{alg: synchronous alg},~\ref{alg: asynchronous alg} and Non-causal Approach}
Fig.~\ref{fig: Batch1k} characterizes the non-causal and causal behaviors along with the performance of FedCau update of Algorithms~\ref{alg: synchronous alg} and \ref{alg: asynchronous alg} for slotted-ALOHA and CSMA/CA protocols. The general network setup has $M=100$, $p_x = 1$, $p_r = 0.2$, and the mini-batch time budget of $T = 0.3$s. We observe that while the behavior of $f(\bw_k)$ is similar across the protocols in Fig.~\ref{subfig: f}, the  iteration-cost function~$C(K)$ of the batch update for slotted-ALOHA is much larger among all the setups in Fig.~\ref{subfig: C}. This behavior affects the multi-objective function $G(K)$ in Fig.~\ref{subfig: G} and causes an earlier stop. However, the test accuracy is not sacrificed, as shown in Fig.~\ref{subfig: AC}. From Fig.~\ref{fig: Batch1k}, we conclude that the batch update of Algorithm~\ref{alg: synchronous alg} satisfies the causal setting and preserves the test accuracy while optimizing both the loss function~$f(\bw_k)$ and the latency over the communication protocols.


{\color{black}
Fig.~\ref{fig: BetaPerf} characterizes the effect of $\beta$ on the performance of batch FedCau update of Algorithm~\ref{alg: synchronous alg} with $M=50, 100$ and CSMA/CA protocol for parameter upload. Fig.~\ref{subfig: ckvsbeta} shows that $C(k_c)$ decreases while $\beta$ takes the values between $(0,1)$. This decreasing behavior is a valid result since the higher values of $\beta$ increase the effect of the term $C(K)$ in scalarized version~\eqref{eq: cost-efficient-distributed-optim-3}. Since $C(K)$ is an increasing function of $K$, the higher values of $C(K)$ result in stopping at the smaller causal iterations, called $k_c$. 
Finally, Fig.~\ref{subfig: accvsbeta} demonstrates the test accuracy we achieve while changing $\beta$. Since $k_c$ decreases as $\beta$ increases, the corresponding test accuracy decreases. Therefore, choosing $\beta \in [0.2,0.5]$ gives us a lower causal iteration cost and sub-optimal test accuracy.

} 

{\color{black}Fig.~\ref{fig: mini1} represents the mini-batch FedCau update of Algorithm~\ref{alg: asynchronous alg} and the FedAvg baseline for CSMA/CA with $p_x = 1$, $p_r = 0.01$ and $M = 50$. Figs.~\ref{subfig: CSminiC}-\ref{subfig: CSminiACC} show the results for $M = 50$, with $T = 0.5, 0.9, 1.1$s. Fig.~\ref{subfig: CSminiC} highlights that with a smaller time budget,~$C(k_c)$ decreases, while Fig.~\ref{subfig: CSminiACC} shows the similarity in the test accuracy. Fig.~\ref{subfig: CSminiAcc_FedAvg1}-\ref{subfig: CSminiAcc_FedAvg3} compare the test accuracy of FedCau in Algorithm~2 with the FedAvg by assigning the time budget~$T=0.5, 0.9, 1.1$ respectively. For the time budget~$T = 0.5, 0.9, 1.1$s, the test accuracy of FedAvg is lower than the results of mini-batch FedCau update of Algorithm~2 with a similar time budget~$T$. These results highlight the role of~$F_f$ combined with~$T$, where $F_f$ ensures participation fairness, especially for the smaller $T$, such as $T=0.5$. Therefore, with the equal~$T$, the FedCau in Algorithm~2 outperforms FedAvg in test accuracy and fairness in worker participation. Figs.~\ref{subfig: CSminiC1}-\ref{subfig: CSminiACC1} reveal the behavior of the mini-batch FedCau
causal latency and test accuracy for $M = 50$, and $T =1.1$s w.r.t.~$F_f$. Fig.~\ref{subfig: CSminiC} demonstrates that the causal latency increases for small and large fairness factors $F_f$. Meanwhile, Fig.~\ref{subfig: CSminiACC1} shows that the test accuracy decreases while $F_f$ increases due to the lack of participation fairness. For smaller $F_f$, the participation fairness results in better test accuracy, while a higher causal latency arises from more frequent transmission of low-power workers. 

}
\begin{figure}[t]
\vspace{4mm}
\centering
\begin{minipage}{0.4\columnwidth}
\vspace{-0.03\textheight}
{\scriptsize\begin{tikzpicture}

\begin{axis}[%
width=0.8\columnwidth,
height=0.5\columnwidth,
at={(0,0)},
scale only axis,
xmin=0, 
xmax=1,
xlabel={Transmission probability $p_x$},
xtick={0,0.2,0.4,0.6,0.8,1},
xtick style={color=black},
ymin=0.1,
ymax=0.5,
ytick={0.2, 0.3, 0.4, 0.5},
grid style={dashed},
grid=both,
ymajorgrids,
ylabel near ticks,
ylabel={\tiny{$C(K)$}},
axis background/.style={fill=white},
legend style={at={(-0.35,1.5)},legend columns=3, font = \tiny,  anchor=north west, legend cell align=left, align=left}
]
\addplot [color=blue, dashdotted, very thick]
  table[row sep=crcr]{%
0.01	0.30781\\
0.0621052631578947	0.277805\\
0.114210526315789	0.277995\\
0.166315789473684	0.292975\\
0.218421052631579	0.28752\\
0.270526315789474	0.29685\\
0.322631578947368	0.29767\\
0.374736842105263	0.276845\\
0.426842105263158	0.28656\\
0.478947368421053	0.29867\\
0.531052631578947	0.302385\\
0.583157894736842	0.28235\\
0.635263157894737	0.29348\\
0.687368421052632	0.30195\\
0.739473684210526	0.28615\\
0.791578947368421	0.30384\\
0.843684210526316	0.30059\\
0.89578947368421	0.279715\\
0.947894736842105	0.297655\\
1	0.308115\\
};
\addlegendentry{$C(k*)$, Slotted ALOHA}

\addplot [color=blue, mark=square, mark options={scale=0.6}]
  table[row sep=crcr]{%
0.01	0.330945\\
0.0621052631578947	0.292175\\
0.114210526315789	0.29785\\
0.166315789473684	0.317695\\
0.218421052631579	0.30587\\
0.270526315789474	0.315705\\
0.322631578947368	0.3199\\
0.374736842105263	0.299065\\
0.426842105263158	0.30588\\
0.478947368421053	0.317335\\
0.531052631578947	0.32252\\
0.583157894736842	0.308645\\
0.635263157894737	0.31269\\
0.687368421052632	0.324685\\
0.739473684210526	0.313095\\
0.791578947368421	0.323215\\
0.843684210526316	0.324595\\
0.89578947368421	0.300165\\
0.947894736842105	0.31923\\
1	0.330915\\
};
\addlegendentry{$C(k_c)$, Slotted ALOHA}

\addplot [color=red, dashed, thick]
  table[row sep=crcr]{%
0.01	0.226139\\
0.0621052631578947	0.241205\\
0.114210526315789	0.2535715\\
0.166315789473684	0.2521275\\
0.218421052631579	0.249958\\
0.270526315789474	0.2482355\\
0.322631578947368	0.2505165\\
0.374736842105263	0.2507975\\
0.426842105263158	0.249494\\
0.478947368421053	0.2501085\\
0.531052631578947	0.243201\\
0.583157894736842	0.2453525\\
0.635263157894737	0.2493835\\
0.687368421052632	0.248088\\
0.739473684210526	0.244103\\
0.791578947368421	0.242152\\
0.843684210526316	0.250099\\
0.89578947368421	0.241347\\
0.947894736842105	0.2455325\\
1	0.245882\\
};
\addlegendentry{$C(k*)$, CSMA/CA}

\addplot [color=red, mark=square*, thin,mark options={scale=0.6}, dotted]
  table[row sep=crcr]{%
0.01	0.2294\\
0.0621052631578947	0.2463965\\
0.114210526315789	0.2600945\\
0.166315789473684	0.2590035\\
0.218421052631579	0.25631\\
0.270526315789474	0.254684\\
0.322631578947368	0.256889\\
0.374736842105263	0.2572895\\
0.426842105263158	0.255868\\
0.478947368421053	0.257127\\
0.531052631578947	0.250048\\
0.583157894736842	0.2517855\\
0.635263157894737	0.2555455\\
0.687368421052632	0.2542415\\
0.739473684210526	0.250484\\
0.791578947368421	0.248498\\
0.843684210526316	0.255671\\
0.89578947368421	0.247338\\
0.947894736842105	0.251085\\
1	0.2518675\\
};
\addlegendentry{$C(k_c)$, CSMA/CA}

\addplot [color=black] 
  table[row sep=crcr]{%
0.01	0.330945\\
0.0621052631578947	0.352\\
0.114210526315789	0.3211\\
0.166315789473684	0.3937\\
0.218421052631579	0.41\\
0.270526315789474	0.3827\\
0.322631578947368	0.3943\\
0.374736842105263	0.3511\\
0.426842105263158	0.3352\\
0.478947368421053	0.4071\\
0.531052631578947	0.38252\\
0.583157894736842	0.348645\\
0.635263157894737	0.37269\\
0.687368421052632	0.384685\\
0.739473684210526	0.413095\\
0.791578947368421	0.423215\\
0.843684210526316	0.3624595\\
0.89578947368421	0.390165\\
0.947894736842105	0.38923\\
1	0.330915\\
};
\addlegendentry{$C(k_c)$, Upper Bound}

\addplot [color=black, thick, dotted]
  table[row sep=crcr]{%
0.01	0.13\\
0.0621052631578947	0.14\\
0.114210526315789	0.112\\
0.166315789473684	0.12\\
0.218421052631579	0.135\\
0.270526315789474	0.1336\\
0.322631578947368	0.14\\
0.374736842105263	0.1442\\
0.426842105263158	0.136\\
0.478947368421053	0.14001\\
0.531052631578947	0.1423\\
0.583157894736842	0.1431\\
0.635263157894737	0.1463\\
0.687368421052632	0.14089\\
0.739473684210526	0.1438\\
0.791578947368421	0.142846\\
0.843684210526316	0.14193\\
0.89578947368421	0.1469\\
0.947894736842105	0.14034\\
1	0.1437\\
};
\addlegendentry{$C(k_c)$, Lower Bound}

\end{axis}
\end{tikzpicture}%
}

\subcaption{$C(k^*)$ and $C(k_c)$ vs $p_x$.}
\label{subfig: ckvspx}
\end{minipage}
\hspace{1.4mm}
\begin{minipage}{0.4\columnwidth}
{\scriptsize\begin{tikzpicture}

\begin{axis}[%
width=0.8\columnwidth,
height=0.5\columnwidth,
at={(0,0)},
scale only axis,
xmin=0, 
xmax=1,
xlabel={Arrival probability $p_r$},
xtick={0,0.2,0.4,0.6,0.8,1},
xtick style={color=black},
ymin=0,
ymax=21,
grid style={dashed},
grid=both,
ymajorgrids,
ylabel near ticks,
ylabel={\tiny{$C(K)$}},
axis background/.style={fill=white},
legend style={at={(0.99,0.99)}, anchor=north east, legend cell align=left, align=left}
]
\addplot [color=blue, dashdotted, very thick]
  table[row sep=crcr]{%
0.01	0.2846\\
0.0621052631578947	0.283165\\
0.114210526315789	0.288525\\
0.166315789473684	0.27904\\
0.218421052631579	0.3053275\\
0.270526315789474	0.295285\\
0.322631578947368	0.3226075\\
0.374736842105263	0.3440275\\
0.426842105263158	0.355105\\
0.478947368421053	0.3452475\\
0.531052631578947	0.357415\\
0.583157894736842	0.3994475\\
0.635263157894737	0.4354575\\
0.687368421052632	0.4154775\\
0.739473684210526	0.4795725\\
0.791578947368421	0.5110575\\
0.843684210526316	0.5556675\\
0.89578947368421	0.612635\\
0.947894736842105	0.679645\\
1	0.63251\\
};

\addplot [color=blue, mark=square, mark options={scale=0.6}]
  table[row sep=crcr]{%
0.01	0.3092875\\
0.0621052631578947	0.308885\\
0.114210526315789	0.31619\\
0.166315789473684	0.3133025\\
0.218421052631579	0.3331025\\
0.270526315789474	0.331465\\
0.322631578947368	0.35575\\
0.374736842105263	0.38136\\
0.426842105263158	0.393805\\
0.478947368421053	0.40039\\
0.531052631578947	0.407775\\
0.583157894736842	0.447305\\
0.635263157894737	0.5061525\\
0.687368421052632	0.4847325\\
0.739473684210526	0.5512775\\
0.791578947368421	0.6037875\\
0.843684210526316	0.654295\\
0.89578947368421	0.7214375\\
0.947894736842105	0.8006325\\
1	0.763885\\
};

\addplot [color=red, dashed, thick]
  table[row sep=crcr]{%
0.01	0.24804515\\
0.0621052631578947	0.25025385\\
0.114210526315789	0.2561846\\
0.166315789473684	0.25929445\\
0.218421052631579	0.2661734\\
0.270526315789474	0.2669679\\
0.322631578947368	0.26134775\\
0.374736842105263	0.2879099\\
0.426842105263158	0.30183245\\
0.478947368421053	0.3299591\\
0.531052631578947	0.556591866666667\\
0.583157894736842	1.3830213\\
0.635263157894737	1.74\\
0.687368421052632	1.8\\
0.739473684210526	2.05\\
0.791578947368421	3.9\\
0.843684210526316	5.7\\
0.89578947368421	7.75\\
0.947894736842105	9.5\\
1	12.8\\
};

\addplot [color=red, mark=square*, thin,mark options={scale=0.6}, dotted]
  table[row sep=crcr]{%
0.01	0.25407375\\
0.0621052631578947	0.25686765\\
0.114210526315789	0.26361785\\
0.166315789473684	0.26728305\\
0.218421052631579	0.27559645\\
0.270526315789474	0.2782219\\
0.322631578947368	0.27671975\\
0.374736842105263	0.30253435\\
0.426842105263158	0.32545355\\
0.478947368421053	0.3675374\\
0.531052631578947	0.671605766666667\\
0.583157894736842	2.2755631\\
0.635263157894737	2.925\\
0.687368421052632	3.6\\
0.739473684210526	4.1\\
0.791578947368421	7.8\\
0.843684210526316	11.4\\
0.89578947368421	15.5\\
0.947894736842105	19\\
1	25.6\\
};
\addplot [color=black]
  table[row sep=crcr]{%
0.01	0.25407375\\
0.0621052631578947	0.27686765\\
0.114210526315789	0.296361785\\
0.166315789473684	0.296728305\\
0.218421052631579	0.297559645\\
0.270526315789474	0.29782219\\
0.322631578947368	0.297671975\\
0.374736842105263	0.32253435\\
0.426842105263158	0.34545355\\
0.478947368421053	0.3975374\\
0.531052631578947	0.871605766666667\\
0.583157894736842	3.2755631\\
0.635263157894737	4.25\\
0.687368421052632	5.16\\
0.739473684210526	6.1\\
0.791578947368421	9.55\\
0.843684210526316	19.4\\
0.89578947368421	25.85\\
0.947894736842105	27\\
1	30\\
};

\end{axis}
\end{tikzpicture}%
}

\subcaption{$C(k^*)$ and $C(k_c)$ vs $p_r$.}
\label{subfig: ckvspr}
\end{minipage}
\hspace{1mm}
\begin{minipage}{0.4\columnwidth}

{\scriptsize\begin{tikzpicture}

\begin{axis}[%
width=0.8\columnwidth,
height=0.5\columnwidth,
at={(0,0)},
scale only axis,
xmin=5, 
xmax=100,
xlabel={Network size $M$},
xtick style={color=black},
ymin=0,
ymax=0.8,
grid style={dashed},
ymajorgrids,
grid=both,
ylabel near ticks,
ylabel={\tiny{$C(K)$}},
axis background/.style={fill=white},
legend style={at={(0,1.5)},legend columns=3, font = \tiny,  anchor=south east, legend cell align=left, align=left}
]
\addplot [color=blue, dashdotted, very thick]
  table[row sep=crcr]{%
5	0.17245\\
10	0.21765\\
20	0.25225\\
30	0.23235\\
40	0.2718\\
50	0.29315\\
60	0.3238\\
70	0.31395\\
90	0.30505\\
100	0.40055\\
};

\addplot [color=blue, mark=square, mark options={scale=0.6}]
  table[row sep=crcr]{%
5	0.17245\\
10	0.2204\\
20	0.2614\\
30	0.25805\\
40	0.2878\\
50	0.3181\\
60	0.3481\\
70	0.364\\
90	0.37845\\
100	0.4519\\
};
\addplot [color=red, dashed, thick]
  table[row sep=crcr]{%
5	0.0766095\\
10	0.187391\\
20	0.2119195\\
30	0.222345\\
40	0.241392\\
50	0.245575\\
60	0.2438695\\
70	0.259444\\
90	0.2670845\\
100	0.2727435\\
};

\addplot [color=red, mark=square*, thin,mark options={scale=0.6}, dotted]
  table[row sep=crcr]{%
5	0.0766095\\
10	0.187391\\
20	0.2139845\\
30	0.2258985\\
40	0.2462125\\
50	0.2516375\\
60	0.251818\\
70	0.267822\\
90	0.278079\\
100	0.284304\\
};

\addplot [color=black]
  table[row sep=crcr]{%
5	0.1945\\
10	0.2804\\
20	0.3514\\
30	0.42805\\
40	0.4878\\
50	0.5381\\
60	0.581\\
70	0.64\\
90	0.67845\\
100	0.70519\\
};
\addplot [color=black, thick, dotted]
  table[row sep=crcr]{%
5	0.01945\\
10	0.0204\\
20	0.02514\\
30	0.02625\\
40	0.0278\\
50	0.0279\\
60	0.0312\\
70	0.033\\
90	0.0352\\
100	0.038\\
};
\end{axis}
\end{tikzpicture}%
}
\subcaption{$C(k^*)$ and $C(k_c)$ vs $M$.}
\label{subfig: ckvsM}
\end{minipage}
\caption{Illustration of the batch FedCau update Algorithm~\ref{alg: synchronous alg}: iteration-cost $C(k^*)$ and $C(k_c)$ and the bounds of Eq.~\eqref{eq: upper comm} vs transmission probability~$p_x$, arrival probability~$p_r$, and network size~$M$. 
}
\label{fig: Batch2c}
\end{figure}
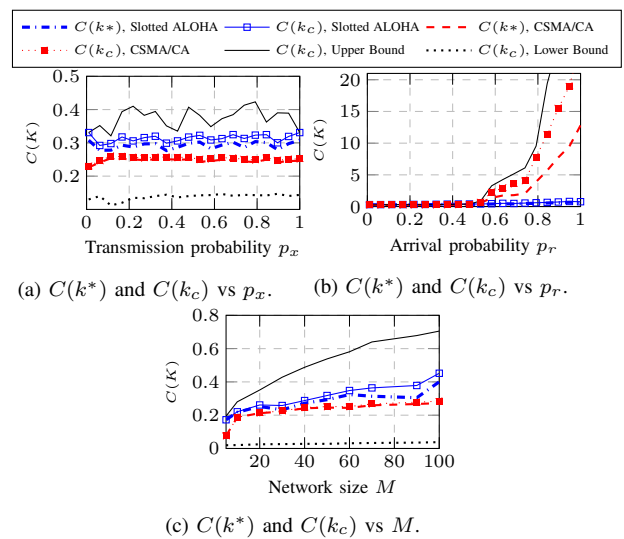

\subsection{Impact of Communication Parameters on FedCau Performance }

Fig.~\ref{fig: Batch2c} characterizes the iteration-cost function~$C(K)$ for the same setup as in Fig.~\ref{fig: Batch1k}. The iteration-cost function for slotted-ALOHA is larger than CSMA/CA, as we see in Figs.~\ref{subfig: ckvspx} and~\ref{subfig: ckvsM}. On the other hand, the iteration-cost function for CSMA/CA increases exponentially when the probability~$p_r$ increases, as shown in Fig.~\ref{subfig: ckvspr}. This result also holds for the bounds of the iteration cost in Eq.~\eqref{eq: upper comm}, as Fig.~\ref{fig: Batch2c} shows. {\color{black}Furthermore, the results from Fig.~\ref{subfig: ckvsM} show that $C(k_c)$ increases on a slower rate than $M$ increases, such that
\begin{equation}
    \frac{C(k_{c_2})-C(k_{c_1})}{M_2 - M_1} < 1,\quad M_2 > M_1,
\end{equation}
where $M_2$ and $M_1$ are number of workers, and $C(k_{c_2})$ and $C(k_{c_1})$ are the total communication-computation with the stopping causal iterations~$k_{c_2}$ and $k_{c_1}$, respectively. Thus, considering full worker participation as the worst case when investigating the scalability, we conclude that the total communication-computation cost of FedCau is scalable in~$M$.}


\subsection{Performance of Non-convex FedCau from Algorithm~\ref{alg: Nonconvex alg}}


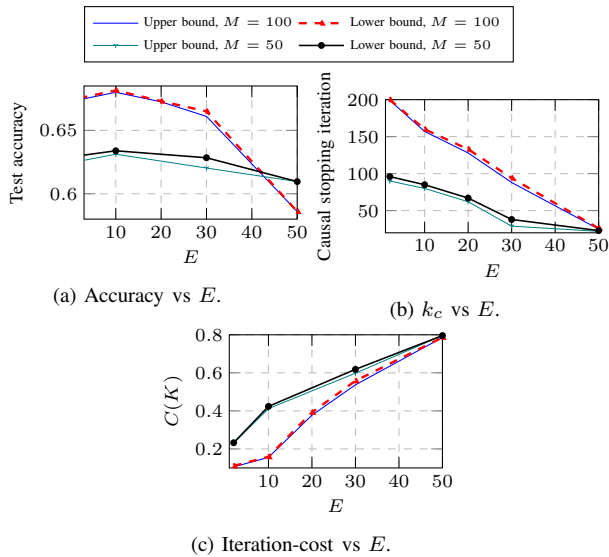
\begin{figure}[t]
\vspace{0.036\textheight}
\centering
\begin{minipage}{0.4\columnwidth}
\vspace{-0.04\textheight}
{\scriptsize\definecolor{tealgreen}{rgb}{0.0, 0.51, 0.5}
\begin{tikzpicture}

\begin{axis}[%
width=0.8\columnwidth,
height=0.5\columnwidth,
at={(0,0)},
scale only axis,
xmin=3, 
xmax=50,
xlabel={$E$},
xtick style={color=black},
ymin=0.58,
ymax=0.685,
xtick={10,20,30,40,50},
grid style={dashed},
ymajorgrids,
grid=both,
ylabel near ticks,
ylabel={Test accuracy},
axis background/.style={fill=white},
legend style={at={(0.0,1.6)},legend columns=2, font = \tiny,  anchor=north west, legend cell align=left, align=left}
]
\addplot [color=blue]
  table[row sep=crcr]{%
2	0.6742\\ 
10	0.68\\ 
20	0.6725\\ 
30	0.661\\ 
50	0.586\\ 
};
\addlegendentry{Upper bound, $M=100$}
\addplot [color=red,line width=0.9pt, mark=triangle,dashed, mark options={scale=0.5}]
  table[row sep=crcr]{%
2	0.6754\\ 
10	0.6815\\ 
20	0.673\\ 
30	0.665\\ 
50	0.586\\ 
};
\addlegendentry{Lower bound, $M=100$}
\addplot [color= tealgreen, mark=Mercedes star flipped, very thin, mark options={scale=0.5}]
  table[row sep=crcr]{%
2	0.6257\\ 
10  0.6312\\ 
30	0.6203\\ 
50	0.6096\\ 
};

\addlegendentry{Upper bound, $M=50$}
\addplot [color=black,line width=0.6pt, mark=* , mark options={scale=0.5}]
  table[row sep=crcr]{%
2	0.63\\ 
10  0.6339\\ 
30	0.6284\\ 
50	0.6096\\ 
};
\addlegendentry{Lower bound, $M=50$}

\end{axis}
\end{tikzpicture}%
}

\subcaption{ Accuracy vs $E$.}
\label{subfig: accvsE}
\end{minipage}
\hspace{3mm}
\begin{minipage}{0.4\columnwidth}
{\scriptsize\definecolor{tealgreen}{rgb}{0.0, 0.51, 0.5}
\begin{tikzpicture}

\begin{axis}[%
width=0.8\columnwidth,
height=0.5\columnwidth,
at={(0,0)},
scale only axis,
xmin=1, 
xmax=50,
xlabel={$E$},
xtick style={color=black},
ymin=20,
ymax=200,
xtick={10,20,30,40,50},
grid style={dashed},
ymajorgrids,
grid=both,
ylabel near ticks,
ylabel={Causal stopping iteration},
axis background/.style={fill=white},
legend style={at={(0.99,0)}, anchor=south east, legend cell align=left, font = \tiny}
]
\addplot [color=blue]
  table[row sep=crcr]{%
2	199\\ 
10	157\\ 
20	128\\ 
30	88\\ 
50	25\\ 
};
\addplot [color=red,line width=0.9pt, mark=triangle,dashed, mark options={scale=0.5}]
  table[row sep=crcr]{%
2	200\\ 
10	160\\ 
20	133\\ 
30	94\\ 
50	26\\ 
};
\addplot [color= tealgreen, mark=Mercedes star flipped, very thin, mark options={scale=0.5}]
  table[row sep=crcr]{%
2	90\\ 
10  80\\ 
20	62\\ 
30	29\\ 
50	22\\ 
};

\addplot [color=black,line width=0.6pt, mark=* , mark options={scale=0.5}]
  table[row sep=crcr]{%
2	96\\ 
10  85\\ 
20	67\\ 
30	38\\ 
50	23\\ 
};

\end{axis}
\end{tikzpicture}%
}

\subcaption{$k_c$ vs $E$.}
\label{subfig: kvsE}
\end{minipage}
\hspace{2mm}
\begin{minipage}{0.4\columnwidth}
{\scriptsize\definecolor{tealgreen}{rgb}{0.0, 0.51, 0.5}
\begin{tikzpicture}

\begin{axis}[%
width=0.8\columnwidth,
height=0.5\columnwidth,
at={(0,0)},
scale only axis,
xmin=1, 
xmax=50,
xlabel={$E$},
xtick style={color=black},
ymin=0.1,
ymax=0.8,
xtick={10,20,30,40,50},
grid style={dashed},
ymajorgrids,
grid=both,
ylabel near ticks,
ylabel={$C(K)$},
axis background/.style={fill=white},
legend style={at={(0.99,0)}, anchor=south east, legend cell align=left, font = \tiny}
]
\addplot [color=blue]
  table[row sep=crcr]{%
2	0.1056\\ 
10	0.1556\\ 
20	0.3772\\ 
30	0.5369\\ 
50	0.786\\ 
};
\addplot [color=red,line width=0.9pt, mark=triangle,dashed, mark options={scale=0.5}]
  table[row sep=crcr]{%
2	0.1107\\ 
10	0.1587\\ 
20	0.3922\\ 
30	0.558\\ 
50	0.786\\ 
};
\addplot [color= tealgreen, mark=Mercedes star flipped, very thin, mark options={scale=0.5}]
  table[row sep=crcr]{%
2	0.2302\\ 
10  0.4120\\ 
30	0.598\\ 
50	0.796\\ 
};

\addplot [color=black,line width=0.6pt, mark=* , mark options={scale=0.5}]
  table[row sep=crcr]{%
2	0.2325\\ 
10  0.4235\\ 
30	0.6184\\ 
50	0.796\\ 
};

\end{axis}
\end{tikzpicture}%
}

\subcaption{Iteration-cost vs $E$.}
\label{subfig: cvsE}
\end{minipage}
\caption{{\color{black}Illustration of the effect of number of local iterations~$E$ on the performance of mini-batch FedCau update of Algorithm~\ref{alg: Nonconvex alg} for non-convex loss functions with CIFAR-10 iid dataset and CSMA/CA, $M=50, 100$, $p_x = 0.8$, and $p_r = 0.01$.
}}
\label{fig: EPerf}
\end{figure}
The experimental results presented in Fig.~\ref{fig: EPerf} investigate the impact of the number of local iterations ($E$) on the performance of mini-batch FedCau updates of Algorithm~3. The study focuses on the CIFAR-10 iid dataset and CNN architecture, employing CSMA/CA with different values of $M=50$ and $M=100$, along with $p_x = 0.8$ and $p_r = 0.01$. Fig.~\ref{subfig: accvsE} reveals distinct behaviors in the causal test accuracy concerning $E$ for $M=50$ and $M=100$. While the changes in test accuracy are less pronounced for $M=50$, the corresponding values are lower than $M=100$. Fig.~\ref{subfig: kvsE} showcases the causal stopping iterations ($k_c^u$ and $k_c^l$), which decrease as $E$ increases. Additionally, the tightness of the interval $(k_c^u, k_c^l)$ established in Proposition~\ref{prop: k_u and k_l} is validated, according to the variations in the non-convex sequences of $\Tilde{F}(\bw_k)$.
Moreover, Fig.~\ref{subfig: cvsE} shows the causal iteration-cost $C(K)$ as a function of $E$, which increases as $E$ increases. This observation highlights the significant impact of computation latency on the performance of the FedCau. Based on the findings in Fig.~\ref{fig: EPerf}, selecting $E=10$ as the optimal number of local iterations is recommended, which provides the best accuracy with a lower causal iteration cost compared to $E>10$. These results offer valuable insights into selecting $E$ and understanding the trade-off between $E$, test accuracy, iteration cost, and causal stopping iterations. 

Fig.~\ref{fig: Noniidcifar} compares the performance of mini-batch FedCau update of Algorithm~\ref{alg: Nonconvex alg} in iid and non-iid data distribution of CIFAR-10, for non-convex loss functions with  CSMA/CA, $M=100$, $E=10$, $p_x = 0.8$, and $p_r = 0.01$. Fig.~\ref{subfig: AccNoniidcifar} compares the test accuracy of training the mini-batch FedCau update of Algorithm~\ref{alg: Nonconvex alg} by iid and non-iid data obtained by $k_c^l$, the highest test accuracy achieved by~Algorithm~\ref{alg: Nonconvex alg} for any non-convex loss function. We observe that for the iid case, with $k_c^l = 160$, the test accuracy is higher than the case for non-iid with $k_c^l = 93$. Figs.~\ref{subfig: Fiidcifar}~and~\ref{subfig: Fnoniidcifar} show the loss functions~$\Tilde{F}(\bw_k)$ and the corresponding upper and lower bounds $F_u(\bw_k)$ and $F_l(\bw_k)$. 
The comparison between Fig.~\ref{subfig: Fiidcifar}~and Fig.~\ref{subfig: Fnoniidcifar} reveals that the iid case results in a lower value of loss function and a higher test accuracy, as shown in Fig.~\ref{subfig: AccNoniidcifar}. Moreover, the difference between the upper bound and the lower bound functions $F_u(\bw_k)$ and $F_l(\bw_k)$ are small in Figs.~\ref{subfig: Fiidcifar}~and~\ref{subfig: Fnoniidcifar}, which shows the high tightness of the bounds. {\color{black}Fig.~\ref{subfig: FedAvg_vs_FedCau_cifar} compares the test accuracy of FedCau and FedAvg with stopping iterations $k_c^l$ and $K^{\max}=200$ after 100 realizations to
have smoother curves}. Notably, FedAvg with $K^{\max}=200$ increases the total iteration cost by $55$\% (non-iid) and $20$\% (iid), but the test accuracy improvement over FedCau is only $2.2$\% (non-iid) and $0.65$\% (iid). {\color{black}We observe that non-iid FedCau terminates at iteration~$k_c^l = 93$ while the FedAvg test accuracy curve becomes flat at iteration~$k = 101$. Moreover, the test accuracy of non-iid FedCau with~$k_c^l = 93$ is~$1.2$\% higher than non-iid FedAvg at iteration~$k = 101$. The communication costs of the local parameters for every extra iteration are high; thus, stopping the training at a proper iteration saves a huge amount of communication resources ($14.7$Mbits per iteration per worker). As a result, FedCau, with the knowledge of when to terminate the training, i.e., $k_c^l = 93$, is significantly superior to FedAvg in terms of saving communication-computation resources and achieving higher test accuracy.

} 
\begin{figure}[t]
\centering
\begin{minipage}{0.4\columnwidth}
{\scriptsize\definecolor{tealgreen}{rgb}{0.0, 0.51, 0.5}
\begin{tikzpicture}

\begin{axis}[%
width=0.8\columnwidth,
height=0.5\columnwidth,
at={(0,0)},
scale only axis,
xmin=0, 
xmax=170,
xlabel={Communication iteration $k$},
xtick style={color=black},
ymin=0,
ymax=0.75,
grid style={dashed},
ymajorgrids,
grid=both,
ylabel near ticks,
ylabel={Test accuracy},
axis background/.style={fill=white},
legend style={at={(0.99,0)}, anchor=south east, legend cell align=left, font = \tiny}
]
\addplot [color=blue, line width=0.5pt]
  table[row sep=crcr]{%
1	0.0994\\
2	0.1825\\
3	0.1895\\
4	0.2527\\
5	0.2762\\
6	0.2638\\
7	0.2871\\
8	0.3093\\
9	0.3386\\
10	0.3649\\
11	0.3745\\
12	0.4078\\
13	0.413\\
14	0.4372\\
15	0.4382\\
16	0.4516\\
17	0.456\\
18	0.4644\\
19	0.4688\\
20	0.481\\
21	0.4913\\
22	0.4909\\
23	0.4978\\
24	0.5106\\
25	0.5066\\
26	0.5125\\
27	0.5139\\
28	0.5248\\
29	0.5298\\
30	0.5326\\
31	0.5357\\
32	0.5428\\
33	0.5376\\
34	0.55\\
35	0.5517\\
36	0.5525\\
37	0.5543\\
38	0.5537\\
39	0.5651\\
40	0.5658\\
41	0.5689\\
42	0.5682\\
43	0.5736\\
44	0.5765\\
45	0.5778\\
46	0.5805\\
47	0.5787\\
48	0.5793\\
49	0.5801\\
50	0.5897\\
51	0.5876\\
52	0.5951\\
53	0.5891\\
54	0.5901\\
55	0.5927\\
56	0.5956\\
57	0.6049\\
58	0.5997\\
59	0.6015\\
60	0.603\\
61	0.6073\\
62	0.6083\\
63	0.6079\\
64	0.6135\\
65	0.6131\\
66	0.6114\\
67	0.6118\\
68	0.6144\\
69	0.6166\\
70	0.6144\\
71	0.618\\
72	0.6168\\
73	0.6177\\
74	0.6211\\
75	0.6242\\
76	0.6253\\
77	0.6201\\
78	0.623\\
79	0.6295\\
80	0.6326\\
81	0.6332\\
82	0.6353\\
83	0.6359\\
84	0.6361\\
85	0.6344\\
86	0.6361\\
87	0.6376\\
88	0.6348\\
89	0.6399\\
90	0.6424\\
91	0.6438\\
92	0.644\\
93	0.6446\\
94	0.6481\\
95	0.6437\\
96	0.6451\\
97	0.6446\\
98	0.6485\\
99	0.6458\\
100	0.6469\\
101	0.6472\\
102	0.6472\\
103	0.6476\\
104	0.6497\\
105	0.6505\\
106	0.6506\\
107	0.6546\\
108	0.6573\\
109	0.6534\\
110	0.6568\\
111	0.6568\\
112	0.6588\\
113	0.658\\
114	0.6592\\
115	0.659\\
116	0.6629\\
117	0.6648\\
118	0.6613\\
119	0.6612\\
120	0.6606\\
121	0.6638\\
122	0.6585\\
123	0.666\\
124	0.664\\
125	0.6635\\
126	0.6672\\
127	0.6647\\
128	0.6637\\
129	0.665\\
130	0.6651\\
131	0.666\\
132	0.6694\\
133	0.6731\\
134	0.6692\\
135	0.6733\\
136	0.671\\
137	0.6711\\
138	0.6707\\
139	0.6735\\
140	0.6701\\
141	0.6745\\
142	0.6731\\
143	0.6706\\
144	0.6729\\
145	0.6739\\
146	0.6717\\
147	0.6781\\
148	0.6766\\
149	0.6757\\
150	0.6767\\
151	0.6796\\
152	0.6771\\
153	0.6791\\
154	0.6779\\
155	0.6785\\
156	0.6799\\
157	0.6809\\
158	0.6817\\
159	0.6825\\
160	0.6815\\
};
\addlegendentry{FedCau, iid } 
\addplot [color=black,line width=0.9pt]
  table[row sep=crcr]{%
1	0.040164319792576\\
2	0.132136423762869\\
3	0.130356955074565\\
4	0.203071\\
5	0.20307\\
6	0.2239\\
7	0.220485498031177\\
8	0.258228478957772\\
9	0.274598078057004\\
10	0.304938564837892\\
11	0.317506559039263\\
12	0.349769458541787\\
13	0.34331897525258\\
14	0.364240584376142\\
15	0.37006198519714\\
16	0.398287959661662\\
17	0.394983926684936\\
18	0.401015694275683\\
19	0.40455752251951\\
20	0.42534964800051\\
21	0.43264801053202\\
22	0.432343231791663\\
23	0.426825522614618\\
24	0.446409420673592\\
25	0.444227152108888\\
26	0.446883307023249\\
27	0.450909305702147\\
28	0.458360535069178\\
29	0.474228661003502\\
30	0.467198596559745\\
31	0.478488910674781\\
32	0.479823335184799\\
33	0.480829505553039\\
34	0.498438980974085\\
35	0.489631360063773\\
36	0.500076718715667\\
37	0.484754816053273\\
38	0.500530483504323\\
39	0.492167323298271\\
40	0.502435085414267\\
41	0.4944482455702\\
42	0.513407013750889\\
43	0.503080237519781\\
44	0.505634806171349\\
45	0.507340086679053\\
46	0.52278591408278\\
47	0.509044087980699\\
48	0.51175262149433\\
49	0.524308378676494\\
50	0.52722875236615\\
51	0.535336620162682\\
52	0.521546322577198\\
53	0.525018798206072\\
54	0.529406368639279\\
55	0.531826678994429\\
56	0.541195414180659\\
57	0.533052470687065\\
58	0.532202919438344\\
59	0.52655305062083\\
60	0.551777683947257\\
61	0.5514574840833\\
62	0.552082604054704\\
63	0.536371620928523\\
64	0.547232988566066\\
65	0.541883514012437\\
66	0.548818016956448\\
67	0.557016306322805\\
68	0.555221719836029\\
69	0.558810037442321\\
70	0.555389255446381\\
71	0.561233043156089\\
72	0.563296142702134\\
73	0.558373884485364\\
74	0.550614784322735\\
75	0.563532663658314\\
76	0.554389153909601\\
77	0.567751293389388\\
78	0.549933569628333\\
79	0.562963366253542\\
80	0.572115438684928\\
81	0.570080878937307\\
82	0.580433012155638\\
83	0.57416945352437\\
84	0.566782758291057\\
85	0.577137124379912\\
86	0.564815813830121\\
87	0.579052904530585\\
88	0.575649563697252\\
89	0.570632656286644\\
90	0.571955700381818\\
};
\addlegendentry{FedCau, non-iid}

\end{axis}
\end{tikzpicture}%
}

\subcaption{ Test accuracy by $k_c^l$.}
\label{subfig: AccNoniidcifar}
\end{minipage}
\hspace{1.4mm}
\begin{minipage}{0.4\columnwidth}
{\scriptsize\definecolor{tealgreen}{rgb}{0.0, 0.51, 0.5}
\begin{tikzpicture}

\begin{axis}[%
width=0.8\columnwidth,
height=0.5\columnwidth,
at={(0,0)},
scale only axis,
xmin=0, 
xmax=160,
xlabel={Communication iteration $k$},
xtick style={color=black},
ymin=1.8,
ymax=2.45,
grid style={dashed},
ymajorgrids,
grid=both,
ylabel near ticks,
ylabel={Loss functions},
axis background/.style={fill=white},
legend style={at={(0.99,0.99)}, font = \tiny}
]
\addplot [color=blue]
  table[row sep=crcr]{%
1	2.302431583\\
2	2.298492193\\
3	2.293401003\\
4	2.274686337\\
5	2.26490593\\
6	2.257844448\\
7	2.240503073\\
8	2.230208874\\
9	2.208050251\\
10	2.198953152\\
11	2.182641506\\
12	2.174900055\\
13	2.152793407\\
14	2.145280123\\
15	2.128489017\\
16	2.126689672\\
17	2.109863758\\
18	2.100760937\\
19	2.090877533\\
20	2.084696054\\
21	2.070654392\\
22	2.065624237\\
23	2.05872941\\
24	2.047087193\\
25	2.038680077\\
26	2.031282663\\
27	2.028793335\\
28	2.010804415\\
29	2.013368845\\
30	2.002903223\\
31	1.997027516\\
32	1.992991567\\
33	1.990642428\\
34	1.981850386\\
35	1.975456238\\
36	1.973824024\\
37	1.964317918\\
38	1.967796087\\
39	1.964011669\\
40	1.963786364\\
41	1.954226375\\
42	1.952615619\\
43	1.953639627\\
44	1.949900031\\
45	1.939165235\\
46	1.935360193\\
47	1.938891768\\
48	1.932192564\\
49	1.930863619\\
50	1.927191019\\
51	1.925608635\\
52	1.923206449\\
53	1.923024654\\
54	1.917713881\\
55	1.91174376\\
56	1.912295938\\
57	1.912168384\\
58	1.906150222\\
59	1.904448271\\
60	1.905060172\\
61	1.906545162\\
62	1.898270845\\
63	1.897207022\\
64	1.898095131\\
65	1.89421916\\
66	1.896362901\\
67	1.892759562\\
68	1.889385581\\
69	1.889730692\\
70	1.89089644\\
71	1.886865616\\
72	1.887536168\\
73	1.882000923\\
74	1.882938623\\
75	1.880197406\\
76	1.878182173\\
77	1.879657984\\
78	1.880534172\\
79	1.875754356\\
80	1.875368357\\
81	1.872233987\\
82	1.873868346\\
83	1.868446827\\
84	1.864873052\\
85	1.866833925\\
86	1.868035555\\
87	1.864229918\\
88	1.863317966\\
89	1.864586353\\
90	1.860514402\\
91	1.856778502\\
92	1.856870174\\
93	1.855707765\\
94	1.854560971\\
95	1.85661459\\
96	1.857172608\\
97	1.851983547\\
98	1.847037911\\
99	1.852628112\\
100	1.850326896\\
101	1.85087347\\
102	1.849301577\\
103	1.850556493\\
104	1.846470356\\
105	1.847065449\\
106	1.848734736\\
107	1.8416394\\
108	1.841173053\\
109	1.84397459\\
110	1.841439843\\
111	1.839085102\\
112	1.840717554\\
113	1.839123487\\
114	1.835454345\\
115	1.839724779\\
116	1.837398052\\
117	1.835766554\\
118	1.837228179\\
119	1.836686373\\
120	1.835899234\\
121	1.831605434\\
122	1.834695935\\
123	1.831231594\\
124	1.835143328\\
125	1.832669735\\
126	1.832396507\\
127	1.831861377\\
128	1.831554294\\
129	1.829040408\\
130	1.829770684\\
131	1.830103278\\
132	1.825448632\\
133	1.825781584\\
134	1.824789047\\
135	1.825139403\\
136	1.822024822\\
137	1.822587729\\
138	1.823976159\\
139	1.822634339\\
140	1.824407458\\
141	1.820381403\\
142	1.821513534\\
143	1.825447798\\
144	1.821183562\\
145	1.820379734\\
146	1.821251988\\
147	1.817798018\\
148	1.818011165\\
149	1.820461154\\
150	1.818886757\\
151	1.816257477\\
152	1.819289088\\
153	1.8185637\\
154	1.819098234\\
155	1.81678319\\
156	1.815844178\\
157	1.81386137\\
158	1.814365029\\
159	1.813604712\\
160	1.811589479\\
};
\addlegendentry{$\Tilde{F}(\bw_k)$}

\addplot [color=red,line width=0.6pt, mark=triangle,dashed, mark options={scale=0.5}]
  table[row sep=crcr]{%
1	2.302431583\\
5	2.26490593\\
10	2.198953152\\
15	2.128489017\\
20	2.084696054\\
25	2.038680077\\
30	2.009418249\\
35	1.981971264\\
40	1.9674176575\\
45	1.941547037\\
50	1.9323064085\\
55	1.9168591495\\
65	1.8966012\\
70	1.89089644\\
75	1.8820368645\\
80	1.87933890066667\\
85	1.868218273\\
90	1.863184571\\
95	1.8576278206\\
100	1.851750791\\
105	1.84959831825\\
110	1.841439843\\
115	1.839724779\\
120	1.835899234\\
125	1.832669735\\
130	1.83064618716667\\
135	1.82545134777778\\
140	1.82544912916667\\
145	1.82265059133333\\
150	1.82007046533333\\
157	1.8147177222\\
};
\addlegendentry{$F_u(\bw_k)$}

\addplot [color= tealgreen, line width= 0.6pt, mark=Mercedes star flipped, mark options={scale=0.5}]
  table[row sep=crcr]{%
1	2.302431583\\
5	2.26490593\\
10	2.198953152\\
15	2.128489017\\
20	2.084696054\\
25	2.038680077\\
30	2.002903223\\
35	1.975456238\\
40	1.963786364\\
45	1.939165235\\
47	1.9337763785\\
50	1.927191019\\
55	1.91174376\\
60	1.902389129\\
65	1.89421916\\
70	1.88770560433333\\
75	1.880197406\\
80	1.875368357\\
85	1.864658674\\
90	1.860514402\\
95	1.85370182966667\\
100	1.846848726\\
105	1.84486003733333\\
110	1.83978108566667\\
115	1.83490450057143\\
120	1.83215527842857\\
125	1.83050119866667\\
130	1.82784314933333\\
135	1.8234069345\\
140	1.8207100868\\
145	1.820379734\\
150	1.81664261225\\
155	1.8159268378\\
160	1.811589479\\
};

\addlegendentry{$F_l(\bw_k)$}
\end{axis}
\end{tikzpicture}%
}
\subcaption{Loss function, iid data.}
\label{subfig: Fiidcifar}
\end{minipage}
\hspace{1.4mm}
\begin{minipage}{0.4\columnwidth}
{\scriptsize\definecolor{tealgreen}{rgb}{0.0, 0.51, 0.5}
\begin{tikzpicture}

\begin{axis}[%
width=0.8\columnwidth,
height=0.5\columnwidth,
at={(0,0)},
scale only axis,
xmin=0, 
xmax=160,
xlabel={Communication iteration $k$},
xtick style={color=black},
ymin=1.8,
ymax=2.45,
grid style={dashed},
ymajorgrids,
grid=both,
ylabel near ticks,
ylabel={Loss functions},
axis background/.style={fill=white},
legend style={at={(0.99,0.99)}, font = \tiny}
]
\addplot [color=blue]
  table[row sep=crcr]{%
1	2.43998396906473\\
2	2.42917105297228\\
3	2.4218293341787\\
4	2.38314201276255\\
5	2.37367358901836\\
6	2.38519451738494\\
7	2.37147506450262\\
8	2.33391544685797\\
9	2.32100806962744\\
10	2.31207149832886\\
11	2.2852957060494\\
12	2.29939506972838\\
13	2.26319944775516\\
14	2.27267144788355\\
15	2.23602908162255\\
16	2.25716813052688\\
17	2.22294493242469\\
18	2.20193900913241\\
19	2.21471258008924\\
20	2.21320028891451\\
21	2.21029648787067\\
22	2.19945416802615\\
23	2.20318589908199\\
24	2.17832669028789\\
25	2.1721848780433\\
26	2.16417740439027\\
27	2.13793739243464\\
28	2.13422014928812\\
29	2.13731274257456\\
30	2.11418773755067\\
31	2.141014377461\\
32	2.11171905940412\\
33	2.12330999372247\\
34	2.11511941670801\\
35	2.09672540983095\\
36	2.08062714194706\\
37	2.10438976439147\\
38	2.08417353881512\\
39	2.068920100451\\
40	2.10382850491905\\
41	2.0893121675085\\
42	2.07778960548683\\
43	2.05914441541577\\
44	2.07482930150657\\
45	2.0753154730448\\
46	2.05866228406666\\
47	2.06017929031072\\
48	2.06155234391395\\
49	2.08012543585975\\
50	2.04205277882859\\
51	2.05027127742565\\
52	2.05414314407865\\
53	2.06977807725631\\
54	2.04978488296695\\
55	2.05003119787339\\
56	2.05212073535555\\
57	2.02147940484179\\
58	2.03647849232033\\
59	2.05176951245693\\
60	2.03630759308638\\
61	2.01779545312912\\
62	2.04000596395646\\
63	2.00159607714817\\
64	2.03213484190954\\
65	2.0277301765858\\
66	2.03615111749524\\
67	2.00439882518101\\
68	2.03588692355236\\
69	1.99222899459932\\
70	2.03298595075466\\
71	2.02254332267995\\
72	2.00317971226315\\
73	2.02494343181811\\
74	2.00733429047583\\
75	1.99413105963529\\
76	1.98173991621901\\
77	1.99305279149919\\
78	1.99304177096309\\
79	2.00538837584472\\
80	2.01339377222061\\
81	1.99619146324585\\
82	2.01291447799455\\
83	1.97632743180506\\
84	1.98172640654744\\
85	1.97602216074168\\
86	1.97834167499487\\
87	1.99077465782241\\
88	1.99784171703327\\
89	1.98020739062431\\
90	1.9611717908305\\
};
\addlegendentry{$\Tilde{F}(\bw_k)$}

\addplot [color=red,line width=0.6pt, mark=triangle,dashed, mark options={scale=0.5}]
  table[row sep=crcr]{%
1	2.43998396906473\\
4	2.40961772858078\\
7	2.38409525512707\\
10	2.32469168895332\\
13	2.28834694058126\\
16	2.25716813052688\\
19	2.21471258008924\\
22	2.20967527759411\\
24	2.19262216021287\\
26	2.17847287431524\\
28	2.1485156192131\\
30	2.14351479137837\\
32	2.13978003838773\\
34	2.11511941670801\\
36	2.10725564895421\\
38	2.10420267790066\\
40	2.10382850491905\\
42	2.08816382605241\\
44	2.08586714314022\\
46	2.08357046022803\\
48	2.08127377731584\\
50	2.07597305955023\\
54	2.05990463756485\\
58	2.05188658675647\\
62	2.04000596395646\\
66	2.03615111749524\\
71	2.03030511110914\\
73	2.02494343181811\\
76	2.01999357770489\\
78	2.01669367496275\\
81	2.01315412510758\\
86	2.00025254693037\\
89	1.9959035646302\\
};
\addlegendentry{$F_u(\bw_k)$}

\addplot [color= tealgreen, line width= 0.6pt, mark=Mercedes star flipped, mark options={scale=0.5}]
  table[row sep=crcr]{%
1	2.43998396906473\\
4	2.38314201276255\\
6	2.37257432676049\\
8	2.33391544685797\\
10	2.31207149832886\\
12	2.27424757690228\\
14	2.24961426468886\\
16	2.22948700702362\\
18	2.20193900913241\\
22	2.19945416802615\\
26	2.16417740439027\\
30	2.11418773755067\\
32	2.11171905940412\\
34	2.10172329302201\\
36	2.08062714194706\\
38	2.07282244761635\\
40	2.06647617919219\\
42	2.06158833667457\\
44	2.05898370496606\\
46	2.05866228406666\\
48	2.05035753144762\\
50	2.04205277882859\\
54	2.03029656512185\\
56	2.02441845826847\\
58	2.02055841691362\\
60	2.01871644105728\\
62	2.00969576513864\\
64	2.00003489672336\\
66	1.99691253587374\\
68	1.99379017502413\\
70	1.9907305548307\\
72	1.98773367529347\\
74	1.98473679575624\\
76	1.98173991621901\\
78	1.98019349210074\\
82	1.9771006438642\\
86	1.97305208675944\\
90	1.9611717908305\\
};

\addlegendentry{$F_l(\bw_k)$}
\end{axis}
\end{tikzpicture}%
}
\subcaption{  Loss function, non-iid.}
\label{subfig: Fnoniidcifar}
\end{minipage}
\hspace{1.4mm}
\begin{minipage}{0.4\columnwidth}
{\scriptsize\input{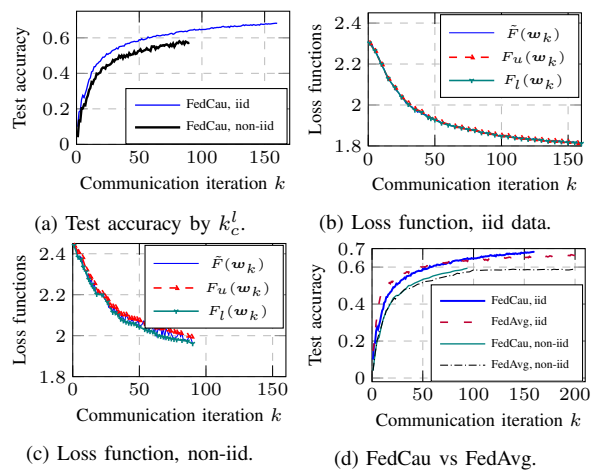}}

\subcaption{{\color{black}FedCau vs FedAvg.}}
\label{subfig: FedAvg_vs_FedCau_cifar}
\end{minipage}
\caption{{\color{black}Performance comparison of CIFAR-10 iid and non-iid data in mini-batch FedCau of Algorithm~\ref{alg: Nonconvex alg} for non-convex loss functions with  CSMA/CA, $M=100$, $E=10$, $K^{\max}=200$, $p_x = 0.8$, and $p_r = 0.01$. a) Test accuracy of CIFAR-10 iid and non-iid dataset obtained by the lower bound causal stopping iteration~$k_c^l$. b) Loss function~$\Tilde{F}(\bw_k)$ with its upper bound~$F_u(\bw_k)$ and lower bound~$F_l(\bw_k)$ for iid data, and c) non-iid data. {\color{black}d) Comparison between FedCau and FedAvg.}
}}
\label{fig: Noniidcifar}
\end{figure}

\subsection{Performance of FedCau Update from Algorithm~\ref{alg: synchronous alg} on Top of LAQ and Top-$q$ }

We choose LAQ because it achieves the same linear convergence as the gradient descent while effecting major savings in the communication resources~\cite{9238427}. Among all the compression methods, we choose top-$q$ sparsification because it suffers the least from non-i.i.d. data, and the training converges. Moreover, applying top-$q$ for the logistic regression classifier trained on MNIST, the convergence does not slow down~\cite{8889996}. Despite the previous numerical results, which characterize the overall latency as the iteration-cost $c_k, k\ge 1$, here we consider the number of bits per each communication iteration as $c_k, k\ge1$. In LAQ, $b$ shows the element-wise number of bits for the local parameters{\color{black}, and we train the FedAvg algorithm over the MNIST dataset}. Moreover, in the top-$q$ method, we change the percentage of the dimension of each local parameter as $0<q \le 1$, but considering that each element contains $32$ bits. TABLE I compares FedAvg and FedCau with and without considering the communication-efficient methods LAQ and Top-$q$. {\color{black} In Table I, FedCau LAQ with $b=2$ achieves $94.2$\% test accuracy, using the least number of bits (total cost of $4.56$Mbits). 

\begin{table}[t]\label{table: Fedcau}
    \centering
    \caption{Comparison between FedCau and FedAvg with and without LAQ and Top-$q$, $M = 50$, and $K^{\max} = 200$.}
    \label{tab:List of Symbols and parameters}
    \renewcommand{\tabcolsep}{4pt}
    \renewcommand{\arraystretch}{0.5}
    \begin{tabular}{|m{3.1cm}|m{1.1cm}| m{1.27cm}| m{1.2cm}|}
     \hline
        \textbf{Method} & \textbf{Stop iteration} & \textbf{Total cost~(Mbits)} & \textbf{Test accuracy~(\%)} \\ \hline
       FedCau LAQ, $b=2$ &57 & $4.56$ & $94.2$ \\ \hline
          \vspace{0.0011\textheight}
        FedCau LAQ, $b=10$ & $43$ & $16.92$ & $87.8$ \\ \hline 
         FedCau Top-$q$, $q=0.1$ &$49$ & $6.19$ & $92.4$ \\ \hline
         FedCau Top-$q$, $q=0.6$ &$43$ & $32.4$ &  $80.9$ \\ \hline 
          {\color{black} FedAvg, LAQ, $b=2$} & {\color{black}$55$} & {\color{black}$4.4$} & {\color{black}$91.13$} \\ \hline
          {\color{black} FedAvg, LAQ, $b=2$} & {\color{black}$60$} & {\color{black}$4.8$} & {\color{black}$94.46$} \\ \hline 
         
        {\color{black} FedAvg} & {\color{black}$55$} & {\color{black}$68.992$} & {\color{black}$94.08$} \\ \hline 
        {\color{black} FedAvg} & {\color{black}$56$} & {\color{black}$70.24$} & {\color{black}$95.9$} \\ \hline 
        FedCau &$56$ & $70.24$ & $96.4$ \\ \hline
        {\color{black} FedAvg} & {\color{black}$57$} & {\color{black}$71.5$} & {\color{black}$96.54$} \\ \hline 
        
        {\color{black} FedAvg} & {\color{black}$60$} & {\color{black}$75.264$} & {\color{black}$96.782$} \\ \hline 
        FedAvg & $200$ & $250.88$ & $99.02$ \\ 
        \hline

\end{tabular}
\end{table}

{\color{black} 
To explore the trade-off between communication cost and test accuracy in the FedAvg baseline, we examine three stopping iterations, namely $55$, $56$, $57$, and $60$, which are close to the FedCau stopping iteration of $k_c=56$. {\color{black}We set these FedAvg stopping iterations because we have obtained~$k_c$ in FedCau. We choose the stopping iterations close to~$k_c$ for FedAvg for fair comparison and to show the superiority of FedCau in test accuracy and overall communication cost. We highlight that these stopping iterations for FedAvg cannot be set beforehand in practice.} When terminating FedAvg at iteration $55$, the achieved accuracy is $2.32$\% lower than FedCau, while offering a $1.82$\% reduction in communication cost. Similarly, FedAvg, with a stopping iteration of $57$, requires a $1.82$\% increase in communication cost to achieve a marginal improvement of $0.14$\% in test accuracy compared to FedCau. 
Furthermore, considering FedAvg at stopping iteration $60$, FedCau significantly saves $7.2$\% in the total cost with only a minor reduction of $0.382$\% in test accuracy compared to FedAvg. These findings highlight the effectiveness of FedCau in selecting the appropriate stopping iteration, ensuring that terminating the training before $k_c$ proves inefficient in terms of test accuracy while continuing after $k_c$ becomes costly with minimal improvements in accuracy. {Moreover, the results for FedAvg with stopping iteration of~$56$, the same as FedCau, show that FedCau with causal termination~$k_c$ outperforms FedAvg in test accuracy. 
}
Furthermore, we compare FedCau LAQ $b=2$ and $k_c=57$ with FedAvg LAQ $b=2$ and stop iterations of $55$ and $60$. The test accuracy results indicate that FedCau LAQ with $b=2$ outperforms FedAvg by increasing the test accuracy by $3.07$\% at the cost of $3.6$\% higher iteration cost. Thus, FedCau achieves the optimal causal stopping iteration in the context of LAQ with $b=2$, considering the trade-off between test accuracy and iteration cost. Furthermore, comparing FedCau with FedAvg at a stopping iteration of $60$, FedAvg achieves a test accuracy of $94.46$\% with an iteration cost of $4.8$~Mbits. Compared to FedCau at $k_c=57$, FedAvg incurs a $5.27$\% increase in iteration cost while gaining only a marginal $0.26$\% improvement in test accuracy. This comparison highlights that beyond $k_c$, the increase in iteration cost becomes significantly higher compared to the increase in test accuracy.
}

{\color{black} We conclude that FedCau obtains the optimal stopping iteration regarding the iteration cost and the achievable test accuracy, even when applying it on top of existing communication-efficient methods, such as LAQ and top-$q$.}
\section{Conclusion}\label{section: Conclusion}
In this paper, we proposed a framework to design cost-aware FL over networks. We characterized the communication-computation cost of running iterations of generic FL algorithms over a shared wireless channel regulated by slotted-ALOHA, CSMA/CA, and OFDMA protocols. We posed the communication-computation latency as the iteration-cost function of FL. We optimized the iteration-termination criteria to minimize the trade-off between FL's achievable objective value and the overall training cost. To this end, we proposed a causal setting, FedCau, utilized in two convex scenarios for batch and mini-batch updates, and for non-convex scenarios as well.

The numerical results showed that in the same background traffic, time budget, and network situation, CSMA/CA has less communication-computation cost than slotted-ALOHA. We also showed that the mini-batch FedCau update could perform more cost-efficiently than the batch update by choosing the proper time budgets. Moreover, the numerical results of the non-convex scenario provided a sub-optimal interval of the causal optimal solution close to the optimal interval, which provides many opportunities for non-convex FL problems. In the end, we applied the FedCau method on top of the existing methods like top-$q$ sparsification and LAQ with characterizing the iteration cost as the number of communication bits. {\color{black}We concluded that FedCau, with or without LAQ and top-$q$, obtains the causal termination iteration and, compared to FedAvg, achieves a significantly better trade-off between test accuracy and the total iteration cost of training.}

Our future work will extend the FedCau update of non-convex scenarios and design communication protocols for cost-efficient FL considering power allocation.



\appendices
\section{}\label{A: proof}






\subsection{Proof of Lemma~\ref{lemma: tightness}}\label{P: lemma: tightness}
The proof is directly obtained from the definitions of $F_u(\bw_k)$, $F_l(\bw_k)$, $\delta_k^u$ and $\delta_k^l$ in  Algorithm~\ref{alg: Nonconvex alg}. Recall that $F_u(\bw_k)=\Tilde{F}(\bw_k)$ (see line~24, 27, 30, 44) or $F_u(\bw_k)= F_u(\bw_{k-1})+\delta_k^l$ (see line~40) and the same arguments considering are valid for $F_l(\bw_k) = \Tilde{F}(\bw_k)$ or $F_l(\bw_k) = F_l(\bw_{k-1})+\delta_k^u$ (see lines 24, 33, 37, 46). Thus, by assuming a finite sequence of $|\Tilde{F}(\bw_k)|, k =1,\ldots, K$, the inequality $|F_u(\bw_k) - F_l(\bw_k)| \le \Tilde{F}_{\max}-\Tilde{F}_{\min}$ is the tightness between the upper bound~$F_u(\bw_k)$ and the lower bound~$F_l(\bw_k)$.

\vspace{-0.005\textheight}{\color{black}
\subsection{Proof of Proposition~\ref{prop: k_u and k_l}}\label{P: prop: k_u and k_l}
The stopping iteration~$k_c$ given by Algorithm~3 is in the form of an interval of $k_c \in [k_c^u, k_c^l]$. This interval's tightness depends on different scenarios, as we explain in the following. Assuming that Algorithm~3 has obtained $k_c^u$, after which we face several situations for updating~$F_l(\boldsymbol{w}_{k})$ according to the behavior of $ \Tilde{F}(\boldsymbol{w}_{k})$ for $k\ge k_c^u+1$. There are three different scenarios, which are explained in the following: 
\begin{itemize}
    \item $\Tilde{F}(\boldsymbol{w}_{k}) < \Tilde{F}(\boldsymbol{w}_{k-1})$ and $\Tilde{F}(\boldsymbol{w}_{k})>F_l(\boldsymbol{w}_{k-1})$:
    According to Algorithm~3 (see lines~44-47), for $k = k_c^u +1$, we have $F_u(\boldsymbol{w}_{k}) = \Tilde{F}(\boldsymbol{w}_{k})$ and $F_l(\boldsymbol{w}_{k}) = F_l(\boldsymbol{w}_{k-1})+\delta_k^u$ where $\delta_k^u = F_u(\boldsymbol{w}_{k})-F_u(\boldsymbol{w}_{k-1}) = \Tilde{F}(\boldsymbol{w}_{k})-F_u(\boldsymbol{w}_{k-1})$. Moreover, since $k \ge k_c^u +1$, the inequality of $G_u(k) > G_u(k)$ gives us 
    \begin{alignat}{3}
    \label{eq: Guk}
      G_u(k)&- G_u(k_c^u) = \left(\beta C(k) + (1-\beta)F_u(\boldsymbol{w}_{k})\right)  
      \\
\nonumber
&-\left(\beta C(k-1) +(1-\beta)F_u(\boldsymbol{w}_{k-1})\right) = \beta c_k\\
\nonumber
   &  - (1-\beta)(F_u(\boldsymbol{w}_{k-1})-F_u(\boldsymbol{w}_{k})) > 0.
    \end{alignat}
Then, we compute~$G_l(k)$ {as in Eq~\eqref{c: lowerG}}:
    \begin{alignat}{3}
\label{eq: sc1}
     G_l(k) &=&&  \beta C(k) + (1-\beta)F_l(\boldsymbol{w}_{k})\\
\nonumber
    &=&& \beta C(k-1)+ \beta c_k +(1-\beta)\left( F_l(\boldsymbol{w}_{k-1})+
        \delta_k^u\right) 
     \\
\nonumber
    &=&& \beta c_k +\beta C(k-1)+\\
\nonumber
    &&&(1-\beta)\left( F_l(\boldsymbol{w}_{k-1}) -F_u(\boldsymbol{w}_{k-1}) +\Tilde{F}(\boldsymbol{w}_{k})\right) \\
\nonumber
    &=&&  \beta c_k + G_l(k-1) - \\
\nonumber
    &&&(1-\beta)\left(F_u(\boldsymbol{w}_{k-1})-\Tilde{F}(\boldsymbol{w}_{k})\right) \stackrel{\small \text{ \eqref{eq: Guk}}}{>} G_l(k-1).
\end{alignat}
Therefore, $k_c^l = k_c^u +1 $.

     \item $\Tilde{F}(\boldsymbol{w}_{k}) < \Tilde{F}(\boldsymbol{w}_{k-1})$ and $\Tilde{F}(\boldsymbol{w}_{k})<F_l(\boldsymbol{w}_{k_{\max}^l})$:
    According to Algorithm~3 (see lines~36-44), for $k = k_c^u +1$, we have $F_l(\boldsymbol{w}_{k})=\Tilde{F}(\boldsymbol{w}_{k})$, and $F_u(\boldsymbol{w}_{k}) = F_u(\boldsymbol{w}_{k-1})+\delta_k^l$, where  $\delta_k^l = F_l(\boldsymbol{w}_{k})-F_l(\boldsymbol{w}_{k-1}) = \Tilde{F}(\boldsymbol{w}_{k})-F_l(\boldsymbol{w}_{k-1})$. Thus, we calculate $G_l(k)$ as
     \begin{alignat}{3}
\label{eq: sc2}
     G_l(k) &=&& \beta C(k) + (1-\beta)F_l(\boldsymbol{w}_{k})\\
\nonumber
    &=&& \beta C(k-1) + \beta c_k +(1-\beta)\Tilde{F}(\boldsymbol{w}_{k})\\
\nonumber
    &=&&\beta C(k-1) + \beta c_k +\\
\nonumber
    &&&  (1-\beta)\left( F_u(\boldsymbol{w}_{k})+F_l(\boldsymbol{w}_{k-1})-F_u(\boldsymbol{w}_{k-1})\right) \\
\nonumber
&=&& (1-\beta)\left (-F_u(\boldsymbol{w}_{k-1})+F_u(\boldsymbol{w}_{k})\right )+ \\
\nonumber
    &&& G_l(k-1) + \beta c_k  \stackrel{\small \text{ \eqref{eq: Guk}}}{>} G_l(k-1),
\end{alignat}
 which results in $k_c^l = k_c^u +1 $.  
\item $\Tilde{F}(\boldsymbol{w}_{k}) > F_u(\boldsymbol{w}_{k-1})$ (see lines~28-34 in Algorithm~3): In this case, the update of $F_u(\boldsymbol{w}_{k})$ and $F_l(\boldsymbol{w}_{k})$ are according to the linear update we proposed in Section~III-D in the revised manuscript. Thus, the update of $\delta_k^u$ is as
\begin{equation}\label{eq: sc3delta}
    \delta_k^u = \frac{\Tilde{F}(\boldsymbol{w}_{k}) - F_u(\boldsymbol{w}_{k_{\max}^u})}{k - k_{\max}^u},~\hspace{2mm} k \ge k_c^u +1,
\end{equation}
and $F_l(\boldsymbol{w}_{k})=F_l(\boldsymbol{w}_{k-1})+ \delta_k^u$, and $F_u(\boldsymbol{w}_{k}) = \Tilde{F}(\boldsymbol{w}_{k})$. Next, we calculate $G_l(k)$ as
\begin{alignat}{3}\label{eq: sc3}
  G_l(k) &=  \beta C(k) + (1-\beta)F_l(\boldsymbol{w}_{k}) \\
\nonumber
&= \beta C(k-1) + \beta c_k +(1-\beta)( F_l(\boldsymbol{w}_{k-1})+
        \delta_k^u)\\
\nonumber
    &= G_l(k-1) + \beta c_k +(1-\beta)\delta_k^u,
\end{alignat}
where $G_l(k)-G_l(k-1) = \beta c_k +(1-\beta)\delta_k^u$. Thus,~$k_c^l$ is obtained when $\beta c_k >-(1-\beta)\delta_k^u$, 
\begin{equation}
    -\delta_k^u = \frac{ F_u(\boldsymbol{w}_{k_{\max}^u})-\Tilde{F}(\boldsymbol{w}_{k}) }{k - k_{\max}^u} < \frac{\beta}{1-\beta}c_k,~\hspace{2mm} k \ge k_c^u +1,
\end{equation}
 where $k_c^l$ is 
 \begin{equation}
     k_c^l =k_{\max}^u +\left\lceil(1-\beta) \frac{F_u(\boldsymbol{w}_{k_{\max}^u})-\Tilde{F}(\boldsymbol{w}_{k})}{\beta c_k} \right \rceil+ 1.
 \end{equation}
\end{itemize}
\vspace{-0.01\textheight}
Therefore, according to the mentioned scenarios, we obtain
\begin{alignat}{3}\label{eq: P_k_u and k_l}
k_c^l &&&= 1 + \\
\nonumber
    &&&\max \left \{k_c^u, k_{\max}^u + \left \lceil \frac{(1-\beta)}{\beta c_{k_d}}\left \{F_u(\boldsymbol{w}_{k_{\max}^u})-\Tilde{F}(\boldsymbol{w}_{{k_d}}) \right \}  \right \rceil \right \},
\end{alignat}
where, for $k~\in [k_c^u+1, K^{\text{max}}]$,
\begin{equation}
   k_d :=  {\textrm{the first value of}}~k \mid  \Tilde{F}(\boldsymbol{w}_{k})>F_u(\boldsymbol{w}_{k-1}).
\end{equation}
}


\bibliographystyle{./MetaFiles/IEEEtran}
\bibliography{./MetaFiles/References}

\end{document}